\documentclass[sigconf]{aamas} 
\usepackage{balance} % for balancing columns on the final page
\usepackage[utf8]{inputenc} % allow utf-8 input
\usepackage[T1]{fontenc}    % use 8-bit T1 fonts
\usepackage{hyperref}       % hyperlinks
\usepackage{url}            % simple URL typesetting
\usepackage{booktabs}       % professional-quality tables
\usepackage{amsfonts}       % blackboard math symbols
\usepackage{nicefrac}       % compact symbols for 1/2, etc.
\usepackage{microtype}      % microtypography
\usepackage{amsmath,amsfonts,amsthm}
\newcommand{\argmax}{\mathop{\rm arg~max}\limits}
\newcommand{\argmin}{\mathop{\rm arg~min}\limits}

\usepackage{algorithm} %algorithmとalgorithmic環境を利用するのに必要．
\usepackage{algpseudocode}
\usepackage{bm}
\usepackage[inline]{enumitem}
\usepackage{multirow}
\usepackage{bbold}
\usepackage{adjustbox}
\allowdisplaybreaks[4]
\newtheorem{theorem}{Theorem}
\newtheorem{definition}{Definition}

\newtheorem{lemma}{Lemma}
\newtheorem{assumption}{Assumption}

\setcopyright{ifaamas}
\acmConference[AAMAS '21]{Proc.\@ of the 20th International Conference on Autonomous Agents and Multiagent Systems (AAMAS 2021)}{May 3--7, 2021}{Online}{U.~Endriss, A.~Now\'{e}, F.~Dignum, A.~Lomuscio (eds.)}
\copyrightyear{2021}
\acmYear{2021}
\acmDOI{}
\acmPrice{}
\acmISBN{}

\acmSubmissionID{191}

\title{Off-Policy Exploitability-Evaluation in Two-Player Zero-Sum Markov Games}

\author{Kenshi Abe}
\affiliation{
  \institution{CyberAgent, Inc.}
  \city{Shibuya, Tokyo}}
\email{abe\_kenshi@cyberagent.co.jp}

\author{Yusuke Kaneko}
\affiliation{
  \institution{CyberAgent, Inc.}
  \city{Shibuya, Tokyo}}
\email{kaneko\_yusuke@cyberagent.co.jp}

\begin{abstract}
Off-policy evaluation (OPE) is the problem of evaluating new policies using historical data obtained from a different policy. In the recent OPE context, most studies have focused on single-player cases, and not on multi-player cases. In this study, we propose OPE estimators constructed by the doubly robust and double reinforcement learning estimators in two-player zero-sum Markov games. The proposed estimators project exploitability that is often used as a metric for determining how close a policy profile (i.e., a tuple of policies) is to a Nash equilibrium in two-player zero-sum games. We prove the exploitability estimation error bounds for the proposed estimators. We then propose the methods to find the best candidate policy profile by selecting the policy profile that minimizes the estimated exploitability from a given policy profile class. We prove the regret bounds of the policy profiles selected by our methods. Finally, we demonstrate the effectiveness and performance of the proposed estimators through experiments.
\end{abstract}


\newcommand{\BibTeX}{\rm B\kern-.05em{\sc i\kern-.025em b}\kern-.08em\TeX}

\begin{document}

%%% The following commands remove the headers in your paper. For final 
%%% papers, these will be inserted during the pagination process.

\pagestyle{fancy}
\fancyhead{}

%%% The next command prints the information defined in the preamble.

\maketitle

%%%%%%%%%%%%%%%%%%%%%%%%%%%%%%%%%%%%%%%%%%%%%%%%%%%%%%%%%%%%%%%%%%%%%%%%

\section{Introduction}
Off-policy evaluation (OPE) is the problem of evaluating new policies using historical data obtained from a different policy.
Because online policy evaluation and learning are usually expensive or risky in various applications of reinforcement learning (RL), such as medicine \cite{murphy2003optimal} and education \cite{mandel2014offline}, OPE is attracting considerable interest \cite{athey2017efficient,kallus2019intrinsically,kitagawa2018should,liu2018breaking,swaminathan2015batch,thomas2016data,zhou2018offline}.
In the recent OPE context, most studies have focused on single-player cases rather than multi-player cases.

Multi-Agent Reinforcement Learning (MARL) is a generalization of single-agent RL for multi-agent environments.
It is widely applicable to situations where there are multi-agent interactions, such as security games, auctions, and negotiations.
In recent years, MARL has achieved many successes in the games Go \cite{silver2016mastering, silver2017mastering} and poker \cite{brown2019superhuman,brown2017libratus}.
MARL is a field with potential real-world applications, such as automated driving \cite{shalev2016safe}.

In this study, we propose OPE estimators in two-player zero-sum Markov games (TZMGs), which is one of the problems dealt with in MARL.
In general, existing OPE estimators in RL estimate the discounted value of a new policy.
However, estimating the discounted value is ineffective when the policy of the other player is unknown.
Unlike these estimators, for OPE in MARL, our OPE estimators evaluate a strategy profile by estimating exploitability, which is a metric for determining how close a strategy profile is to a Nash equilibrium in TZMG.
The proposed exploitability estimators are constructed by the doubly robust (DR) \cite{jiang2016doubly} and double reinforcement learning (DRL) \cite{kallus2019double} value estimators.
We prove that the proposed exploitability estimators are $\sqrt{n}$-consistent estimators for the true exploitability.

We also propose the methods to find the best candidate strategy profile from a given strategy profile class.
The proposed methods select the strategy profile that minimizes the exploitability projected by our exploitability estimators.
Then, we prove that we can consistently select the true lowest-exploitability policy profile using the proposed methods.

To demonstrate the effectiveness of our exploitability estimators, we compare our estimators to the estimators based on the following representative value estimators: importance sampling (IS), marginalized importance sampling (MIS), direct method (DM) value estimators.
The results show that the exploitability estimators based on the DR and DRL value estimators generally outperform the other estimator-based methods.
To the best of our knowledge, this is the first proposed estimators for exploitability for OPE in TZMGs.
    
\section{Preliminary}
\subsection{Two-Player Zero-Sum Markov Game}
A TZMG is defined as a tuple $\langle \mathcal{S}, \mathcal{A}_1, \mathcal{A}_2, T, P_I, P_T, P_R, \gamma \rangle$, where $\mathcal{S}$ represents a finite state space; $\mathcal{A}_i$ represents an action space for player $i \in \{1, 2\}$; $T$ represents a horizon; $P_I : \mathcal{S} \to [0, 1]$ represents an initial state distribution; $P_T : \mathcal{S} \times \mathcal{A}_1 \times \mathcal{A}_2 \times \mathcal{S} \to [0, 1]$ represents a transition probability function; $P_R : \mathcal{S} \times \mathcal{A}_1 \times \mathcal{A}_2 \times \mathbb{R} \to [0, 1]$ represents a reward distribution; and $\gamma \in [0, 1]$ represents a discount factor.
We define $R: \mathcal{S} \times \mathcal{A}_1 \times \mathcal{A}_2$ as a mean reward function of $P_R$.
For $t=1, \cdots, T$, we define $r_t\sim P_R(s_t,a_t^1, a_t^2)$ as a player 1's reward for taking actions $a_t^1$ and $a_t^2$ at state $s_t$, and define $-r_t$ as a player $2$'s reward.
Let $\pi_{i,t} : \mathcal{S} \times \mathcal{A}_i \to [0, 1]$ be a Markov policy for player $i$ at step $t\leq T$, and let $\pi_i=(\pi_{i,t})_{t\leq T}$.
We define $\pi=(\pi_1, \pi_2)$ as a strategy profile or a \textit{policy profile}.
The $T$-step discounted value of the policy profile $(\pi_1, \pi_2)$ for each player is represented as follows:
\begin{align*}
v_1(\pi_1, \pi_2) = \mathbb{E}_{\pi_1,\pi_2}[\sum_{t=1}^T \gamma^{t-1}r_t], ~v_2(\pi_1, \pi_2) = -v_1(\pi_1, \pi_2).
\end{align*}
We further define the state value function of state $s_t$ at step $t~(1\leq t\leq T)$ as follows:
\begin{align*}
V_{1,t}(s_t) &= \mathbb{E}_{\pi_1,\pi_2}[\sum_{k=t}^T \gamma^{k-t}r_k | s_t], ~V_{2,t}(s_t) = -V_{1,t}(s_t).
\end{align*}
Based on the state value function, we define the state-action value function of taking actions $a_t^1$ and $a_t^2$ at state $s_t$ as follows:
\begin{align*}
&Q_{1,t}(s_t, a_t^1, a_t^2) = R(s_t, a_t^1, a_t^2) + \mathbb{E}_{P_T}[\gamma V_{1,{t+1}}(s_{t+1}) | s_t, a_t^1, a_t^2], \\
&Q_{2,t}(s_t, a_t^1, a_t^2) = -Q_{1,t}(s_t, a_t^1, a_t^2).
\end{align*}
For a given policy profile $\pi$, we recursively define the marginal state-action distribution $p^{\pi}_t(s_t,a_t^1,a_t^2)$ at step $t$ as follows:
\begin{align*}
    &p^{\pi}_t(s_t,a_t^1,a_t^2) = \pi_{1,t}(a_{t}^1|s_{t})\pi_{2,t}(a_{t}^2|s_{t}) \\
    &\!\cdot\!\sum_{s_{t-1}\!\in \mathcal{S}}\sum_{a_{t-1}^1\!\in \mathcal{A}_1}\sum_{a_{t-1}^2\!\in \mathcal{A}_2}\!\!P_T(s_t|s_{t-1},a_{t-1}^1,a_{t-1}^2)p^{\pi}_{t-1}(s_{t-1},a_{t-1}^1,a_{t-1}^2),
\end{align*}
where $p^{\pi}_1(s_1,a_1^1,a_1^2) = \pi_{1,1}(a_{1}^1|s_{1})\pi_{2,1}(a_{1}^2|s_{1})P_I(s_1)$.

\subsection{Nash Equilibrium and Exploitability}
A common solution concept for two-player zero-sum games is a Nash equilibrium \cite{nash1951non, shapley1953stochastic}, where no player cannot improve by deviating from their specified strategy.
In TZMGs, a Nash equilibrium $\pi^{\star}=(\pi_1^{\star}, \pi_2^{\star})$ ensures the following condition:
\begin{equation}
\label{eq:nash}
\forall \pi_1 \in \Omega_1, ~\forall \pi_2 \in \Omega_2, ~v_1(\pi_1^{\star}, \pi_2) \geq v_1(\pi_1^{\star}, \pi_2^{\star}) \geq v_1(\pi_1, \pi_2^{\star}),
\end{equation}
where $\Omega_1$ and $\Omega_2$ are the \textit{whole policy sets}, i.e., the sets of all possible Markov policies for players 1 and 2, respectively.
The best response is a policy for player $i$ that is optimal against $\pi_{-i}$, where $\pi_{-i}$ is a policy for a player other than $i$.
Here, we introduce the value known as \textit{exploitability}, which is a metric for measuring how close a policy profile $\pi$ is to a Nash equilibrium $\pi^{\star}=(\pi_1^{\star}, \pi_2^{\star})$ in two-player zero-sum games.
Formally, the exploitability of $\pi_1,\pi_2$ is represented as follows:
\begin{align*}
v^{\mathrm{exp}}(\pi_1,\pi_2) &= \max_{\pi_2^{\prime}\in \Omega_2} v_2(\pi_1, \pi_2^{\prime}) - v_1(\pi_1, \pi_2)  \\
&+ \max_{\pi_1^{\prime}\in \Omega_1} v_1(\pi_1^{\prime}, \pi_2) - v_2(\pi_1, \pi_2) \\ 
&= \max_{\pi_1^{\prime}\in \Omega_1} v_1(\pi_1^{\prime}, \pi_2) + \max_{\pi_2^{\prime}\in \Omega_2} v_2(\pi_1, \pi_2^{\prime}).
\end{align*}
Note that in two-player zero-sum games, we can rewrite the exploitability as $v^{\mathrm{exp}}(\pi_1,\pi_2)=v_1(\pi_1^{\star}, \pi_2^{\star}) - \min_{\pi_2^{\prime}\in \Omega_2} v_1(\pi_1, \pi_2^{\prime}) + v_2(\pi_1^{\star}, \pi_2^{\star}) - \min_{\pi_1^{\prime}\in \Omega_1} v_2(\pi_1^{\prime}, \pi_2)$.
From the definition, a Nash equilibrium $\pi^{\star}$ has the lowest exploitability of $0$.

\section{Off-Policy Evaluation in Two-Player Zero-Sum Markov Games}
In this study, we assume that we can observe the \textit{historical data} 
\begin{equation*}
  \mathcal{D}=\{(s_{i,1}, a_{i,1}^1, a_{i,1}^2, r_{i,1}, \cdots, s_{i,T}, a_{i,T}^1, a_{i,T}^2, r_{i,T}, s_{i,T+1} )\}_{i=1}^{n}, 
\end{equation*}
where $n \in \mathbb{N}$ denotes the number of sampled trajectories.
The data is sampled using a fixed policy profile $\pi^b=(\pi_1^b, \pi_2^b)$.
We refer to this policy profile as a \textit{behavior policy profile}.
The distribution of $\mathcal{D}$ is then defined as follows:
\begin{align*}
   P_I(s_{1})\prod_{t=1}^T \pi_{1,t}^b(a_{t}^1 | s_{t})\pi_{2,t}^b(a_{t}^2 | s_{t})P_R(r_{t} | s_{t},a_{t}^1,a_{t}^2)P_T(s_{t+1} | s_{t}, a_{t}^1, a_{t}^2). 
\end{align*}

In most of the studies related to OPE, the goal is to estimate the discounted value of a given \textit{target policy} from the historical data.
However, this goal is not appropriate for multi-agent environments because, in general, in TZMGs, the policy of the opponent player is unknown, and one may play a game against a different policy than the target policy.
In this case, the discounted value of the target policy depends critically on the opponent player's policy.
Therefore, when the opponent policy is unknown, it is not worth estimating the discounted value against a specific policy.
In this study, for OPE in TZMGs, we estimate the exploitability of a given \textit{target policy profile} $\pi^e=(\pi_1^e,\pi_2^e)$ from the historical data instead of estimating the discounted value.
In other words, we estimate the value against the worst opponent policy for each player.

In this study, we assume that we are constrained to consider each player's policies within pre-defined policy classes $\Pi_1\subset \Omega_1$ and $\Pi_2\subset \Omega_2$.
In this case, if the best responses $\argmax_{\pi_1^{\prime}\in \Pi_1} v_1(\pi_1^{\prime}, \pi_2^e)$ and $\argmax_{\pi_2^{\prime}\in \Pi_2} v_2(\pi_1^e, \pi_2^{\prime})$ are not included in $\Pi_1$ and $\Pi_2$, we cannot calculate the true exploitability $v^{\mathrm{exp}}(\pi_1^e,\pi_2^e)$.
Therefore, instead of calculating $v^{\mathrm{exp}}(\pi_1^e,\pi_2^e)$, our exploitability estimators project the following value:
\begin{align*}
    v^{\mathrm{exp}}_{\Pi}(\pi_1^e,\pi_2^e)=\max_{\pi_1^{\prime}\in \Pi_1} v_1(\pi_1^{\prime}, \pi_2^e) + \max_{\pi_2^{\prime}\in \Pi_2} v_2(\pi_1^e, \pi_2^{\prime}),
\end{align*}
where $\Pi=\Pi_1\times\Pi_2$ is a policy profile class.
Note that our exploitability estimators project the exploitability from the historical data, without the structure information $P_I$, $P_T$, $P_R$, and $R$.
% In this setting, it is very difficult or impossible to compute exact exploitability.

\subsection{Notation}
For simplicity, we abbreviate terms like $V_1(s_t)$ as $V_{1,t}$.
For a policy profile $\pi$, we define the following variables (note that each variable implicitly depends on $\pi$):
\begin{itemize}
     \item $\eta_k=\frac{\pi_{1,k}(a_{k}^1 | s_{k})\pi_{2,k}(a_{k}^2 | s_{k})}{\pi_{1,k}^b(a_{k}^1 | s_{k})\pi_{2,k}^b(a_{k}^2 | s_{k})}$: the density ratio;
     \item $\rho_t=\prod_{k=1}^{t}\eta_k$: the cumulative density ratio;
     \item $\mu_t=\frac{p^{\pi}_t(s_t,a_t^1,a_t^2)}{p^{\pi^b}_t(s_t,a_t^1,a_t^2)}$: the marginal density ratio;
     \item $\hat{\pi}^b_i$: the estimators of $\pi^b_i$;
     \item $\hat{Q}_{1,t}$: the estimators of $Q_{1,t}$;
     \item $\hat{\rho}_t=\prod_{k=1}^{t}\frac{\pi_{1,k}(a_{k}^1 | s_{k})\pi_{2,k}(a_{k}^2 | s_{k})}{\hat{\pi}_{1,k}^b(a_{k}^1 | s_{k})\hat{\pi}_{2,k}^b(a_{k}^2 | s_{k})}$: the estimator of $\rho_t$.
\end{itemize}
Besides, we use the notation $\mathbb{E}_{\mathcal{D}}[f(X)]=\frac{1}{|\mathcal{D}|}\sum_{x\in \mathcal{D}} f(x)$ as an empirical average over $\mathcal{D}$, and we use $\mathbb{V}[\cdot]$ as a variance.

In the proofs presented in this study, we make the following assumptions regarding the overlapping of the policies and bounds of rewards and estimators, which are standard in the existing OPE literature \cite{zhou2018offline, kallus2019double, kallus2019efficiently}:
\begin{assumption}
    \label{asp:overlap_and_reward}
    $0\leq \eta_t\leq C$, $|r_t|\leq R_{\mathrm{max}}~$ for all $1\leq t\leq T$.
\end{assumption}
\begin{assumption}
    \label{asp:estimator_bound}
    $0\leq \hat{\rho}_t\leq C^t$, $0\leq \hat{\mu}_t\leq C^t$, $0\leq |\hat{Q}_{1,t}|\leq (T+1-t)R_{\mathrm{max}}~$ for all $1\leq t\leq T$.
\end{assumption}

\section{Off-Policy Value Estimators}
In this study, we construct the exploitability estimators using DR and DRL value estimators \cite{jiang2016doubly, kallus2019double}, which are the efficient estimators for the discounted value $v_i(\pi_1, \pi_2)$.
Therefore, in this section, we discuss the off-policy value evaluation and propose DR and DRL estimators for the discounted value in TZMGs.
To distinguish these estimators from the exploitability estimators, we refer to them as \textit{value estimators}.

\subsection{Efficiency Bound in Two-Player Zero-Sum Markov Games}
First, we discuss the (semiparametric) efficiency bound, which is the lower bound of the asymptotic mean squared error of OPE, among regular $\sqrt{n}$-consistent estimators.
Following the general literature \cite{tsiatis2007semiparametric}, we discuss the efficiency bound of the discounted value in TZMGs.
An efficiency bound is defined for estimators under several conjectured models of the data generating process.
If the conjectured model is parametric, the efficiency bound is equal to the Cram\'er-Rao lower bound.
Even if the conjectured model is non-parametric or semi-parametric, we can still define a corresponding Cram\'er-Rao lower bound.
Here, we introduce the following theorem from \cite{kallus2019double}.
\begin{theorem}[Efficiency bound in TZMGs]
\label{thm:eb}
The efficiency bound of $v_1(\pi_1,\pi_2)$ in TZMGs is
\begin{equation*}
\Upsilon_{\mathrm{EB}} = \mathbb{V}[V_{1,1}] + \sum_{t=1}^T\mathbb{E}[\gamma^{2(t-1)}\mu_{t}^2\mathbb{V}[r_t+\gamma V_{1,t+1}|s_t,a_t^1,a_t^2]],
\end{equation*}
where $V_{1,T+1}=0$.
\end{theorem}

\subsection{Efficient Off-Policy Value Estimators}
In this section, we propose the DR and DRL value estimators in TZMGs and their asymptotic properties.

{\bf Double Robust Estimator: }
We extend the DR value estimator for Markov decision processes (MDPs) proposed by \cite{jiang2016doubly} to apply to TZMGs.
For the theoretical guarantees, we consider the \textit{cross-fitting} version of the DR value estimator.
We split the historical data into $K$ evenly-sized folds.
Next, for each fold $k$, we construct estimators $\hat{\rho}_t^{-k}$ and $\hat{Q}_{1,t}^{-k}$ based on all the data except fold $k$.
We define the DR value estimator as follows:
\begin{align*}
&\hat{v}_{1}^{\mathrm{DR}}(\pi_1,\pi_2) \!=\! \mathbb{E}_{\mathcal{D}} \!\left[\sum_{t=1}^T \gamma^{t-1}\left(\hat{\rho}_{t}^{-k(i)} \left(r_{t} - \hat{Q}_{1,t}^{-k(i)} \right) \!+\! \hat{\rho}_{t-1}^{-k(i)}\hat{V}_{t}^{-k(i)}\right)\!\right], \\
&\hat{v}_{2}^{\mathrm{DR}}(\pi_1,\pi_2) = -v_{1}^{\mathrm{DR}}(\pi_1,\pi_2),
\end{align*}
where $\hat{V}_t^{-k(i)}=\mathbb{E}_{\pi}[\hat{Q}_{1,t}^{-k(i)} | s_t]$\footnote{$\mathbb{E}_{\pi}[\hat{Q}_{1,t}^{-k(i)} | s_t]$ is the expected value taken only over $a^1\sim \pi_{1,t}(a^1|s_t)$ and $a^2\sim \pi_{2,t}(a^2|s_t)$.} and $k(i)$ denotes the fold that contains the $i$-th data point.
By extending the proof of Theorem 4 in \cite{kallus2019double} to the case of TZMG, we can easily show the asymptotic property of the DR value estimator.
\begin{theorem}[Asymptotic property of the DR value estimator]
\label{thm:ef_dr}
Suppose $1\leq t\leq T, 1\leq k \leq K$, $\| \hat{Q}_{1,t}^{-k} - Q_{1,t} \|_2 = o_p(n^{{-\alpha_{1}}}) , \| \hat{\rho}_t^{-k} - \rho_t \|_2 = o_p(n^{{-\alpha_{2}}})$, where $\alpha_{1} > 0 , \alpha_{2} > 0$, and $\alpha_{1} + \alpha_{2} \geq 1/2$.
Then, 
\begin{align*}
    &\sqrt{n}(\hat{v}^{\mathrm{DR}}_1(\pi_1,\pi_2) - v_1(\pi_1,\pi_2)) \xrightarrow{d} \mathcal{N}(0 , \Upsilon^{\mathrm{DR}}), \\
    &\sqrt{n}(\hat{v}^{\mathrm{DR}}_2(\pi_1,\pi_2) - v_2(\pi_1,\pi_2)) \xrightarrow{d} \mathcal{N}(0 , \Upsilon^{\mathrm{DR}}),
\end{align*}
where $$\Upsilon^{\mathrm{DR}} = \mathbb{V}[V_{1,1}] + \sum_{t=1}^T\mathbb{E}[\gamma^{2(t-1)}\rho_{t}^2\mathbb{V}[r_t+\gamma V_{1,t+1}|\{s_k,a_k^1,a_k^2\}_{k=1}^t]],$$ and $V_{1,T+1}=0$.
\end{theorem}
The proof of this theorem is shown in Appendix \ref{sec:appendix_proof_asm_dr}.
As in \cite{jiang2016doubly,kallus2019double}, we can easily show that $\Upsilon^{\mathrm{DR}}$ is the semiparametric efficiency bound under games where the current state $s_t$ uniquely determines a trajectory.

{\bf Double Reinforcement Learning Estimator: }
In addition to the DR value estimator, we extend a DRL value estimator with cross-fitting for MDPs proposed by \cite{kallus2019double} to one for TZMGs.
The DRL value estimator is defined as follows:
\begin{align*}
\hat{v}^{\mathrm{DRL}}_1(\pi_1,\pi_2) & \!=\! \mathbb{E}_{\mathcal{D}}\!\left[\sum_{t=1}^T \gamma^{t-1}\left(\hat{\mu}_t^{-k(i)}\left(r_{t}\!-\hat{Q}_{1,t}^{-k(i)}\right) \!+\! \hat{\mu}_{t-1}^{-k(i)}\hat{V}_{1,t}^{-k(i)}\right)\!\right]\!, \\
\hat{v}^{\mathrm{DRL}}_2(\pi_1,\pi_2) &= -\hat{v}^{\mathrm{DRL}}_1(\pi_1,\pi_2).
\end{align*}
By extending the proof of Theorem 10 in \cite{kallus2019double} to the TZMG case, we can again show the asymptotic property of the DRL value estimator.
\begin{theorem}[Efficiency of the DRL value estimator]
\label{thm:ef_drl}
Suppose $1\leq t\leq T, 1\leq k \leq K$, $\| \hat{Q}_{1,t}^{-k} - Q_{1,t} \|_2 = o_p(n^{{-\alpha_{1}}}) , \| \hat{\mu}_t^{-k} - \mu_t \|_2 = o_p(n^{{-\alpha_{2}}})$, where $\alpha_{1} > 0 , \alpha_{2} > 0$, and $\alpha_{1} + \alpha_{2} \geq 1/2$.
Then, 
\begin{align*}
    &\sqrt{n}(\hat{v}_1^{\mathrm{DRL}}(\pi_1,\pi_2) - v_1(\pi_1,\pi_2)) \xrightarrow{d} \mathcal{N}(0 , \Upsilon_{\mathrm{EB}}), \\
    &\sqrt{n}(\hat{v}_2^{\mathrm{DRL}}(\pi_1,\pi_2) - v_1(\pi_1,\pi_2)) \xrightarrow{d} \mathcal{N}(0, \Upsilon_{\mathrm{EB}}),
\end{align*}
where $\Upsilon_{\mathrm{EB}}$ is an efficiency bound in Theorem \ref{thm:eb}.
\end{theorem}
According to this result, the DRL value estimator is efficient under mild assumptions, whereas the IS, MIS, DM, and DR estimators may be inefficient.

\subsection{Other Candidates of Value Estimators}
In this study, we compare our exploitability estimators to the estimators constructed by the IS, MIS, and DM value estimators.
This section summarizes these value estimators.

{\bf Importance Sampling Estimator:}
An IS estimator is represented as follows:
\begin{align*}
\hat{v}^{\mathrm{IS}}_1(\pi_1,\pi_2) = \mathbb{E}_{\mathcal{D}}\left[\sum_{t=1}^T \gamma^{t-1}\hat{\rho}_{t}r_{t}\right]
, \hat{v}^{\mathrm{IS}}_2(\pi_1,\pi_2) = -\hat{v}^{\mathrm{IS}}_1(\pi_1,\pi_2).
\end{align*}
When the behavior policy profile is known, i.e., $\hat{\rho}_t=\rho_t$, the IS estimator is an unbiased and consistent estimator of $v_1(\pi_1,\pi_2)$ and $v_2(\pi_1,\pi_2)$.
However, in general, the variance of the IS estimator grows exponentially with respect to horizon $T$ \cite{jiang2016doubly}.

{\bf Marginalized Importance Sampling Estimator:}
A MIS estimator is represented as follows:
\begin{align*}
    \hat{v}^{\mathrm{MIS}}_1(\pi_1,\pi_2) = \mathbb{E}_{\mathcal{D}}\left[\sum_{t=1}^T \gamma^{
    t-1} \hat{\mu}_t r_t\right], \hat{v}^{\mathrm{MIS}}_2(\pi_1,\pi_2) = -\hat{v}^{\mathrm{MIS}}_1(\pi_1,\pi_2).
\end{align*}
The MIS estimator can be regarded as one of the IS-type estimators.
Although the MIS estimator addresses the curse of horizon by exploiting the Markov decision process (MDP) structure, it is inefficient \cite{kallus2019double,xie2019towards}.

{\bf Direct Method Estimator:}
A DM estimator is represented as follows:
\begin{align*}
&\hat{v}^{\mathrm{DM}}_1(\pi_1,\pi_2)  = \mathbb{E}_{\mathcal{D}}\left[\mathbb{E}_{\pi}[\hat{Q}_{1,1}(s_1, a_1^1, a_1^2) | s_1]\right], \\
&\hat{v}^{\mathrm{DM}}_2(\pi_1,\pi_2)  = -\hat{v}^{\mathrm{DM}}_1(\pi_1,\pi_2).
\end{align*}
The DM estimator is not consistent if $\hat{Q}_{1,1}$ is not consistent, and it is not unbiased if $\hat{Q}_{1,1}$ is not correct.

\section{Off-Policy Exploitability Estimators}
\begin{algorithm}[t!]
  \caption{Off-Policy Exploitability Estimator with $\hat{v}_i^{\mathrm{DR}}$}
  \label{algo:ope_estimator_dr}
  \begin{algorithmic}[1]
    \Require Historical data $\mathcal{D}$
    \Require A target policy profile $\pi^e=(\pi^e_1, \pi^e_2)$
    \Require A policy classes $\Pi_1$ and $\Pi_2$
    \State Take a $K$-fold random partition $(I_k)_{k=1}^K$ of observation indices $\{1,\cdots,n\}$ such that the size of each fold $I_k$ is $n/K$.
    \State Let $\mathcal{D}_k=\{\mathcal{D}^{(i)} | i \in I_k\}, \mathcal{D}_{-k}=\{\mathcal{D}^{(i)} | i \notin I_k\}$
    \State Construct value estimators
    \begin{align*}
    &v^k_1(\pi_1, \pi_2)=\mathbb{E}_{\mathcal{D}_k}\left[\sum_{t=1}^T \gamma^{t-1}\left(\hat{\rho}_{t}^{-k} \left(r_{t} - \hat{Q}_{1,t}^{-k} \right) + \hat{\rho}_{t-1}^{-k}\hat{V}_{t}^{-k}\right)\right],\\
    &v^k_2(\pi_1, \pi_2)=\mathbb{E}_{\mathcal{D}_k}\left[\sum_{t=1}^T \gamma^{t-1}\left(\hat{\rho}_{t}^{-k} \left(-r_{t} + \hat{Q}_{1,t}^{-k} \right) - \hat{\rho}_{t-1}^{-k}\hat{V}_{t}^{-k}\right)\right],
    \end{align*}
    where $\hat{Q}_{1,t}^{-k}$ and $\hat{\rho}^{-k}_t$ are the estimators of $Q_{i,t}$ and $\rho_t$, rerspectively, constructed using $\mathcal{D}_{-k}$.
    \Ensure $\max\limits_{\pi_1\in \Pi_1}\frac{1}{K}\sum_{k=1}^K v^k_1(\pi_1,\pi_2^e) + \max\limits_{\pi_2\in \Pi_2}\frac{1}{K}\sum_{k=1}^K v^k_2(\pi_1^e,\pi_2)$
 \end{algorithmic}
\end{algorithm}

% \begin{algorithm}[t!]
%   \caption{Off-Policy Exploitability Estimator with $\hat{v}_1^{\mathrm{DRL}}$}
%   \label{algo:ope_estimator_drl}
%   \begin{algorithmic}[1]
%     \Require Historical data $\mathcal{D}$
%     \Require A target policy profile $\pi^e=(\pi^e_1, \pi^e_2)$
%     \Require A policy classes $\Pi_1$ and $\Pi_2$
%     \State Take a $K$-fold random partition $(I_k)_{k=1}^K$ of observation indices $\{1,\cdots,n\}$ such that the size of each fold $I_k$ is $n/K$.
%     \State Let $\mathcal{D}_k=\{\mathcal{D}^{(i)} | i \in I_k\}, \mathcal{D}_{-k}=\{\mathcal{D}^{(i)} | i \notin I_k\}$
%     \State Construct value estimators
%     \begin{align*}
%     &v^k_1(\pi_1, \pi_2)=\mathbb{E}_{\mathcal{D}_k}\left[\sum_{t=1}^T \gamma^{t-1}\left(\hat{\mu}_{t}^{-k} \left(r_{t} - \hat{Q}_{1,t}^{-k} \right) + \hat{\mu}_{t-1}^{-k}\hat{V}_{t}^{-k}\right)\right],\\
%     &v^k_2(\pi_1, \pi_2)=\mathbb{E}_{\mathcal{D}_k}\left[\sum_{t=1}^T \gamma^{t-1}\left(\hat{\mu}_{t}^{-k} \left(-r_{t} + \hat{Q}_{1,t}^{-k} \right) - \hat{\mu}_{t-1}^{-k}\hat{V}_{t}^{-k}\right)\right],
%     \end{align*}
%     where $\hat{Q}_{1,t}^{-k}, \hat{\mu}^{-k}_t$ are the estimators of $Q_{i,t}, \mu_t$ constructed using $\mathcal{D}_{-k}$.
%     \Ensure $\max\limits_{\pi_1\in \Pi_1}\frac{1}{K}\sum_{k=1}^K v^k_1(\pi_1,\pi_2^e) + \max\limits_{\pi_2\in \Pi_2}\frac{1}{K}\sum_{k=1}^K v^k_2(\pi_1^e,\pi_2)$
%  \end{algorithmic}
% \end{algorithm}

For OPE in TZMGs, we propose the following exploitability estimators constructed by the DR and DRL value estimators, respectively:
\begin{align*}
    \hat{v}^{\mathrm{exp}}_{\mathrm{DR}}(\pi^e_1,\pi^e_2) &=\max_{\pi_1\in \Pi_1} \hat{v}^{\mathrm{DR}}_1(\pi_1, \pi_2^e) + \max_{\pi_2\in \Pi_2} \hat{v}^{\mathrm{DR}}_2(\pi_1^e, \pi_2), \\
    \hat{v}^{\mathrm{exp}}_{\mathrm{DRL}}(\pi^e_1,\pi^e_2) &=\max_{\pi_1\in \Pi_1} \hat{v}^{\mathrm{DRL}}_1(\pi_1, \pi_2^e) + \max_{\pi_2\in \Pi_2} \hat{v}^{\mathrm{DRL}}_2(\pi_1^e, \pi_2).
\end{align*}
Similarly, we define $\hat{v}^{\mathrm{exp}}_{\mathrm{IS}}$, $\hat{v}^{\mathrm{exp}}_{\mathrm{MIS}}$, and $\hat{v}^{\mathrm{exp}}_{\mathrm{DM}}$ as the exploitability estimators based on the IS, MIS, and DM value estimators, respectively.
We present the pseudocode of the proposed estimator with $\hat{v}_i^{\mathrm{DR}}$ in Algorithm \ref{algo:ope_estimator_dr}.
The procedure of the exploitability estimator with $\hat{v}_i^{\mathrm{DRL}}$ is the same as Algorithm \ref{algo:ope_estimator_dr} except that $\hat{\rho}_t$ is replaced with $\hat{\mu}_t$.

In this section, we demonstrate the exploitability estimation error bounds of  $\hat{v}^{\mathrm{exp}}_{\mathrm{DR}}(\pi^e_1,\pi^e_2)$ and $\hat{v}^{\mathrm{exp}}_{\mathrm{DRL}}(\pi^e_1,\pi^e_2)$.
To obtain theoretical implications, we define the $\epsilon$-Hamming covering number $N_H(\epsilon, \Pi)$ under the Hamming distance $H_{n}(\pi^a,\pi^b)=\frac{1}{n}\sum_{i=1}^n\bm{1}(\{\bigvee_{t=1}^T\pi_{1,t}^a(s_{i,t})\neq\pi_{1,t}^b(s_{i,t})\}\vee\{\bigvee_{t=1}^T\pi_{2,t}^a(s_{i,t})\neq\pi_{2,t}^b(s_{i,t})\})$ and its entropy integral $\kappa(\Pi)=\int_{0}^{\infty}\sqrt{\log N_H(\epsilon^2,\Pi)}$.
In the proofs of the remaining theorems, we make the following assumptions on the covering number $N_H(\epsilon,\Pi)$:
\begin{assumption}
\label{asp:covering_number}
For any $0<\epsilon<1, N_H(\epsilon,\Pi)\leq D_1\exp(D_2(\frac{1}{\epsilon})^{\omega})$ for some constants $D_1,D_2>0, 0\leq \omega<0.5$.
\end{assumption}
Assumption \ref{asp:covering_number} is precisely the same as the assumption in the proof of \cite{zhou2018offline,kato2020off}, and this is not strong assumption \cite{zhou2018offline}.
Furthermore, to establish uniform error bounds on $\hat{Q}_{1,t}$ and $\hat{\mu}_{t}$, in the remaining theorems, we assume that $\hat{Q}_{1,t}$ and $\hat{\mu}_{t}$ are computed using the estimated TZMG model $\hat{R}$, $\hat{P}_T$, $\hat{p}^{\pi^b}_t$.
Under similar consistency assumptions as in \cite{zhou2018offline, kato2020off}, the estimation error bounds of $\hat{v}^{\mathrm{exp}}_{\mathrm{DR}}$ and $\hat{v}^{\mathrm{exp}}_{\mathrm{DRL}}$ are then obtained as follows:
\begin{theorem}[Estimation error bound of $\hat{v}^{\mathrm{exp}}_{\mathrm{DR}}(\pi^e_1,\pi^e_2)$]
\label{thm:ope_dr}
Let us define $\hat{\pi}_{l}^{b,-k}(a_{l}^1,a_{l}^2 | s_{l})\!=\!\hat{\pi}_{1,l}^{b,-k}(a_{l}^1 | s_{l})\hat{\pi}_{2,l}^{b,-k}(a_{l}^2 | s_{l})$ and $\pi_{l}^{b}(a_{l}^1,a_{l}^2 | s_{l})\!=\!\pi_{1,l}^{b}(a_{l}^1 | s_{l})\pi_{2,l}^{b}(a_{l}^2 | s_{l})$.
Assume Assumptions \ref{asp:overlap_and_reward}, \ref{asp:estimator_bound}, \ref{asp:covering_number},
\begin{enumerate*}[label=(\ref{thm:ope_dr}\alph*)]
    \item \label{aspp:dr_3} $1\leq t\leq T$ and $t\leq t^{\prime}\leq T$, 
\end{enumerate*}
\begin{align*}
    \mathbb{E}&\left[\left(\hat{R}^{-k}(s_{t^{\prime}},a_{t^{\prime}}^1,a_{t^{\prime}}^2)\prod_{l=t}^{{t^{\prime}}-1}\hat{P}_T^{-k}(s_{l+1}|s_{l},a_{l}^1,a_{l}^2)\right.\right.\\
    & \left.\left.- R(s_{t^{\prime}},a_{t^{\prime}}^1,a_{t^{\prime}}^2)\prod_{l=t}^{{t^{\prime}}-1}P_T(s_{l+1}|s_{l},a_{l}^1,a_{l}^2)\right)^2\right]=o(n^{-2\alpha_1}),
\end{align*}
and 
\begin{enumerate*}[label=(\ref{thm:ope_dr}b)]
    \item \label{aspp:dr_4} $1\leq t\leq T$, 
\end{enumerate*}
\begin{align*}
    \mathbb{E}\left[\left( \prod_{l=1}^{t}\frac{1}{\hat{\pi}_{l}^{b,-k}(a_{l}^1,a_{l}^2 | s_{l})} - \prod_{l=1}^{t}\frac{1}{\pi_{l}^{b}(a_{l}^1,a_{l}^2 | s_{l})}\right)^2\right]=o(n^{-2\alpha_2}),
\end{align*}
where $\alpha_1>0,\alpha_2>0$, and $\alpha_1+\alpha_2\geq 1/2$.
Then, for any $\delta>0$, there exists $C>0, N_{\delta}>0$, such that with probability at least $1-2\delta$ and for all $n\geq N_{\delta}$:
\begin{align*}
    |v^{\mathrm{exp}}_{\Pi}(\pi^e_1,\pi^e_2) - \hat{v}^{\mathrm{exp}}_{\mathrm{DR}}(\pi^e_1,\pi^e_2)| \leq C\left(\kappa(\Pi) + \sqrt{\log (1/\delta)}\right)\sqrt{\Upsilon^{\ast}_{\mathrm{DR}}/n},
\end{align*}
where $\Upsilon^{\ast}_{\mathrm{DR}}=\sup\limits_{\pi\in \Pi}\mathbb{E}\left[\left(\sum_{t=1}^T \gamma^{t-1}\left(\rho_t(r_t-Q_{1,t}) + \rho_{t-1}V_{1,t}\right)\right)^2\right]$.
\end{theorem}

\begin{theorem}[Estimation error bound of $\hat{v}^{\mathrm{exp}}_{\mathrm{DRL}}(\pi^e_1,\pi^e_2)$]
\label{thm:ope_drl}
Assume Assumptions \ref{asp:overlap_and_reward}, \ref{asp:estimator_bound}, \ref{asp:covering_number}, \ref{aspp:dr_3}, and
\begin{enumerate*}[label=(\ref{thm:ope_drl}\alph*)]
    \item \label{aspp:drl_1} $1\leq t\leq T$, 
\end{enumerate*}
\begin{align*}
    \mathbb{E}&\left[\left( \frac{\prod_{{t^{\prime}}=1}^{t} \hat{P}_T^{-k}(s_{{t^{\prime}}}|s_{{t^{\prime}}-1},a_{{t^{\prime}}-1}^1,a_{{t^{\prime}}-1}^2)}{\hat{p}_{b,t}^{-k}(s_t,a_t^1,a_t^2)} \right.\right. \\
    & \left.\left. - \frac{\prod_{{t^{\prime}}=1}^{t} P_T(s_{{t^{\prime}}}|s_{{t^{\prime}}-1},a_{{t^{\prime}}-1}^1,a_{{t^{\prime}}-1}^2)}{p_{b,t}(s_t,a_t^1,a_t^2)}\right)^2\right]=o(n^{-2\alpha_2}),
\end{align*}
where $\alpha_1>0,\alpha_2>0$, and $\alpha_1+\alpha_2\geq 1/2$.
Then, for any $\delta>0$, there exists $C>0, N_{\delta}>0$, such that with probability at least $1-2\delta$ and for all $n\geq N_{\delta}$:
\begin{align*}
    |v^{\mathrm{exp}}_{\Pi}(\pi^e_1,\pi^e_2) - \hat{v}^{\mathrm{exp}}_{\mathrm{DRL}}(\pi^e_1,\pi^e_2)| \leq C\left(\kappa(\Pi) + \sqrt{\log (1/\delta)}\right)\sqrt{\Upsilon^{\ast}_{\mathrm{DRL}}/n},
\end{align*}
where $\Upsilon^{\ast}_{\mathrm{DRL}}=\sup\limits_{\pi\in \Pi}\mathbb{E}\left[\left(\sum_{t=1}^T \gamma^{t-1}\left(\mu_t(r_t-Q_{1,t}) + \mu_{t-1}V_{1,t}\right)\right)^2\right]$.
\end{theorem}

% Assumptions \ref{aspp:dr_1} and \ref{aspp:dr_2} are precisely the same as the assumptions in the proof of \cite{zhou2018offline}, and these are not strong assumptions.
% For example, for the common policy class of finite-depth threes, these conditions are satisfied \cite{zhou2018offline}.
% Similarly, assumptions \ref{aspp:dr_3}, \ref{aspp:dr_4}, and \ref{aspp:drl_1} are also not strong or restrictive.
% When the estimators $\hat{R}^{-k}(s_{t^{\prime}},a_{t^{\prime}}^1,a_{t^{\prime}}^2)\prod_{l=t}^{{t^{\prime}}-1}\hat{P}_T^{-k}(s_{l+1}|s_{l},a_{l}^1,a_{l}^2)
% $ and $\prod_{l=1}^{t}\frac{1}{\hat{\pi}_1^{b,-k}(a_{l}^1 | s_{l})\hat{\pi}_2^{b,-k}(a_{l}^2 | s_{l})}$ are sufficiently smooth, Assumptions \ref{aspp:dr_3}, \ref{aspp:dr_4}, and \ref{aspp:drl_1} easily hold \cite{zhou2018offline}.

Theorems \ref{thm:ope_dr} and \ref{thm:ope_drl} mean that $\hat{v}^{\mathrm{exp}}_{\mathrm{DR}}$ and $\hat{v}^{\mathrm{exp}}_{\mathrm{DRL}}$ are $\sqrt{n}$-consistent estimators for the true exploitability defined among $\Pi$.
In particular, when $\Pi=\Omega_1\times\Omega_2$, the error between the estimated exploitability and the true exploitability $v^{\mathrm{exp}}(\pi^e_1,\pi^e_2)$ converges to $0$ at a rate $O_p(\frac{1}{\sqrt{n}})$.
Because $\Upsilon^{\ast}_{\mathrm{DR}}=\sup\limits_{\pi\in \Pi}(\Upsilon_{\mathrm{DR}}+v_1^2(\pi_1,\pi_2))$ and $\Upsilon^{\ast}_{\mathrm{DRL}}=\sup\limits_{\pi_1, \pi_2\in \Pi}(\Upsilon_{\mathrm{EB}}+v_1^2(\pi_1,\pi_2))$, it is necessary to use the value estimator with a small (asymptotic) variance to reduce the exploitability estimation error.
That is, the exploitability estimation error would be small using the value estimator with a small asymptotic variance.
Therefore, from Theorems \ref{thm:ef_dr} and \ref{thm:ef_drl}, using the efficient value estimator $\hat{v}^{\mathrm{exp}}_{\mathrm{DRL}}$ would result in a small estimation error.
Note that we do not assume that the behavior policy profile is known in Theorems \ref{thm:ope_dr} and \ref{thm:opl_drl}.
We sketch the proof of Theorem \ref{thm:ope_dr}.
The proof of Theorem \ref{thm:ope_drl} is almost the same as Theorem \ref{thm:ope_dr}.

\paragraph{Proof sketch of Theorem~\ref{thm:ope_dr}}
First, we define the DR value estimator with oracles $Q_{1,t}$ and $\rho_t$ as follows:
\begin{align*}
v_{1}^{\mathrm{DR}}(\pi^e_1,\pi^e_2) &= \mathbb{E}_{\mathcal{D}} \left[\sum_{t=1}^T \gamma^{t-1}\left(\rho_t \left(r_{t} - Q_{1,t} \right) + \rho_{t-1}V_{t}\right)\right], \\
v_{2}^{\mathrm{DR}}(\pi^e_1,\pi^e_2) &= -v_{1}^{\mathrm{DR}}(\pi^e_1,\pi^e_2).
\end{align*}
Besides, we define the value difference between two policy profiles $\pi^{\alpha}$ and $\pi^{\beta}$ in $\Pi$ as follows:
\begin{align*}
    \Delta(\pi^{\alpha},\pi^{\beta})&=v_1(\pi^{\alpha}_1,\pi^{\alpha}_2)-v_1(\pi^{\beta}_1,\pi^{\beta}_2), \\
    \hat{\Delta}(\pi^{\alpha},\pi^{\beta})&=\hat{v}^{\mathrm{DR}}_1(\pi^{\alpha}_1,\pi^{\alpha}_2)-\hat{v}^{\mathrm{DR}}_1(\pi^{\beta}_1,\pi^{\beta}_2), \\
    \tilde{\Delta}(\pi^{\alpha},\pi^{\beta})&=v_1^{\mathrm{DR}}(\pi^{\alpha}_1,\pi^{\alpha}_2)-v_1^{\mathrm{DR}}(\pi^{\beta}_1,\pi^{\beta}_2).
\end{align*}
We mainly show the uniform concentration of these difference functions following the proof of \cite{zhou2018offline}.

{\bf Uniform concentration of the difference of influence functions: }
First, we prove that the influence difference function $\tilde{\Delta}(\cdot,\cdot)$ concentrates uniformly around its mean $\Delta(\cdot,\cdot)$:
\begin{lemma}
\label{lem:tilde_delta_dr}
Under Assumptions \ref{asp:overlap_and_reward} and \ref{asp:covering_number}, for any $\delta>0$, with probability at least $1-2\delta$,
\begin{align*}
    \sup_{\pi^{\alpha}, \pi^{\beta} \in \Pi}&\left|\tilde{\Delta}(\pi^{\alpha},\pi^{\beta}) - \Delta(\pi^{\alpha},\pi^{\beta})\right| \\
    &\leq O\left(\left(\kappa(\Pi)+\sqrt{\log\frac{1}{\delta}}\right)\sqrt{\frac{\Upsilon^{\ast}_{\mathrm{DR}}}{n}}\right) + o(\frac{1}{\sqrt{n}}).
\end{align*}
\end{lemma}
The proof of Lemma \ref{lem:tilde_delta_dr} is the extension of the concentration result in \cite{zhou2018offline} to the TZMG setting.
The proof of this lemma is shown in Appendix \ref{sec:appendix_proof_tilde_delta_dr}.

{\bf Uniform concentration of the estimated value difference function: }
Next, we prove that with high probability, the estimated value difference function $\hat{\Delta}(\cdot,\cdot)$ concentrates around $\tilde{\Delta}(\cdot,\cdot)$ uniformly at a rate $o_p(\frac{1}{\sqrt{n}})$:
\begin{lemma}
\label{lem:tilde_hat_delta_dr}
Under Assumptions \ref{asp:overlap_and_reward}, \ref{asp:estimator_bound}, \ref{asp:covering_number}, \ref{aspp:dr_3}-\ref{aspp:dr_4}:
\begin{align*}
    \sup_{\pi_{\alpha},\pi_{\beta}\in\Pi} \left|\hat{\Delta}(\pi^{\alpha},\pi^{\beta}) - \tilde{\Delta}(\pi^{\alpha},\pi^{\beta})\right| = o_p(\frac{1}{\sqrt{n}}).
\end{align*}
\end{lemma}
The proof of this lemma is shown in Appendix \ref{sec:appendix_proof_tilde_hat_delta_dr}.
Here, we have:
\begin{align*}
    &\sup_{\pi^{\alpha},\pi^{\beta}\in \Pi}\left|\hat{\Delta}(\pi^{\alpha}, \pi^{\beta}) - \Delta(\pi^{\alpha}, \pi^{\beta})\right| \\
    &\!\leq\! \sup_{\pi^{\alpha},\pi^{\beta}\in \Pi}\!\left|\hat{\Delta}(\pi^{\alpha}, \pi^{\beta}) \!-\! \tilde{\Delta}(\pi^{\alpha}, \pi^{\beta}) \!-\! \Delta(\pi^{\alpha}, \pi^{\beta}) \!+\! \tilde{\Delta}(\pi^{\alpha}, \pi^{\beta})\right| \\
    &\leq \sup_{\pi^{\alpha},\pi^{\beta}\in \Pi}\left|\hat{\Delta}(\pi^{\alpha}, \pi^{\beta}) - \tilde{\Delta}(\pi^{\alpha}, \pi^{\beta})\right| \\
    &+ \sup_{\pi^{\alpha},\pi^{\beta}\in \Pi}\left|\tilde{\Delta}(\pi^{\alpha}, \pi^{\beta}) - \Delta(\pi^{\alpha}, \pi^{\beta})\right|.
\end{align*}
Therefore, combining Lemmas \ref{lem:tilde_delta_dr} and \ref{lem:tilde_hat_delta_dr}, we can show the uniform concentration of $\hat{\Delta}(\cdot,\cdot)$ on $\Delta(\cdot,\cdot)$:
\begin{lemma}
\label{lem:hat_delta_dr}
Assume Assumptions \ref{asp:overlap_and_reward}, \ref{asp:estimator_bound}, \ref{asp:covering_number}, \ref{aspp:dr_3}-\ref{aspp:dr_4}.
Then, for any $\delta>0$, there exists $C>0, N_{\delta}>0$, such that with probability at least $1-2\delta$ and for all $n\geq N_{\delta}$:
\begin{align*}
    \sup_{\pi^{\alpha},\pi^{\beta}\in\Pi} \!\left|\hat{\Delta}(\pi^{\alpha},\pi^{\beta}) - \Delta(\pi^{\alpha},\pi^{\beta})\right| \!\leq\! C\left(\kappa(\Pi) \!+\! \sqrt{\log (1/\delta)}\right)\!\sqrt{\frac{\Upsilon^{\ast}_{\mathrm{DR}}}{n}}.
\end{align*}
\end{lemma}

{\bf Estimation error bound of the exploitability estimator: }
Next, we define the best response and the estimated best response as follows:
\begin{align*}
    &\pi_1^{\dagger}=\argmax_{\pi_1\in \Pi_1} v_1(\pi_1,\pi^e_2), ~\pi_2^{\dagger}=\argmax_{\pi_2\in \Pi_2} v_2(\pi^e_1,\pi_2), \\
    &\hat{\pi}_1^{\dagger}=\argmax_{\pi_1\in \Pi_1}\hat{v}_1^{\mathrm{DR}}(\pi_1,\pi^e_2), ~\hat{\pi}_2^{\dagger}=\argmax_{\pi_2\in \Pi_2}\hat{v}_2^{\mathrm{DR}}(\pi^e_1,\pi_2).
\end{align*}
Then, by some algebra, we have:
\begin{align*}
    &v_{\Pi}^{\mathrm{exp}}(\pi^e_1,\pi^e_2) - \hat{v}_{\mathrm{DR}}^{\mathrm{exp}}(\pi^e_1,\pi^e_2) \\
    &\!\leq\! 3\!\sup_{\pi^{\alpha}\in \Pi,\pi^{\beta}\in \Pi}\!|\Delta((\pi^{\alpha}_1,\pi^{\alpha}_2),(\pi^{\beta}_1,\pi^{\beta}_2)) \!-\! \hat{\Delta}((\pi^{\alpha}_1,\pi^{\alpha}_2),(\pi^{\beta}_1,\pi^{\beta}_2))|,
\end{align*}
and
\begin{align*}
    & v_{\Pi}^{\mathrm{exp}}(\pi^e_1,\pi^e_2) - \hat{v}_{\mathrm{DR}}^{\mathrm{exp}}(\pi^e_1,\pi^e_2) \\
    &\!\geq\! -3\!\sup_{\pi^{\alpha}\in \Pi,\pi^{\beta}\in \Pi}\!|\Delta((\pi^{\alpha}_1,\pi^{\alpha}_2),(\pi^{\beta}_1,\pi^{\beta}_2)) \!-\! \hat{\Delta}((\pi^{\alpha}_1,\pi^{\alpha}_2),(\pi^{\beta}_1,\pi^{\beta}_2))|.
\end{align*}
Therefore, we have:
\begin{align*}
    &|v_{\Pi}^{\mathrm{exp}}(\pi^e_1,\pi^e_2) - \hat{v}_{\mathrm{DR}}^{\mathrm{exp}}(\pi^e_1,\pi^e_2)| \\
    &\!\leq\! 3\!\sup_{\pi^{\alpha}\in \Pi,\pi^{\beta}\in \Pi}\!|\Delta((\pi^{\alpha}_1,\pi^{\alpha}_2),(\pi^{\beta}_1,\pi^{\beta}_2)) \!-\! \hat{\Delta}((\pi^{\alpha}_1,\pi^{\alpha}_2),(\pi^{\beta}_1,\pi^{\beta}_2))|.
\end{align*}
Then, from Lemma \ref{lem:hat_delta_dr} and this equation, the statement is concluded.
For further details on the proof, see Appendix \ref{sec:appendix_proof_ope_dr}.

\section{Best Evaluation Policy Profile Selection}
\begin{algorithm}[t!]
  \caption{Off-Policy Best Evaluation Policy Profile Selection with $\hat{v}^{\mathrm{exp}}_{\mathrm{DR}}$}
  \label{algo:opl_estimator_dr}
  \begin{algorithmic}[1]
    \Require Historical data $\mathcal{D}$
    \Require A policy classes $\Pi_1$ and $\Pi_2$
    \State Take a $K$-fold random partition $(I_k)_{k=1}^K$ of observation indices $\{1,\cdots,n\}$ such that the size of each fold $I_k$ is $n/K$.
    \State Let $\mathcal{D}_k=\{\mathcal{D}^{(i)} | i \in I_k\}, \mathcal{D}_{-k}=\{\mathcal{D}^{(i)} | i \notin I_k\}$.
    \State Construct value estimators
    \begin{align*}
    &v^k_1(\pi_1, \pi_2)=\mathbb{E}_{\mathcal{D}_k}\left[\sum_{t=1}^T \gamma^{t-1}\left(\hat{\rho}_{t}^{-k} \left(r_{t} - \hat{Q}_{1,t}^{-k} \right) + \hat{\rho}_{t-1}^{-k}\hat{V}_{t}^{-k}\right)\right],\\
    &v^k_2(\pi_1, \pi_2)=\mathbb{E}_{\mathcal{D}_k}\left[\sum_{t=1}^T \gamma^{t-1}\left(\hat{\rho}_{t}^{-k} \left(-r_{t} + \hat{Q}_{1,t}^{-k} \right) - \hat{\rho}_{t-1}^{-k}\hat{V}_{t}^{-k}\right)\right],
    \end{align*}
    where $\hat{Q}_{1,t}^{-k}$ and $\hat{\rho}^{-k}_t$ are the estimators of $Q_{i,t}$ and $\rho_t$, respectively, constructed using $\mathcal{D}_{-k}$.
    \State Obtain $\hat{\pi}_1$ and $\hat{\pi}_2$ by solving the following optimization problem:
    \begin{align*}
        &\hat{\pi}_1=\argmax_{\pi_1\in \Pi_1}\min\limits_{\pi_2\in \Pi_2}\frac{1}{K}\sum_{k=1}^K v^k_1(\pi_1,\pi_2), \\
        &\hat{\pi}_2=\argmax_{\pi_2\in \Pi_2}\min\limits_{\pi_1\in \Pi_1}\frac{1}{K}\sum_{k=1}^K v^k_2(\pi_1,\pi_2).
    \end{align*}
    \Ensure $(\hat{\pi}_1, \hat{\pi}_2)$
 \end{algorithmic}
\end{algorithm}
In this section, we consider the problem of selecting the best candidate policy profile from a given policy profile class, one of the most practical applications of OPE.
For given historical data $\mathcal{D}$, our goal is to select the best policy profile with the lowest exploitability from the candidate policy profile class $\Pi$, i.e.,
\begin{align*}
    (\pi^{\ast}_1, \pi^{\ast}_2) = \argmin_{\pi_1,\pi_2\in \Pi_1\times \Pi_2} v^{\mathrm{exp}}_{\Pi}(\pi_1, \pi_2).
\end{align*}
According to Equation (\ref{eq:nash}), when $\Pi_1=\Omega_1$ and $\Pi_2=\Omega_2$, the policy profile $(\pi^{\ast}_1,\pi^{\ast}_2)$ is a Nash equilibrium.

To this end, we propose methods based on the exploitability estimators proposed in the previous section.
Based on the exploitability estimator $\hat{v}^{\mathrm{exp}}_{\mathrm{DR}}$, we select the policy profile that minimizes the estimated exploitability as follows:
\begin{align*}
    (\hat{\pi}_1^{\mathrm{DR}}, \hat{\pi}_2^{\mathrm{DR}}) = \argmin_{\pi_1,\pi_2\in \Pi_1\times \Pi_2} \hat{v}^{\mathrm{exp}}_{\mathrm{DR}}(\pi_1, \pi_2).
\end{align*}
From the definition of $\hat{v}^{\mathrm{exp}}_{\mathrm{DR}}$, we can rewrite the $\hat{\pi}_1^{\mathrm{DR}}$ and $\hat{\pi}_2^{\mathrm{DR}}$, respectively, as follows:
\begin{align*}
    \hat{\pi}_1^{\mathrm{DR}}&=\argmax_{\pi_1\in\Pi_1} \min_{\pi_2\in\Pi_2} \hat{v}_1^{\mathrm{DR}}(\pi_1,\pi_{2}), \\
    \hat{\pi}_2^{\mathrm{DR}}&=\argmax_{\pi_2\in\Pi_2} \min_{\pi_1\in\Pi_1} \hat{v}_2^{\mathrm{DR}}(\pi_1,\pi_{2}).
\end{align*}
Similarly, we define $\hat{\pi}^{\mathrm{IS}}$, $\hat{\pi}^{\mathrm{MIS}}$, $\hat{\pi}^{\mathrm{DM}}$, and $\hat{\pi}^{\mathrm{DRL}}$ as the estimators based on $\hat{v}^{\mathrm{exp}}_{\mathrm{IS}}$, $\hat{v}^{\mathrm{exp}}_{\mathrm{MIS}}$, $\hat{v}^{\mathrm{exp}}_{\mathrm{DM}}$, and $\hat{v}^{\mathrm{exp}}_{\mathrm{DRL}}$, respectively.
We describe the pseudocode of the proposed method with $\hat{v}^{\mathrm{exp}}_{\mathrm{DR}}$ in Algorithm \ref{algo:opl_estimator_dr}.
The procedure of the proposed method with $\hat{v}^{\mathrm{exp}}_{\mathrm{DRL}}$ is the same as Algorithm \ref{algo:opl_estimator_dr} except that $\hat{\rho}_t$ is replaced with $\hat{\mu}_t$.

We can derive the exploitability bounds of $\hat{\pi}^{\mathrm{DR}}$ and $\hat{\pi}^{\mathrm{DRL}}$ similarly as in the proofs of Theorems \ref{thm:ope_dr} and \ref{thm:ope_drl}.
\begin{theorem}[Exploitability bound of $\hat{\pi}^{\mathrm{DR}}$]
\label{thm:opl_dr}
Assume Assumptions \ref{asp:overlap_and_reward}, \ref{asp:estimator_bound}, \ref{asp:covering_number}, \ref{aspp:dr_3}-\ref{aspp:dr_4}.
Then, for any $\delta>0$, there exists $C>0, N_{\delta}>0$, such that with probability at least $1-2\delta$ and for all $n\geq N_{\delta}$:
\begin{align*}
    v^{\mathrm{exp}}(\hat{\pi}_1^{\mathrm{DR}},\hat{\pi}_2^{\mathrm{DR}}) - v^{\mathrm{exp}}(\pi_1^{\ast},\pi_2^{\ast}) \leq C\left(\kappa(\Pi) + \sqrt{\log (1/\delta)}\right)\sqrt{\frac{\Upsilon^{\ast}_{\mathrm{DR}}}{n}}.
\end{align*}
\end{theorem}

\begin{theorem}[Exploitability bound of $\hat{\pi}^{\mathrm{DRL}}$]
\label{thm:opl_drl}
Assume Assumptions \ref{asp:overlap_and_reward}, \ref{asp:estimator_bound}, \ref{asp:covering_number}, \ref{aspp:dr_3}, and \ref{aspp:drl_1}.
Then, for any $\delta>0$, there exists $C>0, N_{\delta}>0$, such that with probability at least $1-2\delta$ and for all $n\geq N_{\delta}$:
\begin{align*}
    v^{\mathrm{exp}}(\hat{\pi}_1^{\mathrm{DRL}},\hat{\pi}_2^{\mathrm{DRL}}) \!-\! v^{\mathrm{exp}}(\pi_1^{\ast},\pi_2^{\ast}) \!\leq\! C\left(\kappa(\Pi) + \sqrt{\log (1/\delta)}\right)\sqrt{\frac{\Upsilon^{\ast}_{\mathrm{DRL}}}{n}}.
\end{align*}
\end{theorem}

These theorems mean that we can consistently select the true lowest-exploitability policy profile $\pi^{\ast}$ using the proposed methods.
According to the minimax theorem, if $\Pi_1=\Omega_1$ and $\Pi_2=\Omega_2$, then $v^{\mathrm{exp}}(\pi_1^{\ast},\pi_2^{\ast})=0$.
Therefore, in this case, the exploitability of the selected policy profile converges asymptotically to $0$.
This means that the selected policy profile converges asymptotically to a Nash equilibrium when $\Pi_1=\Omega_1$ and $\Pi_2=\Omega_2$.
We sketch the proof of Theorem \ref{thm:opl_dr}.
The proof of Theorem \ref{thm:opl_drl} is almost the same as Theorem \ref{thm:opl_dr}.

\paragraph{Proof sketch of Theorem~\ref{thm:opl_dr}}
Let define:
\begin{align*}
    \mathcal{B}_i(\pi_{-i})=\argmax_{\pi_i^{\prime}\in \Omega_i} v_i(\pi_i^{\prime}, \pi_{-i}), ~ \hat{\mathcal{B}}_i(\pi_{-i})=\argmax_{\pi_i\in \Pi_i} \hat{v}^{\mathrm{DR}}_i(\pi_i, \pi_{-i}).
\end{align*}
Besides, for simplicity, we write $\hat{\pi}^{\mathrm{DR}}_i$ as $\hat{\pi}_i$ and $\hat{v}^{\mathrm{DR}}_i$ as $\hat{v}_i$.
From the definitions of $\pi^{\ast}_i$ and $\hat{\pi}_i$, we have:
\begin{align*}
    &v_1(\hat{\mathcal{B}}_1(\pi_2^{\ast}),\pi_2^{\ast}) \leq v_1(\mathcal{B}_1(\pi_2^{\ast}), \pi_2^{\ast}), \\
    &v_1(\pi_1^{\ast},\mathcal{B}_2(\pi_1^{\ast})) \leq v_1(\pi_1^{\ast},\hat{\mathcal{B}}_2(\pi_1^{\ast})), \\
    &\hat{v}_1(\mathcal{B}_1(\hat{\pi}_2),\hat{\pi}_2) \leq \hat{v}_1(\hat{\mathcal{B}}_1(\hat{\pi}_2),\hat{\pi}_2) \leq \hat{v}_1(\hat{\mathcal{B}}_1(\pi_2^{\ast}),\pi_2^{\ast}), \\
    &\hat{v}_1(\hat{\pi}_1, \mathcal{B}_2(\hat{\pi}_1)) \geq \hat{v}_1(\hat{\pi}_1, \hat{\mathcal{B}}_2(\hat{\pi}_1)) \geq \hat{v}_1(\pi_1^{\ast},\hat{\mathcal{B}}_2(\pi_1^{\ast})).
\end{align*}
Therefore, the exploitability bound of $\hat{\pi}$ is:
\begin{align*}
    &v^{\mathrm{exp}}(\hat{\pi}_1,\hat{\pi}_2) - v^{\mathrm{exp}}(\pi_1^{\ast},\pi_2^{\ast}) \\
    &= \Delta((\mathcal{B}_1(\hat{\pi}_2),\hat{\pi}_2), (\hat{\pi}_1,\mathcal{B}_2(\hat{\pi}_1))) - \hat{\Delta}((\mathcal{B}_1(\hat{\pi}_2),\hat{\pi}_2), (\hat{\pi}_1,\mathcal{B}_2(\hat{\pi}_1))) \\
    &- \Delta((\mathcal{B}_1(\pi_2^{\ast}), \pi_2^{\ast}), (\pi_1^{\ast},\mathcal{B}_2(\pi_1^{\ast}))) + \hat{\Delta}((\mathcal{B}_1(\pi_2^{\ast}), \pi_2^{\ast}), (\pi_1^{\ast},\mathcal{B}_2(\pi_1^{\ast}))) \\
    &\!+\! \hat{v}_1(\mathcal{B}_1(\hat{\pi}_2),\hat{\pi}_2) \!-\! \hat{v}_1(\hat{\pi}_1,\mathcal{B}_2(\hat{\pi}_1)) \!-\! \hat{v}_1(\mathcal{B}_1(\pi_2^{\ast}), \pi_2^{\ast}) \!+\! \hat{v}_1(\pi_1^{\ast},\mathcal{B}_2(\pi_1^{\ast})) \\
    &\leq \Delta((\mathcal{B}_1(\hat{\pi}_2),\hat{\pi}_2), (\hat{\pi}_1,\mathcal{B}_2(\hat{\pi}_1))) - \hat{\Delta}((\mathcal{B}_1(\hat{\pi}_2),\hat{\pi}_2), (\hat{\pi}_1,\mathcal{B}_2(\hat{\pi}_1))) \\
    &- \Delta((\mathcal{B}_1(\pi_2^{\ast}), \pi_2^{\ast}), (\pi_1^{\ast},\mathcal{B}_2(\pi_1^{\ast}))) + \hat{\Delta}((\mathcal{B}_1(\pi_2^{\ast}), \pi_2^{\ast}), (\pi_1^{\ast},\mathcal{B}_2(\pi_1^{\ast}))) \\
    &+ \hat{\Delta}((\hat{\mathcal{B}}_1(\pi_2^{\ast}),\pi_2^{\ast}),(\mathcal{B}_1(\pi_2^{\ast}), \pi_2^{\ast})) - \Delta((\hat{\mathcal{B}}_1(\pi_2^{\ast}),\pi_2^{\ast}),(\mathcal{B}_1(\pi_2^{\ast}), \pi_2^{\ast})) \\
    &- \hat{\Delta}((\pi_1^{\ast},\hat{\mathcal{B}}_2(\pi_1^{\ast})),(\pi_1^{\ast},\mathcal{B}_2(\pi_1^{\ast})))  + \Delta((\pi_1^{\ast},\hat{\mathcal{B}}_2(\pi_1^{\ast})),(\pi_1^{\ast},\mathcal{B}_2(\pi_1^{\ast}))) \\
    &\leq 4\sup_{\pi^{\alpha}\in \Pi,\pi^{\beta}\in \Pi}|\Delta((\pi^{\alpha}_1,\pi^{\alpha}_2),(\pi^{\beta}_1,\pi^{\beta}_2)) - \hat{\Delta}((\pi^{\alpha}_1,\pi^{\alpha}_2),(\pi^{\beta}_1,\pi^{\beta}_2))|.
\end{align*}
Then, from Lemma \ref{lem:hat_delta_dr} and this equation, the statement is concluded.

\section{Experiments}
We conduct experiments to analyze and evaluate the proposed exploitability estimators and the policy profile selection methods.
We conduct our experiments in two environments: repeated biased rock-paper-scissors (RBRPS) and Markov soccer \cite{littman1994markov}.

In all the experiments, we first prepare a near optimal policy profile $\pi_d$ using Minimax-Q learning \cite{littman1994markov}, after which we construct the behavior and target policy profiles using $\pi_d$.
We use an off-policy temporal difference learning \cite{sutton1998rli} to construct a Q-function model, and we use a histogram estimator for $\mu$, as in Section 5.2 in \cite{kallus2019double}.
In our experiments, we assume that the behavior policy profile is known and fixed.

\subsection{Environments}
\begin{figure*}[t!]
    \centering
    \includegraphics[width=\textwidth]{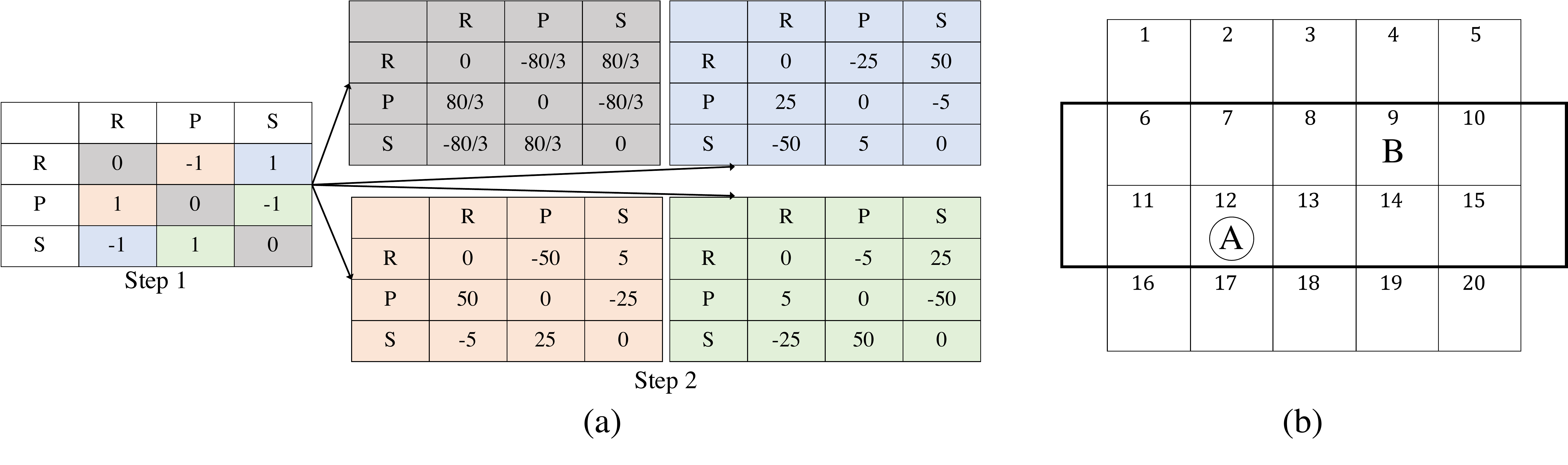}
    \caption{(a) Payoff matrices and a state transition graph in repeated biased rock-paper-scissors.
    When the result at the first step is a draw, the payoff matrix at the second step will be the gray one.
    When either player wins by rock/paper/scissors, the payoff matrix at the next step will be the blue/red/green one.
    (b) An initial board in Markov soccer.
    }
    \label{fig:env}
\end{figure*}
RBRPS is a simple TZMG where two players play an one-shot biased rock-paper-scissors game \cite{schaeffer2009comparing} multiple times.
We refer to a game that is repeated once as RBRPS1 and a game that is repeated two times as RBRPS2.
Note that RBRPS1 is precisely the same as the conventional rock-paper-scissors game.
Figure \ref{fig:env} (a) shows the payoff matrices and the state transition graph of RBRPS2.
In the first step, the payoff matrix is the same as in the conventional rock-paper-scissors game.
Depending on the result of the one-shot game, the next state and the payoff matrix transition.
There are five states in RBRPS2, and each state corresponds to each payoff matrix.

Markov soccer is a 1 vs 1 soccer game on a $4 \times 5$ grid , as shown in Figure \ref{fig:env} (b).
A and B denote players 1 and 2, respectively, and the circle in the figure represents the ball.
In each turn, each player can move to one of the neighboring cells or stay in place, and the actions of the two players are executed in random order.
When a player tries to move to the cell occupied by the other player, the ball's possession goes to the stationary player, and the positions of both players remain unchanged.
When the player with the ball reaches the goal (right of cell 10 or 15 for A, left of cell 6 or 11 for B), the game is over.
At this time, the player receives a reward of $+1$, and the opponent receives a reward of $-1$.
The player's positions and the ball's possession are initialized as shown in Figure \ref{fig:env} (b).

\subsection{Exploitability Evaluation}
In the first experiment, we compare the performance of $\hat{v}^{\mathrm{exp}}_{\mathrm{IS}}$, $\hat{v}^{\mathrm{exp}}_{\mathrm{MIS}}$,  $\hat{v}^{\mathrm{exp}}_{\mathrm{DM}}$, $\hat{v}^{\mathrm{exp}}_{\mathrm{DR}}$, and $\hat{v}^{\mathrm{exp}}_{\mathrm{DRL}}$ in RBRPS1 and RBRPS2.
We define the behavior policy profile as $\pi^b_1=0.7\pi^d_1 + 0.3\pi^{r}$ and $\pi^b_2=0.7\pi^d_2 + 0.3\pi^{p}$, where $\pi^r$ is a deterministic policy that always chooses rock, and $\pi^p$ is one that always chooses paper.
Similarly, we define the target policy profile as $\pi^e_1=0.9\pi^d_1 + 0.1\pi^{r}$ and $\pi^e_2=0.5\pi^d_2 + 0.5\pi^{p}$.
We define the policy classes as $\Pi_1=\Omega_1,\Pi_2=\Omega_2$.
We conduct $100$ trials using varying historical data sizes.

Tables \ref{tab:exp_ope_rps1} and \ref{tab:exp_ope_rps2} show the root-mean-squared error (RMSE) of each exploitability estimator in RBRPS1 and RBRPS2, where bold font indicates the best estimator in each case.
For further details on the results, see Appendix \ref{sec:appendix_results}.
We find that $\hat{v}^{\mathrm{exp}}_{\mathrm{DR}}$ and $\hat{v}^{\mathrm{exp}}_{\mathrm{DRL}}$ generally outperform the other estimators.
Note that $\hat{v}^{\mathrm{exp}}_{\mathrm{DRL}}$ has no advantage over $\hat{v}^{\mathrm{exp}}_{\mathrm{DR}}$ because the current state $s_t$ uniquely determines a trajectory.
Because the exploitability evaluation requires estimating best response value using historical data, the estimation error of the discounted value must be small.
Therefore, $\hat{v}^{\mathrm{exp}}_{\mathrm{DR}}$ and $\hat{v}^{\mathrm{exp}}_{\mathrm{DRL}}$, with a small estimation error of the discounted value, would perform better than the other estimators.

\subsection{Best Evaluation Policy Profile Selection}
In the second experiment, we analyze the performance of our policy profile selectors in RBRPS1, RBRPS2, and Markov soccer.
We compare the five policy profiles $\hat{\pi}^{\mathrm{IS}}$, $\hat{\pi}^{\mathrm{MIS}}$, $\hat{\pi}^{\mathrm{DM}}$, $\hat{\pi}^{\mathrm{DR}}$, and $\hat{\pi}^{\mathrm{DRL}}$, which are selected by each policy profile selector.

\begin{table}[t!]
    \centering
    \caption{Off-policy exploitability evaluation in RBRPS1: RMSE.}
    \begin{tabular}{c|c|c|c|c|c} \hline
         $N$ & $\hat{v}^{\mathrm{exp}}_{\mathrm{IS}}$ & $\hat{v}^{\mathrm{exp}}_{\mathrm{MIS}}$ & $\hat{v}^{\mathrm{exp}}_{\mathrm{DM}}$ & $\hat{v}^{\mathrm{exp}}_{\mathrm{DR}}$ & $\hat{v}^{\mathrm{exp}}_{\mathrm{DRL}}$ \\ \hline \hline
         $250$ & $0.085$ & $0.232$ & $4.8\times 10^{-3}$ & $\mathbf{3.6\times 10^{-3}}$ & $4.5\times 10^{-3}$ \\ \hline
         $500$ & $0.065$ & $0.230$ & $6.9\times 10^{-5}$ & $\mathbf{3.6\times 10^{-5}}$ & $6.1\times 10^{-5}$ \\ \hline
         $1000$ & $0.044$ & $0.226$ & $2.9\times 10^{-9}$ & $\mathbf{1.1\times 10^{-9}}$ & $2.5\times 10^{-9}$ \\ \hline
    \end{tabular}
    \label{tab:exp_ope_rps1}
\end{table}

\begin{table}[t!]
    \centering
    \caption{Off-policy exploitability evaluation in RBRPS2: RMSE.}
    \begin{tabular}{c|c|c|c|c|c} \hline
         $N$ & $\hat{v}^{\mathrm{exp}}_{\mathrm{IS}}$ & $\hat{v}^{\mathrm{exp}}_{\mathrm{MIS}}$ & $\hat{v}^{\mathrm{exp}}_{\mathrm{DM}}$ & $\hat{v}^{\mathrm{exp}}_{\mathrm{DR}}$ & $\hat{v}^{\mathrm{exp}}_{\mathrm{DRL}}$ \\ \hline\hline
         $250$ & $36.6$ & $11.3$ & $7.07$ & $8.98$ & $\mathbf{6.52}$ \\ \hline
         $500$ & $21.7$ & $11.2$ & $6.04$ & $6.10$ & $\mathbf{5.56}$ \\ \hline
         $1000$ & $15.5$ & $11.1$ & $4.87$ & $\mathbf{4.33}$ & $4.39$ \\ \hline
    \end{tabular}
    \label{tab:exp_ope_rps2}
\end{table}

\begin{table*}[t!]
    \centering
    \caption{Best evaluation policy profile selection in RBRPS: Exploitability (and standard errors).}
    \begin{tabular}{c|c|c|c|c|c|c} \hline
         & $\pi^b$ & $\hat{\pi}^{\mathrm{IS}}$ & $\hat{\pi}^{\mathrm{MIS}}$ & $\hat{\pi}^{\mathrm{DM}}$ & $\hat{\pi}^{\mathrm{DR}}$ & $\hat{\pi}^{\mathrm{DRL}}$ \\ \hline\hline
         RBRPS1 & $1.00$ & $0.236 (0.04)$ & $0.738 (0.05)$ & $0.058 (0.01)$ & $\mathbf{0.036 (0.01)}$ & $0.054 (0.01)$ \\ \hline
         RBRPS2 & $39.6$ & $29.2 (5.12)$ & $37.4 (4.33)$ & $22.5 (2.49)$ & $20.5 (0.66)$ & $\mathbf{19.4 (0.45)}$ \\ \hline
    \end{tabular}
    \label{tab:eq_opl_rps}
\end{table*}

\begin{table*}[t!]
    \centering
    \caption{Best evaluation policy profile selection in Markov soccer: Win rate $\times 100$ (and standard errors).}
    \begin{tabular}{|c|c|c|c|c|c|c|c|} \cline{3-8}
         \multicolumn{2}{c|}{} &\multicolumn{6}{c|}{Player 2}\\ \cline{3-8}
         \multicolumn{2}{c|}{} & $\pi^b_2$ & $\hat{\pi}^{\mathrm{IS}}_2$ & $\hat{\pi}^{\mathrm{MIS}}_2$ & $\hat{\pi}^{\mathrm{DM}}_2$ & $\hat{\pi}^{\mathrm{DR}}_2$ & $\hat{\pi}^{\mathrm{DRL}}_2$ \\ \cline{3-8}\hline
         \multirow{6}{*}{\rotatebox{90}{Player 1}} & $\pi^b_1$ & $48.9 (0.52)$ & $31.7 (9.5)$ & $54.2 (10.7)$ & $18.2 (3.4)$ & $22.6 (3.6)$ & $\mathbf{15.6 (0.9)}$ \\ \cline{2-8}
         & $\hat{\pi}^{\mathrm{IS}}_1$ & $81.2 (3.0)$ & $54.9 (7.9)$ & $74.9 (8.0)$ & $46.8 (6.0)$ & $53.5 (5.3)$ & $\mathbf{44.7 (4.7)}$ \\ \cline{2-8}
         & $\hat{\pi}^{\mathrm{MIS}}_1$ & $88.1 (1.6)$ & $65.5 (6.2)$ & $79.7 (6.4)$ & $57.8 (3.7)$ & $63.2 (5.0)$ & $\mathbf{55.5 (3.0)}$ \\ \cline{2-8}
         & $\hat{\pi}^{\mathrm{DM}}_1$ & $88.8 (3.1)$ & $65.5 (6.7)$ & $81.3 (6.2)$ & $58.3 (6.0)$ & $67.0 (4.5)$ & $\mathbf{56.7 (4.9)}$ \\ \cline{2-8}
         & $\hat{\pi}^{\mathrm{DR}}_1$ & $89.0 (3.0)$ & $\mathbf{70.0 (5.5)}$ & $82.0 (5.6)$ & $60.8 (5.8)$ & $66.2 (6.0)$ & $\mathbf{57.5 (4.1)}$ \\ \cline{2-8}
         & $\hat{\pi}^{\mathrm{DRL}}_1$ & $\mathbf{92.2 (1.5)}$ & $69.8 (5.9)$ & $\mathbf{82.5 (5.8)}$ & $\mathbf{63.6 (4.5)}$ & $\mathbf{71.0 (5.1)}$ & $\mathbf{62.4 (3.2)}$ \\ \hline
    \end{tabular}
    \label{tab:eq_opl_markov_soccer}
\end{table*}

In the experiments on RBRPS1 and RBRPS2, we define the behavior policy profile as $\pi^b_1=0.5\pi^d_1 + 0.5\pi^{r}$ and $\pi^b_2=0.5\pi^d_2 + 0.5\pi^{p}$.
We define the candidate policy classes as $\Pi_1=\Omega_1,\Pi_2=\Omega_2$ in RBRPS1, and set them to $\Pi_1=\{\{\alpha_1(s)\pi_1^d(s)+(1-\alpha_1(s))\pi^r(s)\}_{s\in S} | 0\leq \alpha_1(s)\leq 1\}$ and $\Pi_2=\{\{\alpha_2(s)\pi_2^d(s)+(1-\alpha_2(s))\pi^p(s)\}_{s\in S} | 0\leq \alpha_2(s)\leq 1\}$ in RBRPS2.
Note that the number of policy parameters is reduced to simplify minimax optimization in RBRPS2.
We conduct ten trials in each experiment with a historical data size of $250$.

Table \ref{tab:eq_opl_rps} shows the exploitability of each selected policy profile in RBRPS1 and RBRPS2.
We find that all selected policies are better than the behavior policy profile.
Again, bold font indicates the best policy profile in each case.
Notably, $\hat{\pi}^{\mathrm{DR}}$ and $\hat{\pi}^{\mathrm{DRL}}$ outperform the policy profiles obtained by the other estimators.

In the Markov soccer experiment, we define the behavior policy profile as $\pi^b_1=0.3\pi^d_1 + 0.7\pi^{u}$ and $\pi^b_2=0.5\pi^d_2 + 0.5\pi^{u}$, where $\pi^u$ is a uniform random policy.
We set the candidate policy classes to $\Pi_1=\{\alpha_1\pi_1^d+(1-\alpha_1)\pi^u | 0\leq \alpha_1\leq 1\}$ and $\Pi_2=\{\alpha_2\pi_2^d+(1-\alpha_2)\pi^u | 0\leq \alpha_1\leq 1\}$.
As before, we conduct ten trials in each experiment with a historical data size of $250$.
Because it is difficult to calculate the exploitability accurately in Markov soccer accurately, we compare the selected policy's winning rates against other policies.
Here, we approximate the winning rate using the rate of reaching the goal in $10,000$ games.
Note that player 1 has an advantage over player 2 because the possession of the ball always goes to player 1 at the initial state.

Table \ref{tab:eq_opl_markov_soccer} shows the winning rates of each selected policy in Markov soccer.
In this table, we show the winning rate of player 1.
The winning rates of $\hat{\pi}^{\mathrm{DRL}}_1$ and $\hat{\pi}^{\mathrm{DRL}}_2$ are generally higher than those of the other policies.
Unlike the results in RBRPS, the policy profile selected using $\hat{v}^{\mathrm{exp}}_{\mathrm{DRL}}$ is more robust and better than that obtained using $\hat{v}^{\mathrm{exp}}_{\mathrm{DR}}$.
These results suggest that we can select the policy profile the lowest exploitability when using $\hat{v}^{\mathrm{exp}}_{\mathrm{DRL}}$.

\section{Related Work}
In the context of OPE, there are many previous studies focusing on the theoretical properties of the value estimators, such as the IS \cite{hirano2003efficient}, MIS \cite{xie2019towards}, DR \cite{chernozhukov2018double,dudik2014doubly,farajtabar2018more,jiang2016doubly,liu2018representation,robins1994estimation,thomas2016data}, and DRL \cite{kallus2019double,kallus2019efficiently} estimators.
% Many of these studies are related to estimating the treatment effect, with the aim of adjusting for bias due to covariates.
% Our study is also located in the context of causal inference, in which the effect of online policies is estimated by correcting for bias using offline data.
In particular, the DRL estimator has the crucial advantage of using Markov properties to avoid the curse of horizon.
The main difference between these studies and our study is that we propose exploitability estimators for OPE in MARL. 

There are some studies on inverse MARL that assume the situation where the historical data is obtained in multi-agent environments \cite{lin2017multiagent,reddy2012inverse,wang2018competitive,natarajan2010multi,zhang2019non,yu2019multi}.
These studies differ from ours in that they aim to restore the reward function from the historical data.
In contrast, our study uses the historical data to estimate the exploitability of a given policy profile.
% Also, in the context of these studies, if there is enough data and the game is small, the parameters of the game can be recovered first and then the game can be solved.
% However, although these studies use DM for estimation, it is not an efficient estimation, as we have shown in our study.

MARL in Markov games has been studied extensively in the literature \cite{hu2003nash,littman1994markov,littman1996generalized,bai2020provable,zhang2019multi,busoniu2008comprehensive}.
Most existing studies on MARL focus on online policy learning.
In contrast, our study focuses on offline policy evaluation.

%As well as policy learning in Markov games, there are many previous studies on policy learning in extensive-form games.
%There are some diifferences between these studies and ours.
%First, our study is about a Markov game, not an extensive-form game.
%Secondly, counterfactual regret minimization (CFR), which is often used in the study of extensive-form games, is used for calculating a Nash equilibrium, not for policy estimation as we use it.
%Third, there are some studies on strategy evaluation in extensive-form games, but there is no paper on evaluation using exploitability as we have done.
As with policy learning in Markov games, there is a large body of literature on policy learning in extensive-form games \cite{mccracken2004safe,southey2009effective,zinkevich2008regret,davis2019low,schmid2019variance,gibson2012generalized}.
These studies focus on developing efficient method for computing Nash equilibria in extensive-form games, such as counterfactual regret minimization \cite{zinkevich2008regret}.
On the other hand, we focus on policy evaluation in Markov games.
Various works have investigated policy evaluation in extensive-form games \cite{zinkevich2006optimal,bowling2008strategy,johanson2009data,davidson2013baseline,bard2013online,davis2014using}.
While these studies have focused on online strategy evaluation with known structure, our study focuses on offline estimating exploitability without structural information.

There are several studies on the best policy selection in bandit problems or RL \cite{athey2017efficient,kitagawa2018should,swaminathan2015batch,zhou2018offline,kato2020off}.
Unlike these studies, we propose the policy selection methods in multi-agent settings.
Various studies on batch MARL \cite{perolat2017learning, zhang2018finite} also have considered the off-policy data setting.
The most significant difference between these studies and our study is that our study's main objective is to develop OPE estimators in MARL.
Furthermore, we consider the situation where candidate policies belong to a restricted policy class.
This has advantages in practical situations where only specific policies can be implemented.
% Nash-Q learning \cite{hu2003nash} and Minimax-Q learning \cite{littman1994markov,littman1996generalized} are learning algorithms, with a proven convergence. Unlike these studies, we derive the bounds of the exploitability of learned strategy profiles, depending on the number of trajectories.

\section{Conclusion}
In this study, we proposed estimators for TZMGs.
The proposed estimators project the exploitability of a target policy profile from historical data. 
We proved the exploitability estimation error bounds for the proposed estimators.
Besides, we proposed the methods for selecting the best policy profile from a given policy profile class based on our exploitability estimators.
We proved the exploitability bounds of the policy profiles selected by the proposed methods.
In future studies, we will explore the application of our exploitability estimators in more general settings, such as large extensive-form games.

%%%%%%%%%%%%%%%%%%%%%%%%%%%%%%%%%%%%%%%%%%%%%%%%%%%%%%%%%%%%%%%%%%%%%%%%

%%% The next two lines define, first, the bibliography style to be 
%%% applied, and, second, the bibliography file to be used.

\bibliographystyle{ACM-Reference-Format} 
\bibliography{ref}

\clearpage

\appendix
\newgeometry{
textheight=9in,
textwidth=5.5in,
top=30truemm,bottom=30truemm,left=25truemm,right=25truemm
}

\section{Notations}
\label{sec:appendix_notations}
In this section, we summarize the notation we use in Table \ref{tab:notation}.
We abbreviate terms like $Q_{1,t}(s_{i,t},a_{i,t}^1,a_{i,t}^2)$ as $Q_{1,i,t}$.
For simplicity, in our proofs, we assume that $|\mathcal{A}_1|=|\mathcal{A}_2|=d$.

\begin{table}[h!]
    \centering
    \caption{Notation}
    % \begin{adjustbox}{max width=\textwidth}
    \begin{tabular}{c|c} \hline
    $d$ & Number of possible actions $|\mathcal{A}_1|$ and $|\mathcal{A}_2|$ for each player \\ \hline
    $a_t$ & Tuple of actions $(a_{t}^{1},a_{t}^{2})$ at step $t$ \\ \hline
    $\pi_t(a_t|s_t)$ & Instantaneous density $\pi_{1,t}(a_t^1|s_t)\pi_{2,t}(a_t^2|s_t)$ \\ \hline
    $R(s_t,a_t)$ & Mean reward function $R(s_t,a_t^1,a_t^2)$ \\ \hline
    $Q_{1,t}(s_t,a_t)$ & Q-function $Q_{1,t}(s_t,a_t^1,a_t^2)$ at step $t$ \\ \hline
    $P(s_{t+1}|s_t,a_t)$ & Transition probability $P(s_{t+1}|s_t,a_t^1,a_t^2)$ \\ \hline
    $p_t^{\pi^b}(s_t,a_t),p_{b,t}(s_t,a_t)$ & Marginal state-action density $p_t^{\pi^b}(s_t,a_t^1,a_t^2)$ \\ \hline
    $\mathcal{D}$ & Historical data \\ \hline
    $\mathcal{D}_k$ & Historical data in fold $k$ \\ \hline
    $A\otimes B$ & \begin{tabular}{c}
         Kronecker product $\left(\begin{array}{ccc}
         a_{11}B & \cdots & a_{1n}B \\
         a_{m1}B & \cdots & a_{mn}B
    \end{array}\right)$,  \\
          where $A$ is a $m\times n$ matrix and $B$ is a $p\times q$ matrix.
    \end{tabular}\\ \hline
    $A_{t}^1$ & \begin{tabular}{c}
         $d$-dimensional vectors $A_{t}^1=(0,\cdots,1,\cdots,0)^T$ where $1$ appears and \\
         only appears in the $a_{i,t}^1$-th component and the rest are all zeros
    \end{tabular} \\ \hline
    $A_{t}^2$ & \begin{tabular}{c}
         $d$-dimensional vectors $A_{t}^2=(0,\cdots,1,\cdots,0)^T$ where $1$ appears and \\
         only appears in the $a_{i,t}^2$-th component and the rest are all zeros
    \end{tabular} \\ \hline
    $A_{t}$ & $A_{t}^1\otimes A_{t}^2$ \\ \hline
    $Q_{1,t}(s_t)$ & \begin{tabular}{c}
        Q-function vector at step $t$ \\
        $Q_{1,t}(s_t,a_1^1,a_t^2)=(Q_{1,t}(s_t,a_1^1,a_1^2),Q_{1,t}(s_t,a_1^1,a_2^2),\cdots,Q_{1,t}(s_t,a_1^d,a_1^d))^T$
    \end{tabular} \\ \hline
    $\pi_t(s_{t})$ & \begin{tabular}{c}
        Policy vector \\
        $\pi_t(s_t)=(\pi_{1,t}(a_1^1|s_t)\pi_{2,t}(a_1^2|s_t),\pi_{1,t}(a_1^1|s_t)\pi_{2,t}(a_2^2|s_t),\cdots,\pi_{1,t}(a_d^1|s_t)\pi_{2,t}(a_d^2|s_t))^T$
    \end{tabular} \\ \hline
    $\pi(s_{t^{\prime}:t})$ & $\pi(s_{t^{\prime}:t})=\pi_{t^{\prime}}(s_{t^{\prime}})\otimes \pi_{t^{\prime}+1}(s_{t^{\prime}+1})\otimes \cdots \otimes\pi_t(s_t)$ \\ \hline
    $\mathbb{1}_{\alpha}$ & $\alpha$-dimensional vector $(1,\cdots,1)^T$ where all components are $1$ \\ \hline
    $\mathbb{E}_{\mathcal{D}}[f(X)]$ & Empirical average $\frac{1}{|\mathcal{D}|}\sum_{X\in \mathcal{D}}f(X)$ \\ \hline
    $\mathbb{G}_{\mathcal{D}}[f(X)]$ & Empirical process $\sqrt{|\mathcal{D}|}(\mathbb{E}_{\mathcal{D}}[f(X)]-\mathbb{E}[f(X)])$ \\ \hline
    $\bigvee_{t=1}^T e_t $ & Logical disjunction $e_1\vee e_2\vee\cdots \vee e_t$. \\ \hline
    \end{tabular}
    % \end{adjustbox}
    \label{tab:notation}
\end{table}

\section{Proofs of Theorems}
\label{sec:appendix_proofs}
\subsection{Proof of Theorem \ref{thm:eb}}
\begin{proof}
We omit the proof since it is almost the same as Theorem 2 in \cite{kallus2019double}.
\end{proof}

\subsection{Proof of Theorem \ref{thm:ef_dr}}
\label{sec:appendix_proof_asm_dr}
\begin{proof}
We prove the statement following in \cite{kallus2019double}.
We define 
\begin{equation*}
    \psi(\{\hat{\rho}_t\},\{\hat{Q}_{1,t}\}) = \sum_{t=1}^T\gamma^{t-1}\left(\hat{\rho}_t r_t - \hat{\rho}_t\hat{Q}_{1,t} + \hat{\rho}_{t-1}\hat{V}_t\right).
\end{equation*}
Then , $\hat{v}^{\mathrm{DR}}_1(\pi_1,\pi_2)$ is given by
\begin{equation*}
    \sum_{k=1}^K\frac{n_k}{n} \mathbb{E}_{\mathcal{D}_k} [\psi(\{\hat{\rho}_t^{-k}\},\{\hat{Q}_{1,t}^{-k}\})],
\end{equation*}
where $n_k=|\mathcal{D}_k|$. 

Then, we have
\begin{align*}
    \sqrt{n}(\mathbb{E}_{\mathcal{D}_k} [\psi(\{\hat{\rho}_t^{-k}\},\{\hat{Q}_{1,t}^{-k}\})] -  v_1(\pi_1,\pi_2)) &= \sqrt{n/n_k} \mathbb{G}_{\mathcal{D}_k}[\psi(\{\hat{\rho}_t^{-k}\},\{\hat{Q}_{1,t}^{-k}\}) - \psi(\{\rho_t\} , \{Q_{1,t}\})] \\
    &+  \sqrt{n/n_k} \mathbb{G}_{\mathcal{D}_k} [\psi(\{\rho_t\} , \{Q_{1,t}\})] \\
    &+ \sqrt{n}(\mathbb{E}[\psi(\{\hat{\rho}_t^{-k}\},\{\hat{Q}_{1,t}^{-k}\}) | \{\hat{\rho}_t^{-k}\},\{\hat{Q}_{1,t}^{-k}\}] - v_1(\pi_1,\pi_2)).
\end{align*}
We analyze each term.
First , we prove that $\sqrt{n/n_k} \mathbb{G}_{\mathcal{D}_k}[\psi(\{\hat{\rho}_t^{-k}\},\{\hat{Q}_{1,t}^{-k}\}) - \psi(\{\rho_t\} , \{Q_{1,t}\})] = o_p(1)$.
If for any $\epsilon > 0$ , 
\begin{equation}
    \label{eq:limit_psi_con}
    \begin{aligned}
    \lim_{n_k \to \infty} \sqrt{n_k}P[&\mathbb{E}_{\mathcal{D}_k}[\psi(\{\hat{\rho}_t^{-k}\},\{\hat{Q}_{1,t}^{-k}\}) - \psi(\{\rho_t\} , \{Q_{1,t}\})] \\
    &- \mathbb{E}[\psi(\{\hat{\rho}_t^{-k}\},\{\hat{Q}_{1,t}^{-k}\}) - \psi(\{\rho_t\} , \{Q_{1,t}\}) | \{\hat{\rho}_t^{-k}\},\{\hat{Q}_{1,t}^{-k}\}] > \epsilon | \mathcal{D}_{-k}] = 0.
    \end{aligned}
\end{equation}
Then, from bounded convergence theorem, 
\begin{align*}
    \lim_{n_k \to \infty} \sqrt{n_k}P[&\mathbb{E}_{\mathcal{D}_k}[\psi(\{\hat{\rho}_t^{-k}\},\{\hat{Q}_{1,t}^{-k}\}) - \psi(\{\rho_t\} , \{Q_{1,t}\})] \\
    &- \mathbb{E}[\psi(\{\hat{\rho}_t^{-k}\},\{\hat{Q}_{1,t}^{-k}\}) - \psi(\{\rho_t\} , \{Q_{1,t}\}) | \{\hat{\rho}_t^{-k}\},\{\hat{Q}_{1,t}^{-k}\}] > \epsilon] = 0.
\end{align*}
To show Equation (\ref{eq:limit_psi_con}), we show that this conditional mean is 0 and conditional variance is $o_p(1)$.
The conditional mean part is
\begin{equation*}
    \mathbb{E}[\mathbb{E}_{\mathcal{D}_k}[\psi(\{\hat{\rho}_t^{-k}\},\{\hat{Q}_{1,t}^{-k}\}) - \psi(\{\rho_t\} , \{Q_{1,t}\})] - \mathbb{E}[\psi(\{\hat{\rho}_t^{-k}\},\{\hat{Q}_{1,t}^{-k}\}) - \psi(\{\rho_t\} , \{Q_{1,t}\}) | \{\hat{\rho}_t^{-k}\},\{\hat{Q}_{1,t}^{-k}\}] | \mathcal{D}_{-k}] = 0,
\end{equation*}
because $\{\hat{\rho}_t^{-k}\},\{\hat{Q}_{1,t}^{-k}\}$ only depend on $\mathcal{D}_{-k}$ and $\mathcal{D}_k$, $\mathcal{D}_{-k}$ are independent.
The conditional variance part is
\begin{align*}
    &\mathbb{V}[\sqrt{n_k}\mathbb{E}_{\mathcal{D}_k}[\psi(\{\hat{\rho}_t^{-k}\},\{\hat{Q}_{1,t}^{-k}\}) - \psi(\{\rho_t\} , \{Q_{1,t}\})] | \mathcal{D}_{-k}] = \mathbb{V}[\frac{1}{\sqrt{n_k}}\sum_{\mathcal{D}_k}\psi(\{\hat{\rho}_t^{-k}\},\{\hat{Q}_{1,t}^{-k}\}) - \psi(\{\rho_t\} , \{Q_{1,t}\}) | \mathcal{D}_{-k}] \\
    &= \frac{1}{n_k}\sum_{\mathcal{D}_k}\mathbb{V}[\psi(\{\hat{\rho}_t^{-k}\},\{\hat{Q}_{1,t}^{-k}\}) - \psi(\{\rho_t\} , \{Q_{1,t}\}) | \mathcal{D}_{-k}] = \mathbb{V}[\psi(\{\hat{\rho}_t^{-k}\},\{\hat{Q}_{1,t}^{-k}\}) - \psi(\{\rho_t\} , \{Q_{1,t}\}) | \mathcal{D}_{-k}] \\
    &\leq \mathbb{E}[D_1^2+D_2^2+D_3^2+2D_1D_2+2D_1D_2+2D_2D_3 | \mathcal{D}_{-k}] = T^2 \max\{o_p(n^{-2\alpha_{1}}),o_p(n^{-2\alpha_{2}}),o_p(n^{-\alpha_{1}-\alpha_{2}})\} \\
    &= o_p(1),
\end{align*}
where
\begin{align*}
    D_1 &= \sum_{t=1}^T \gamma^{t-1}\left((\hat{\rho}_t^{-k}-\rho_t)(-\hat{Q}_{1,t}^{-k}+Q_{1,t}) + (\hat{\rho}_{t-1}^{-k}-\rho_{t-1})(\hat{V}_t^{-k}-V_{1,t})\right), \\
    D_2 &= \sum_{t=1}^T \gamma^{t-1}\left(\rho_t(-\hat{Q}_{1,t}^{-k}+Q_{1,t}) + \rho_{t-1}(\hat{V}_t^{-k}-V_{1,t})\right), \\
    D_3 &= \sum_{t=1}^T \gamma^{t-1}\left((\hat{\rho}_t^{-k}-\rho_t)(r_t-Q_{1,t}+\gamma V_{1,t+1})\right).
\end{align*}
Here, we used the convergence rate assumption.
Then , from Chebyshev's inequality , 
\begin{align*}
    \sqrt{n_k}P[&\mathbb{E}_{\mathcal{D}_k}[\psi(\{\hat{\rho}_t^{-k}\},\{\hat{Q}_{1,t}^{-k}\}) - \psi(\{\rho_t\} , \{Q_{1,t}\})] \\
    &- \mathbb{E}[\psi(\{\hat{\rho}_t^{-k}\},\{\hat{Q}_{1,t}^{-k}\}) - \psi(\{\rho_t\} , \{Q_{1,t}\}) | \{\hat{\rho}_t^{-k}\},\{\hat{Q}_{1,t}^{-k}\}] > \epsilon | \mathcal{D}_{-k}] \\
    \leq \frac{1}{\epsilon^2} \mathbb{V}&[\sqrt{n_k}\mathbb{E}_{\mathcal{D}_k}[\psi(\{\hat{\rho}_t^{-k}\},\{\hat{Q}_{1,t}^{-k}\}) - \psi(\{\rho_t\} , \{Q_{1,t}\})] | \mathcal{D}_{-k}]  = o_p(1).
\end{align*}

Next, We prove that $\sqrt{n}(\mathbb{E}[\psi(\{\hat{\rho}_t^{-k}\},\{\hat{Q}_{1,t}^{-k}\}) | \{\hat{\rho}_t^{-k}\},\{\hat{Q}_{1,t}^{-k}\}] - v_1(\pi_1,\pi_2))$ is $o_p(1)$.
We have:
\begin{align*}
    &\sqrt{n}(\mathbb{E}[\psi(\{\hat{\rho}_t^{-k}\},\{\hat{Q}_{1,t}^{-k}\}) | \{\hat{\rho}_t^{-k}\},\{\hat{Q}_{1,t}^{-k}\}] - \mathbb{E}[\psi(\{\rho_t\},\{Q_{1,t}\}) | \{\hat{\rho}_t^{-k}\},\{\hat{Q}_{1,t}^{-k}\}]) \\
    &= \sqrt{n}\mathbb{E}[\sum_{t=1}^T \gamma^{t-1}\left((\hat{\rho}_t^{-k}-\rho_t)(-\hat{Q}_{1,t}^{-k}+Q_{1,t}) + (\hat{\rho}_{t-1}^{-k}-\rho_{t-1})(\hat{V}_t^{-k}-V_{1,t})\right) | \{\hat{\rho}_t^{-k}\},\{\hat{Q}_{1,t}^{-k}\}] \\
    &+ \sqrt{n}\mathbb{E}[\sum_{t=1}^T \gamma^{t-1}\left(\rho_t(-\hat{Q}_{1,t}^{-k}+Q_{1,t}) + \rho_{t-1}(\hat{V}_t^{-k}-V_{1,t})\right) | \{\hat{\rho}_t^{-k}\},\{\hat{Q}_{1,t}^{-k}\}] \\
    &+ \sqrt{n}\mathbb{E}[\sum_{t=1}^T \gamma^{t-1}\left((\hat{\rho}_t^{-k}-\rho_t)(r_t-Q_{1,t}+\gamma V_{1,t+1})\right) | \{\hat{\rho}_t^{-k}\},\{\hat{Q}_{1,t}^{-k}\}] \\
    &= \sqrt{n}\mathbb{E}[\sum_{t=1}^T \gamma^{t-1}\left((\hat{\rho}_t^{-k}-\rho_t)(-\hat{Q}_{1,t}^{-k}+Q_{1,t}) + (\hat{\rho}_{t-1}^{-k}-\rho_{t-1})(\hat{V}_t^{-k}-V_{1,t})\right) | \{\hat{\rho}_t^{-k}\},\{\hat{Q}_{1,t}^{-k}\}] \\
    &= \sqrt{n}\sum_{t=1}^T O\left(\|\hat{\rho}_t^{-k} - \rho_t \|_2 \|\hat{Q}_{1,t}^{-k} - Q_{1,t} \|_2\right) = \sqrt{n}\sum_{t=1}^T o_p(n^{-\alpha_1-\alpha_2}) = o_p(1).
\end{align*}

From above results, for $1\leq k\leq K$,
\begin{equation*}
    \sqrt{n}(\mathbb{E}_{\mathcal{D}_k} [\psi(\{\hat{\rho}_t^{-k}\},\{\hat{Q}_{1,t}^{-k}\})] -  v_1(\pi_1,\pi_2)) = \sqrt{n/n_k} \mathbb{G}_{\mathcal{D}_k} [\psi(\{\rho_t\} , \{Q_{1,t}\})] + o_p(1).
\end{equation*}
Therefore,
\begin{align*}
    &\sqrt{n}(\hat{v}^{\mathrm{DR}}_1(\pi_1,\pi_2)-v_1(\pi_1,\pi_2)) \\
    &= \sum_{k=1}^K\frac{n_k}{n}\sqrt{n}(\mathbb{E}_{\mathcal{D}_k} [\psi(\{\hat{\rho}_t^{-k}\},\{\hat{Q}_{1,t}^{-k}\})] -  v_1(\pi_1,\pi_2)) = \sum_{k=1}^K\sqrt{\frac{n_k}{n}} \mathbb{G}_{\mathcal{D}_k} [\psi(\{\rho_t\} , \{Q_{1,t}\})] + o_p(1) \\ 
    &\leq \mathbb{G}_{\mathcal{D}} [\psi(\{\rho_t\} , \{Q_{1,t}\})] + o_p(1).
\end{align*}
Here, we can easily show that
\begin{align*}
    \mathbb{V}[\psi(\{\rho_t\} , \{Q_{1,t}\})] = \mathbb{V}[V_{1,1}] + \sum_{t=1}^T\mathbb{E}[\gamma^{2(t-1)}\rho_{t}^2\mathbb{V}[r_t+\gamma V_{1,t+1}|s_1,a_1^1,a_1^2,\cdots,s_t,a_t^1,a_t^2]].
\end{align*}
Then, from Assumption \ref{asp:overlap_and_reward} and central limit theorem, this statement is concluded.
\end{proof}

\subsection{Proof of Theorem \ref{thm:ef_drl}}
\begin{proof}
The proof is similar to that of Theorem \ref{thm:ef_dr}.
\end{proof}

\subsection{Proof of Theorem \ref{thm:ope_dr}}
\label{sec:appendix_proof_ope_dr}
\begin{proof}
Let define
\begin{align*}
    \Delta(\pi^{\alpha},\pi^{\beta})&=v_1(\pi^{\alpha}_1,\pi^{\alpha}_2)-v_1(\pi^{\beta}_1,\pi^{\beta}_2), \\
    \hat{\Delta}(\pi^{\alpha},\pi^{\beta})&=\hat{v}^{\mathrm{DR}}_1(\pi^{\alpha}_1,\pi^{\alpha}_2)-\hat{v}^{\mathrm{DR}}_1(\pi^{\beta}_1,\pi^{\beta}_2), \\
    \tilde{\Delta}(\pi^{\alpha},\pi^{\beta})&=v_1^{\mathrm{DR}}(\pi^{\alpha}_1,\pi^{\alpha}_2)-v_1^{\mathrm{DR}}(\pi^{\beta}_1,\pi^{\beta}_2),
\end{align*}
and
\begin{align*}
    &\pi_1^{\dagger}=\argmax_{\pi_1\in \Pi_1} v_1(\pi_1,\pi^e_2), ~\pi_2^{\dagger}=\argmax_{\pi_2\in \Pi_2} v_2(\pi^e_1,\pi_2), \\
    &\hat{\pi}_1^{\dagger}=\argmax_{\pi_1\in \Pi_1}\hat{v}_1^{\mathrm{DR}}(\pi_1,\pi^e_2), ~\hat{\pi}_2^{\dagger}=\argmax_{\pi_2\in \Pi_2}\hat{v}_2^{\mathrm{DR}}(\pi^e_1,\pi_2).
\end{align*}
We have:
\begin{align*}
    &v_{\Pi}^{\mathrm{exp}}(\pi^e_1,\pi^e_2) - \hat{v}_{\mathrm{DR}}^{\mathrm{exp}}(\pi^e_1,\pi^e_2) = v_1(\pi_1^{\dagger},\pi^e_2) - \hat{v}_1^{\mathrm{DR}}(\hat{\pi}_1^{\dagger},\pi^e_2) + v_2(\pi^e_1,\pi_2^{\dagger})  - \hat{v}^{\mathrm{DR}}_2(\pi^e_1,\hat{\pi}_2^{\dagger}) \\
    &= \Delta((\pi_1^{\dagger},\pi^e_2),(\hat{\pi}_1^{\dagger},\pi^e_2)) - \hat{\Delta}((\pi_1^{\dagger},\pi^e_2),(\hat{\pi}_1^{\dagger},\pi^e_2)) + v_1(\hat{\pi}_1^{\dagger},\pi^e_2) - \hat{v}^{\mathrm{DR}}_1(\hat{\pi}_1^{\dagger},\pi^e_2) + \hat{\Delta}((\pi_1^{\dagger},\pi^e_2),(\hat{\pi}_1^{\dagger},\pi^e_2)) \\
    &- \Delta((\pi^e_1,\pi_2^{\dagger}),(\pi^e_1,\hat{\pi}_2^{\dagger})) + \hat{\Delta}((\pi^e_1,\pi_2^{\dagger}),(\pi^e_1,\hat{\pi}_2^{\dagger})) - v_1(\pi^e_1,\hat{\pi}_2^{\dagger}) + \hat{v}^{\mathrm{DR}}_1(\pi^e_1,\hat{\pi}_2^{\dagger}) - \hat{\Delta}((\pi^e_1,\pi_2^{\dagger}),(\pi^e_1,\hat{\pi}_2^{\dagger}))\\
    &\leq \Delta((\pi_1^{\dagger},\pi^e_2),(\hat{\pi}_1^{\dagger},\pi^e_2)) - \hat{\Delta}((\pi_1^{\dagger},\pi^e_2),(\hat{\pi}_1^{\dagger},\pi^e_2)) + v_1(\hat{\pi}_1^{\dagger},\pi^e_2) - \hat{v}^{\mathrm{DR}}_1(\hat{\pi}_1^{\dagger},\pi^e_2) \\
    &- \Delta((\pi^e_1,\pi_2^{\dagger}),(\pi^e_1,\hat{\pi}_2^{\dagger})) + \hat{\Delta}((\pi^e_1,\pi_2^{\dagger}),(\pi^e_1,\hat{\pi}_2^{\dagger})) - v_1(\pi^e_1,\hat{\pi}_2^{\dagger}) + \hat{v}^{\mathrm{DR}}_1(\pi^e_1,\hat{\pi}_2^{\dagger}) \\
    &\leq \Delta((\pi_1^{\dagger},\pi^e_2),(\hat{\pi}_1^{\dagger},\pi^e_2)) - \hat{\Delta}((\pi_1^{\dagger},\pi^e_2),(\hat{\pi}_1^{\dagger},\pi^e_2)) + \Delta((\hat{\pi}_1^{\dagger},\pi^e_2)),(\pi^e_1,\hat{\pi}_2^{\dagger})) - \Delta^{\mathrm{DR}}((\hat{\pi}_1^{\dagger},\pi^e_2)),(\pi^e_1,\hat{\pi}_2^{\dagger})) \\
    &- \Delta((\pi^e_1,\pi_2^{\dagger}),(\pi^e_1,\hat{\pi}_2^{\dagger})) + \hat{\Delta}((\pi^e_1,\pi_2^{\dagger}),(\pi^e_1,\hat{\pi}_2^{\dagger})) \\
    &\leq 3\sup_{\pi^{\alpha}\in \Pi,\pi^{\beta}\in \Pi}|\Delta((\pi^{\alpha}_1,\pi^{\alpha}_2),(\pi^{\beta}_1,\pi^{\beta}_2)) - \hat{\Delta}((\pi^{\alpha}_1,\pi^{\alpha}_2),(\pi^{\beta}_1,\pi^{\beta}_2))|,
\end{align*}
and
\begin{align*}
    & v_{\Pi}^{\mathrm{exp}}(\pi^e_1,\pi^e_2) - \hat{v}_{\mathrm{DR}}^{\mathrm{exp}}(\pi^e_1,\pi^e_2) = v_1(\pi_1^{\dagger},\pi^e_2) - \hat{v}_1^{\mathrm{DR}}(\hat{\pi}_1^{\dagger},\pi^e_2) + v_2(\pi^e_1,\pi_2^{\dagger})  - \hat{v}^{\mathrm{DR}}_2(\pi^e_1,\hat{\pi}_2^{\dagger}) \\
    &= -\Delta((\pi_1^{\dagger},\pi^e_2),(\hat{\pi}_1^{\dagger},\pi^e_2)) + \hat{\Delta}((\pi_1^{\dagger},\pi^e_2),(\hat{\pi}_1^{\dagger},\pi^e_2)) + v_1(\pi_1^{\dagger},\pi^e_2) - \hat{v}^{\mathrm{DR}}_1(\pi_1^{\dagger},\pi^e_2) + \Delta((\pi_1^{\dagger},\pi^e_2),(\hat{\pi}_1^{\dagger},\pi^e_2)) \\
    & +\Delta((\pi^e_1,\pi_2^{\dagger}),(\pi^e_1,\hat{\pi}_2^{\dagger})) - \hat{\Delta}((\pi^e_1,\pi_2^{\dagger}),(\pi^e_1,\hat{\pi}_2^{\dagger})) - v_1(\pi^e_1,\pi_2^{\dagger}) + \hat{v}^{\mathrm{DR}}_1(\pi^e_1,\pi_2^{\dagger}) - \Delta((\pi^e_1,\pi_2^{\dagger}),(\pi^e_1,\hat{\pi}_2^{\dagger}))\\
    &\geq -\Delta((\pi_1^{\dagger},\pi^e_2),(\hat{\pi}_1^{\dagger},\pi^e_2)) + \hat{\Delta}((\pi_1^{\dagger},\pi^e_2),(\hat{\pi}_1^{\dagger},\pi^e_2)) + v_1(\pi_1^{\dagger},\pi^e_2) - \hat{v}^{\mathrm{DR}}_1(\pi_1^{\dagger},\pi^e_2) \\
    & +\Delta((\pi^e_1,\pi_2^{\dagger}),(\pi^e_1,\hat{\pi}_2^{\dagger})) - \hat{\Delta}((\pi^e_1,\pi_2^{\dagger}),(\pi^e_1,\hat{\pi}_2^{\dagger})) - v_1(\pi^e_1,\pi_2^{\dagger}) + \hat{v}^{\mathrm{DR}}_1(\pi^e_1,\pi_2^{\dagger}) \\
    &\geq -\Delta((\pi_1^{\dagger},\pi^e_2),(\hat{\pi}_1^{\dagger},\pi^e_2)) + \hat{\Delta}((\pi_1^{\dagger},\pi^e_2),(\hat{\pi}_1^{\dagger},\pi^e_2)) + \Delta((\pi_1^{\dagger},\pi^e_2),(\pi^e_1,\pi_2^{\dagger})) - \hat{\Delta}((\pi_1^{\dagger},\pi^e_2),(\pi^e_1,\pi_2^{\dagger})) \\
    & +\Delta((\pi^e_1,\pi_2^{\dagger}),(\pi^e_1,\hat{\pi}_2^{\dagger})) - \hat{\Delta}((\pi^e_1,\pi_2^{\dagger}),(\pi^e_1,\hat{\pi}_2^{\dagger})) \\
    &\geq -3\sup_{\pi^{\alpha}\in \Pi,\pi^{\beta}\in \Pi}|\Delta((\pi^{\alpha}_1,\pi^{\alpha}_2),(\pi^{\beta}_1,\pi^{\beta}_2)) - \hat{\Delta}((\pi^{\alpha}_1,\pi^{\alpha}_2),(\pi^{\beta}_1,\pi^{\beta}_2))|.
\end{align*}
Therefore, we have:
\begin{align*}
    &|v_{\Pi}^{\mathrm{exp}}(\pi^e_1,\pi^e_2) - \hat{v}_{\mathrm{DR}}^{\mathrm{exp}}(\pi^e_1,\pi^e_2)| \leq 3\sup_{\pi^{\alpha}\in \Pi,\pi^{\beta}\in \Pi}|\Delta((\pi^{\alpha}_1,\pi^{\alpha}_2),(\pi^{\beta}_1,\pi^{\beta}_2)) - \hat{\Delta}((\pi^{\alpha}_1,\pi^{\alpha}_2),(\pi^{\beta}_1,\pi^{\beta}_2))|.
\end{align*}
Based on Lemma \ref{lem:hat_delta_dr}, for $\delta>0$, there exists $C>0$, $N_{\delta}>0$, such that with probability at least $1-2\delta$ and for all $n\geq N_{\delta}$:
\begin{align*}
    |v_{\Pi}^{\mathrm{exp}}(\pi^e_1,\pi^e_2) - \hat{v}_{\mathrm{DR}}^{\mathrm{exp}}(\pi^e_1,\pi^e_2)| \leq C\left(\left(\kappa(\Pi)+\sqrt{\log\frac{1}{\delta}}\right)\sqrt{\frac{\Upsilon^{\ast}_{\mathrm{DR}}}{n}}\right).
\end{align*}
\end{proof}

\subsection{Proof of Theorem \ref{thm:ope_drl}}
\begin{proof}
Let define
\begin{align*}
    \Delta(\pi^{\alpha},\pi^{\beta})&=v_1(\pi^{\alpha}_1,\pi^{\alpha}_2)-v_1(\pi^{\beta}_1,\pi^{\beta}_2), \\
    \hat{\Delta}^{\mathrm{DRL}}(\pi^{\alpha},\pi^{\beta})&=\hat{v}^{\mathrm{DRL}}_1(\pi^{\alpha}_1,\pi^{\alpha}_2)-\hat{v}^{\mathrm{DRL}}_1(\pi^{\beta}_1,\pi^{\beta}_2), \\
    \tilde{\Delta}^{\mathrm{DRL}}(\pi^{\alpha},\pi^{\beta})&=v_1^{\mathrm{DRL}}(\pi^{\alpha}_1,\pi^{\alpha}_2)-v_1^{\mathrm{DRL}}(\pi^{\beta}_1,\pi^{\beta}_2).
\end{align*}
As in the proof of Theorem \ref{thm:ope_dr}, we have:
\begin{align*}
    &|v_{\Pi}^{\mathrm{exp}}(\pi^e_1,\pi^e_2) - \hat{v}_{\mathrm{DRL}}^{\mathrm{exp}}(\pi^e_1,\pi^e_2)| \leq 3\sup_{\pi^{\alpha},\pi^{\beta}\in \Pi}|\Delta((\pi^{\alpha}_1,\pi^{\alpha}_2),(\pi^{\beta}_1,\pi^{\beta}_2)) - \hat{\Delta}^{\mathrm{DRL}}((\pi^{\alpha}_1,\pi^{\alpha}_2),(\pi^{\beta}_1,\pi^{\beta}_2))|.
\end{align*}
Here, we introduce the following lemma.
\begin{lemma}
\label{lem:tilde_delta_drl}
Assume Assumptions \ref{asp:overlap_and_reward}, \ref{asp:estimator_bound}, \ref{asp:covering_number}, \ref{aspp:dr_3}, and \ref{aspp:drl_1}.
Then, for any $\delta>0$, there exists $C>0, N_{\delta}>0$, such that with probability at least $1-2\delta$ and for all $n\geq N_{\delta}$:
\begin{align*}
    \sup_{\pi^{\alpha},\pi^{\beta}\in\Pi} \left|\hat{\Delta}^{\mathrm{DRL}}(\pi^{\alpha},\pi^{\beta}) - \Delta(\pi^{\alpha},\pi^{\beta})\right| \leq C\left(\kappa(\Pi) + \sqrt{\log (1/\delta)}\right)\sqrt{\Upsilon^{\ast}_{\mathrm{DRL}}/n}.
\end{align*}
\end{lemma}
The proof of this lemma is shown in Section \ref{sec:regret_bound_drl}.
Based on Lemma \ref{lem:tilde_delta_drl}, for $\delta>0$, there exists $C>0$, $N_{\delta}>0$, such that with probability at least $1-2\delta$ and for all $n\geq N_{\delta}$:
\begin{align*}
    |v_{\Pi}^{\mathrm{exp}}(\pi^e_1,\pi^e_2) - \hat{v}_{\mathrm{DRL}}^{\mathrm{exp}}(\pi^e_1,\pi^e_2)| \leq C\left(\left(\kappa(\Pi)+\sqrt{\log\frac{1}{\delta}}\right)\sqrt{\frac{\Upsilon^{\ast}_{\mathrm{DRL}}}{n}}\right).
\end{align*}
\end{proof}

\subsection{Proof of Theorem \ref{thm:opl_dr}}
\begin{proof}
We have:
\begin{align*}
    &v^{\mathrm{exp}}(\hat{\pi}^{\mathrm{DR}}_1,\hat{\pi}^{\mathrm{DR}}_2) - v^{\mathrm{exp}}(\pi_1^{\ast},\pi_2^{\ast}) = v_1(\mathcal{B}(\hat{\pi}^{\mathrm{DR}}_2),\hat{\pi}^{\mathrm{DR}}_2) + v_2(\hat{\pi}^{\mathrm{DR}}_1,\mathcal{B}(\hat{\pi}^{\mathrm{DR}}_1)) - v_1(\mathcal{B}(\pi_2^{\ast}), \pi_2^{\ast}) - v_2(\pi_1^{\ast},\mathcal{B}(\pi_1^{\ast})) \\
    &= \Delta((\mathcal{B}(\hat{\pi}^{\mathrm{DR}}_2),\hat{\pi}^{\mathrm{DR}}_2), (\hat{\pi}^{\mathrm{DR}}_1,\mathcal{B}(\hat{\pi}^{\mathrm{DR}}_1))) - \Delta((\mathcal{B}(\pi_2^{\ast}), \pi_2^{\ast}), (\pi_1^{\ast},\mathcal{B}(\pi_1^{\ast}))) \\
    &= \Delta((\mathcal{B}(\hat{\pi}^{\mathrm{DR}}_2),\hat{\pi}^{\mathrm{DR}}_2), (\hat{\pi}^{\mathrm{DR}}_1,\mathcal{B}(\hat{\pi}^{\mathrm{DR}}_1))) - \hat{\Delta}((\mathcal{B}(\hat{\pi}^{\mathrm{DR}}_2),\hat{\pi}^{\mathrm{DR}}_2), (\hat{\pi}^{\mathrm{DR}}_1,\mathcal{B}(\hat{\pi}^{\mathrm{DR}}_1))) \\
    &- \Delta((\mathcal{B}(\pi_2^{\ast}), \pi_2^{\ast}), (\pi_1^{\ast},\mathcal{B}(\pi_1^{\ast}))) + \hat{\Delta}((\mathcal{B}(\pi_2^{\ast}), \pi_2^{\ast}), (\pi_1^{\ast},\mathcal{B}(\pi_1^{\ast}))) \\
    &+ \hat{v}^{\mathrm{DR}}_1(\mathcal{B}(\hat{\pi}^{\mathrm{DR}}_2),\hat{\pi}^{\mathrm{DR}}_2) - \hat{v}^{\mathrm{DR}}_1(\hat{\pi}^{\mathrm{DR}}_1,\mathcal{B}(\hat{\pi}^{\mathrm{DR}}_1)) - \hat{v}^{\mathrm{DR}}_1(\mathcal{B}(\pi_2^{\ast}), \pi_2^{\ast}) + \hat{v}^{\mathrm{DR}}_1(\pi_1^{\ast},\mathcal{B}(\pi_1^{\ast})),
\end{align*}
where $\mathcal{B}(\hat{\pi}^{\mathrm{DR}}_1)=\argmax_{\pi_2\in \Omega_2} v_2(\hat{\pi}^{\mathrm{DR}}_1,\pi^e_2)$ and $\mathcal{B}(\hat{\pi}^{\mathrm{DR}}_2)=\argmax_{\pi_1\in \Omega_1} v_1(\pi_1, \hat{\pi}^{\mathrm{DR}}_2)$.
Let define $\hat{\mathcal{B}}(\pi_1)=\argmax_{\pi_2\in \Pi_2} \hat{v}^{\mathrm{DR}}_2(\pi_1,\pi_2)$ and $\hat{\mathcal{B}}(\pi_2)=\argmax_{\pi_1\in \Pi_1} \hat{v}^{\mathrm{DR}}_1(\pi_1, \pi_2)$.
Then, we have:
\begin{align*}
    &\hat{v}^{\mathrm{DR}}_1(\mathcal{B}(\hat{\pi}^{\mathrm{DR}}_2),\hat{\pi}^{\mathrm{DR}}_2) \leq \hat{v}^{\mathrm{DR}}_1(\hat{\mathcal{B}}(\hat{\pi}^{\mathrm{DR}}_2),\hat{\pi}^{\mathrm{DR}}_2) \leq \hat{v}^{\mathrm{DR}}_1(\hat{\mathcal{B}}(\pi_2^{\ast}),\pi_2^{\ast}) \\
    &\hat{v}^{\mathrm{DR}}_1(\hat{\pi}^{\mathrm{DR}}_1, \mathcal{B}(\hat{\pi}^{\mathrm{DR}}_1)) \geq \hat{v}^{\mathrm{DR}}_1(\hat{\pi}^{\mathrm{DR}}_1, \hat{\mathcal{B}}(\hat{\pi}^{\mathrm{DR}}_1)) \geq \hat{v}^{\mathrm{DR}}_1(\pi_1^{\ast},\hat{\mathcal{B}}(\pi_1^{\ast}))
\end{align*}
Therefore, we have:
\begin{align*}
    &v^{\mathrm{exp}}(\hat{\pi}^{\mathrm{DR}}_1,\hat{\pi}^{\mathrm{DR}}_2) - v^{\mathrm{exp}}(\pi_1^{\ast},\pi_2^{\ast}) \\
    &\leq \Delta((\mathcal{B}(\hat{\pi}^{\mathrm{DR}}_2),\hat{\pi}^{\mathrm{DR}}_2), (\hat{\pi}^{\mathrm{DR}}_1,\mathcal{B}(\hat{\pi}^{\mathrm{DR}}_1))) - \hat{\Delta}((\mathcal{B}(\hat{\pi}^{\mathrm{DR}}_2),\hat{\pi}^{\mathrm{DR}}_2), (\hat{\pi}^{\mathrm{DR}}_1,\mathcal{B}(\hat{\pi}^{\mathrm{DR}}_1))) \\
    &- \Delta((\mathcal{B}(\pi_2^{\ast}), \pi_2^{\ast}), (\pi_1^{\ast},\mathcal{B}(\pi_1^{\ast}))) + \hat{\Delta}((\mathcal{B}(\pi_2^{\ast}), \pi_2^{\ast}), (\pi_1^{\ast},\mathcal{B}(\pi_1^{\ast}))) \\
    &+ \hat{v}^{\mathrm{DR}}_1(\hat{\mathcal{B}}(\pi_2^{\ast}),\pi_2^{\ast}) - \hat{v}^{\mathrm{DR}}_1(\mathcal{B}(\pi_2^{\ast}), \pi_2^{\ast}) - \hat{v}^{\mathrm{DR}}_1(\pi_1^{\ast},\hat{\mathcal{B}}(\pi_1^{\ast})) + \hat{v}^{\mathrm{DR}}_1(\pi_1^{\ast},\mathcal{B}(\pi_1^{\ast})), \\
    &\leq \Delta((\mathcal{B}(\hat{\pi}^{\mathrm{DR}}_2),\hat{\pi}^{\mathrm{DR}}_2), (\hat{\pi}^{\mathrm{DR}}_1,\mathcal{B}(\hat{\pi}^{\mathrm{DR}}_1))) - \hat{\Delta}((\mathcal{B}(\hat{\pi}^{\mathrm{DR}}_2),\hat{\pi}^{\mathrm{DR}}_2), (\hat{\pi}^{\mathrm{DR}}_1,\mathcal{B}(\hat{\pi}^{\mathrm{DR}}_1))) \\
    &- \Delta((\mathcal{B}(\pi_2^{\ast}), \pi_2^{\ast}), (\pi_1^{\ast},\mathcal{B}(\pi_1^{\ast}))) + \hat{\Delta}((\mathcal{B}(\pi_2^{\ast}), \pi_2^{\ast}), (\pi_1^{\ast},\mathcal{B}(\pi_1^{\ast}))) \\
    &+ \hat{v}^{\mathrm{DR}}_1(\hat{\mathcal{B}}(\pi_2^{\ast}),\pi_2^{\ast}) - \hat{v}^{\mathrm{DR}}_1(\mathcal{B}(\pi_2^{\ast}), \pi_2^{\ast}) - v_1(\hat{\mathcal{B}}(\pi_2^{\ast}),\pi_2^{\ast}) + v_1(\mathcal{B}(\pi_2^{\ast}), \pi_2^{\ast}) \\
    &- \hat{v}^{\mathrm{DR}}_1(\pi_1^{\ast},\hat{\mathcal{B}}(\pi_1^{\ast})) + \hat{v}^{\mathrm{DR}}_1(\pi_1^{\ast},\mathcal{B}(\pi_1^{\ast}))  + v_1(\pi_1^{\ast},\hat{\mathcal{B}}(\pi_1^{\ast})) - v_1(\pi_1^{\ast},\mathcal{B}(\pi_1^{\ast})) \\
    &= \Delta((\mathcal{B}(\hat{\pi}^{\mathrm{DR}}_2),\hat{\pi}^{\mathrm{DR}}_2), (\hat{\pi}^{\mathrm{DR}}_1,\mathcal{B}(\hat{\pi}^{\mathrm{DR}}_1))) - \hat{\Delta}((\mathcal{B}(\hat{\pi}^{\mathrm{DR}}_2),\hat{\pi}^{\mathrm{DR}}_2), (\hat{\pi}^{\mathrm{DR}}_1,\mathcal{B}(\hat{\pi}^{\mathrm{DR}}_1))) \\
    &- \Delta((\mathcal{B}(\pi_2^{\ast}), \pi_2^{\ast}), (\pi_1^{\ast},\mathcal{B}(\pi_1^{\ast}))) + \hat{\Delta}((\mathcal{B}(\pi_2^{\ast}), \pi_2^{\ast}), (\pi_1^{\ast},\mathcal{B}(\pi_1^{\ast}))) \\
    &+ \hat{\Delta}((\hat{\mathcal{B}}(\pi_2^{\ast}),\pi_2^{\ast}),(\mathcal{B}(\pi_2^{\ast}), \pi_2^{\ast})) - \Delta((\hat{\mathcal{B}}(\pi_2^{\ast}),\pi_2^{\ast}),(\mathcal{B}(\pi_2^{\ast}), \pi_2^{\ast})) \\
    &- \hat{\Delta}((\pi_1^{\ast},\hat{\mathcal{B}}(\pi_1^{\ast})),(\pi_1^{\ast},\mathcal{B}(\pi_1^{\ast})))  + \Delta((\pi_1^{\ast},\hat{\mathcal{B}}(\pi_1^{\ast})),(\pi_1^{\ast},\mathcal{B}(\pi_1^{\ast}))) \\
    &\leq 4\sup_{\pi^{\alpha}\in \Pi,\pi^{\beta}\in \Pi}|\Delta((\pi^{\alpha}_1,\pi^{\alpha}_2),(\pi^{\beta}_1,\pi^{\beta}_2)) - \hat{\Delta}((\pi^{\alpha}_1,\pi^{\alpha}_2),(\pi^{\beta}_1,\pi^{\beta}_2))|,
\end{align*}
Therefore, based on Lemma \ref{lem:hat_delta_dr}, for $\delta>0$, there exists $C>0$, $N_{\delta}>0$, such that with probability at least $1-2\delta$ and for all $n\geq N_{\delta}$:
\begin{align*}
    v^{\mathrm{exp}}(\hat{\pi}^{\mathrm{DR}}_1,\hat{\pi}^{\mathrm{DR}}_2) - v^{\mathrm{exp}}(\pi_1^{\ast},\pi_2^{\ast}) \leq C\left(\left(\kappa(\Pi)+\sqrt{\log\frac{1}{\delta}}\right)\sqrt{\frac{\Upsilon^{\ast}_{\mathrm{DR}}}{n}}\right).
\end{align*}
\end{proof}

\subsection{Proof of Theorem \ref{thm:opl_drl}}
\begin{proof}
As in the proof of Theorem \ref{thm:opl_dr}, we have:
\begin{align*}
    &v^{\mathrm{exp}}(\hat{\pi}^{\mathrm{DRL}}_1,\hat{\pi}^{\mathrm{DRL}}_2) - v^{\mathrm{exp}}(\pi_1^{\ast},\pi_2^{\ast}) \leq 4\sup_{\pi^{\alpha}\in \Pi,\pi^{\beta}\in \Pi}|\Delta((\pi^{\alpha}_1,\pi^{\alpha}_2),(\pi^{\beta}_1,\pi^{\beta}_2)) - \hat{\Delta}^{\mathrm{DRL}}((\pi^{\alpha}_1,\pi^{\alpha}_2),(\pi^{\beta}_1,\pi^{\beta}_2))|,
\end{align*}
Therefore, based on Lemma \ref{lem:tilde_delta_drl}, for $\delta>0$, there exists $C>0$, $N_{\delta}>0$, such that with probability at least $1-2\delta$ and for all $n\geq N_{\delta}$:
\begin{align*}
    v^{\mathrm{exp}}(\hat{\pi}^{\mathrm{DRL}}_1,\hat{\pi}^{\mathrm{DRL}}_2) - v^{\mathrm{exp}}(\pi_1^{\ast},\pi_2^{\ast}) \leq C\left(\left(\kappa(\Pi)+\sqrt{\log\frac{1}{\delta}}\right)\sqrt{\frac{\Upsilon^{\ast}_{\mathrm{DRL}}}{n}}\right).
\end{align*}
\end{proof}

\section{Proofs of Lemmas}
\subsection{Proof of Lemma \ref{lem:tilde_delta_dr}}
\label{sec:appendix_proof_tilde_delta_dr}
\begin{proof}
The proof divides into two main components.

We can rewrite $v^{\mathrm{DR}}_1(\pi_1^e,\pi_2^e)$ as
\begin{equation}
\label{eq:qqq}
\begin{aligned}
    &v^{\mathrm{DR}}_1(\pi_1^e,\pi_2^e) = \frac{1}{n}\sum_{i=1}^n\left(\frac{\prod_{t^{\prime}=1}^t\pi^e_t(a_{i,t^{\prime}}|s_{i,t^{\prime}})}{\prod_{t^{\prime}=1}^t\pi^b_t(a_{i,t^{\prime}}|s_{i,t^{\prime}})} R_{i,t} \right.\\
    &- \left.\frac{\prod_{t^{\prime}=1}^t\pi^e_t(a_{i,t^{\prime}}|s_{i,t^{\prime}})}{\prod_{t^{\prime}=1}^t\pi^b_t(a_{i,t^{\prime}}|s_{i,t^{\prime}})} \sum_{s_{t+1}\in \mathcal{S}}P_T(s_{t+1}|s_{i,t},a_{i,t})\sum_{t^{\prime}=t+1}^T\gamma^{t^{\prime}-t}\sum_{\tau_{t+1:t^{\prime}}}\left(\prod_{l=t+1}^{t^{\prime}}\pi^e_l(a_{l}|s_{l})\right)\left(R_{t^{\prime}}\prod_{l=t+1}^{t^{\prime}-1}P_T(s_{l+1}|s_{l},a_{l})\right) \right.\\
    &+ \left.\frac{\prod_{t^{\prime}=1}^{t-1}\pi^e_t(a_{t^{\prime}}|s_{t^{\prime}})}{\prod_{t^{\prime}=1}^{t-1}\pi^b_t(a_{t^{\prime}}|s_{t^{\prime}})}\sum_{a\in\mathcal{A}}\pi^e_t(a|s_t)\sum_{s_{t+1}\in \mathcal{S}}P_T(s_{t+1}|s_{i,t},a)\sum_{t^{\prime}=t+1}^T\gamma^{t^{\prime}-t}\sum_{\tau_{t+1:t^{\prime}}}\!\left(\!\prod_{l=t+1}^{t^{\prime}}\pi^e_l(a_{l}|s_{l})\!\right)\!\!\left(\!R_{t^{\prime}}\prod_{l=t+1}^{t^{\prime}-1}P_T(s_{l+1}|s_{l},a_{l})\!\right)\!\!\right)\!,
\end{aligned}
\end{equation}
where $\tau_{t:t^{\prime}}=(s_t,a_t^1,a_t^2,\cdots,s_{t^{\prime}},a_{t^{\prime}}^1,a_{t^{\prime}}^2)$.
Therefore, 
we can write
\begin{align*}
    v^{\mathrm{DR}}_1(\pi_1^e,\pi_2^e) = \frac{1}{n}\sum_{i=1}^n\sum_{s_{1:T}} \langle \pi^e(s_{1:T}), \Gamma_{i,s_{1:T}} \rangle
\end{align*}
where $s_{1:T}=(s_1,\cdots,s_T)$ and $\Gamma_{i,s_{1:T}}$ is a random variable that is independent of $\pi^e$.
By using this form, we can write $v^{\mathrm{DR}}_1(\pi_1^{\alpha},\pi_2^{\alpha})-v^{\mathrm{DR}}_1(\pi_1^{\beta},\pi_2^{\beta})$ as 
\begin{align*}
    v^{\mathrm{DR}}_1(\pi_1^{\alpha},\pi_2^{\alpha})-v^{\mathrm{DR}}_1(\pi_1^{\beta},\pi_2^{\beta}) = \frac{1}{n}\sum_{i=1}^n\sum_{s_{1:T}} \langle \pi^{\alpha}(s_{1:T})-\pi^{\beta}(s_{1:T}), \Gamma_{i,s_{1:T}} \rangle,
\end{align*}
because $\Gamma_{i,s_{1:T}}$ is independent of $\pi^{\alpha}$ and $\pi^{\beta}$.

Hereafter, we prove the statement following \cite{zhou2018offline}.
We extend the proofs of \cite{zhou2018offline} to TZMG cases.

\paragraph{Step 1: Bounding Rademacher complexity.}

First, we bound the Rademacher complexity.
We introduce the following definitions of the Rademacher complexity.
\begin{definition}
Let $\Pi^D=\{\sum_{s_{1:T}} \langle \pi^{\alpha}(\cdot)-\pi^{\beta}(\cdot),\cdot\rangle\}$ and $Z_i$'s be {\bf iid} Rademacher random variables: $P(Z_i=1)=P(Z_i=-1)=\frac{1}{2}$.
\begin{enumerate}
    \item The empirical Rademacher complexity $\mathcal{R}_n(\Pi^D)$ of the functon class $\Pi^D$ is defined as:
    \begin{align*}
        &\mathcal{R}_n(\Pi^D; \{\{s_{i,t}\}, \{\Gamma_{i,s_{1:T}}\}\}_{i=1}^n) \\
        &= \mathbb{E}\left[\sup_{\pi^{\alpha},\pi^{\beta}\in \Pi}\frac{1}{n} |\sum_{i=1}^n Z_i\sum_{s_{1:T}} \langle \pi^{\alpha}(s_{1:T})-\pi^{\beta}(s_{1:T}), \Gamma_{i,s_{1:T}} \rangle| \middle| \{\{s_{i,t}\}, \{\Gamma_{i,s_{1:T}}\}\}_{i=1}^n \right],
    \end{align*}
    where the expectation is taken with respect to $Z_1, \cdots, Z_n$.
    \item The Rademacher complexity $\mathcal{R}_n(\Pi^D)$ of the function class $\Pi^D$ is the expected value (taken with respect to the sample $\{\{s_{i,t}\}, \{\Gamma_{i,s_{1:T}}\}\}_{i=1}^n$) of the empirical Rademacher complexity: $\mathcal{R}_n(\Pi^D) = \mathbb{E}[\mathcal{R}_n(\Pi^D; \{\{s_{i,t}\}, \{\Gamma_{i,s_{1:T}}\}\}_{i=1}^n)]$.
\end{enumerate}
\end{definition}
Using these definitions, we can derive the following Lemma.
\begin{lemma}
\label{lem:rademacher}
Let $\{\{\Gamma_{i,t}\}\}_{i=1}^n$ be {\bf iid} set of weights with bounded support.
Then under Assumption \ref{asp:overlap_and_reward} and \ref{asp:covering_number}:
\begin{align}
    \mathcal{R}_n(\Pi^D) = O\left(\kappa(\Pi)\sqrt{\frac{\sup\limits_{\pi^{\alpha},\pi^{\beta}\in \Pi}\mathbb{E}\left[\left(\sum_{s_{1:T}} \langle \pi^{\alpha}(s_{1:T})-\pi^{\beta}(s_{1:T}), \Gamma_{i,s_{1:T}} \rangle\right)^2\right]}{n}}\right) + o(\frac{1}{\sqrt{n}}).
\end{align}
\end{lemma}

\paragraph{Step 2: Expected uniform bound on maximum deviation.}

Since $v_1^{\mathrm{DR}}(\pi_1,\pi_2)$ is consistent, classical results on Rademacher complexity \cite{bartlett2002rademacher} give:
\begin{align*}
    \mathbb{E}\left[\sup_{\pi^{\alpha}, \pi^{\beta} \in \Pi}\left|\tilde{\Delta}(\pi^{\alpha},\pi^{\beta}) - \Delta(\pi^{\alpha}, \pi^{\beta})\right|\right] \leq 2 \mathcal{R}_n(\Pi^D).
\end{align*}
Therefore, from Lemma \ref{lem:rademacher}, we have:
\begin{equation}
\begin{aligned}
\label{eq:tilde_delta_o}
    \mathbb{E}\left[\sup_{\pi^{\alpha}, \pi^{\beta} \in \Pi}\left|\tilde{\Delta}(\pi^{\alpha}, \pi^{\beta}) - \Delta(\pi^{\alpha}, \pi^{\beta})\right|\right] &\leq O\left(\kappa(\Pi)\sqrt{\frac{\sup_{\pi^{\alpha}, \pi^{\beta}}\mathbb{E}\left[\left(\mathcal{M}(\pi^{\alpha},\pi^{\beta},\{s_{i,t}\}, \{\Gamma_{i,s_{1:T}}\})\right)^2\right]}{n}}\right) + o(\frac{1}{\sqrt{n}}) \\
    &\leq 4\cdot O\left(\kappa(\Pi)\sqrt{\frac{\Upsilon^{\ast}_{\mathrm{DR}}}{n}}\right) + o(\frac{1}{\sqrt{n}})
\end{aligned}
\end{equation}

\paragraph{Step 3: High probability bound on maximum deviation via Talagrand inequality.}

From the previous step, it remains to bound the difference between $\sup_{\pi^{\alpha}, \pi^{\beta} \in \Pi}\left|\tilde{\Delta}(\pi^{\alpha}, \pi^{\beta}) - \Delta(\pi^{\alpha}, \pi^{\beta})\right|$ and $\mathbb{E}\left[\sup_{\pi^{\alpha}, \pi^{\beta} \in \Pi}\left|\tilde{\Delta}(\pi^{\alpha}, \pi^{\beta}) - \Delta(\pi^{\alpha}, \pi^{\beta})\right|\right]$.
Here, we introduce the following version of Talagrand's concentration inequality in \cite{gine2006concentration, zhou2018offline}:
\begin{lemma}
\label{lem:talagrand_2}
Let $X_1, \cdots, X_n$ be independent $\mathcal{X}$-valued random variables and $\mathcal{F}$ be a class of functions where each $f:\mathcal{X}\to \mathbb{R}$ in $\mathcal{F}$ satisfies $\sup_{x\in \mathcal{X}}|f(x)|\leq 1$.
Then:
\begin{align*}
    P\left(\left|\sup_{f\in \mathcal{F}} |\sum_{i=1}^n f(X_i)| - \mathbb{E}\left[\sup_{f\in \mathcal{F}} |\sum_{i=1}^n f(X_i)|\right]\right| \geq t\right) \leq 2\exp\left(-\frac{1}{2}t\log(1+\frac{t}{V})\right), \forall > 0,
\end{align*}
where $V$ is any number satisfying $V\geq \mathbb{E}\left[\sup_{f\in \mathcal{F}} \sum_{i=1}^n f^2(X_i)\right]$.
\end{lemma}

We apply Lemma \ref{lem:talagrand_2} to the current context: we identify $X_i$ in Lemma \ref{lem:talagrand_2} with $(\{s_{i,t}\}, \{\Gamma_{i,s_{1:T}}\})$ here and $f(\{s_{i,t}\}, \{\Gamma_{i,s_{1:T}}\})=\frac{\mathcal{M}(\pi_1,\pi_2,\{s_{i,t}\}, \{\Gamma_{i,s_{1:T}}\}) - \mathbb{E}[\mathcal{M}(\pi_1,\pi_2,\{s_{i,t}\}, \{\Gamma_{i,s_{1:T}}\})]}{2U}$, where $U$ satisfies $|\mathcal{M}(\pi_1,\pi_2,\{s_{i,t}\}, \{\Gamma_{i,s_{1:T}}\})|\leq U$.
Consequently, we have:
\begin{align*}
    &P\left(\left|\sup_{\pi_1,\pi_2\in\Pi}\left|\sum_{i=1}^n\frac{\mathcal{M}(\pi_1,\pi_2,\{s_{i,t}\}, \{\Gamma_{i,s_{1:T}}\}) - \mathbb{E}[\mathcal{M}(\pi_1,\pi_2,\{s_{i,t}\}, \{\Gamma_{i,s_{1:T}}\})]}{2U}\right| \right.\right. \\
    &\left.\left. - \mathbb{E}\left[\sup_{\pi_1,\pi_2\in\Pi}\left|\sum_{i=1}^n\frac{\mathcal{M}(\pi_1,\pi_2,\{s_{i,t}\}, \{\Gamma_{i,s_{1:T}}\}) - \mathbb{E}[\mathcal{M}(\pi_1,\pi_2,\{s_{i,t}\}, \{\Gamma_{i,s_{1:T}}\})]}{2U}\right|\right] \right| \geq t\right) \\
    &= P\left(\left|\sup_{\pi_1,\pi_2\in\Pi}\frac{n}{2U}\left|\tilde{\Delta}(\pi_1,\pi_2)-\Delta(\pi_1,\pi_2)\right| - \mathbb{E}\left[\sup_{\pi_1,\pi_2\in\Pi}\frac{n}{2U}\left|\tilde{\Delta}(\pi_1,\pi_2)-\Delta(\pi_1,\pi_2)\right|\right] \right| \geq t\right) \\
    &\leq 2\exp\left(-\frac{1}{2}t\log(1+\frac{t}{V})\right).
\end{align*}
Here, let $t=2\sqrt{2\left(\log \frac{1}{\delta}\right)V} + 2\log \frac{1}{\delta}$, we have:
\begin{align*}
    &\exp\left(-\frac{1}{2}t\log(1+\frac{t}{V})\right) = \exp\left(-\frac{2\sqrt{2\left(\log \frac{1}{\delta}\right)V} + 2\log \frac{1}{\delta}}{2}\log(1+\frac{2\sqrt{2\left(\log \frac{1}{\delta}\right)V} + 2\log \frac{1}{\delta}}{V})\right) \\
    &\leq \exp\left(-\frac{2\sqrt{2\left(\log \frac{1}{\delta}\right)V} + 2\log \frac{1}{\delta}}{2}\frac{\frac{2\sqrt{2\left(\log \frac{1}{\delta}\right)V} + 2\log \frac{1}{\delta}}{V}}{1+\frac{2\sqrt{2\left(\log \frac{1}{\delta}\right)V} + 2\log \frac{1}{\delta}}{V})}\right) = \exp\left(-\frac{1}{2}\frac{(2\sqrt{2\left(\log \frac{1}{\delta}\right)V} + 2\log \frac{1}{\delta})^2}{V + 2\sqrt{2\left(\log \frac{1}{\delta}\right)V} + 2\log \frac{1}{\delta}}\right) \\
    &= \exp\left(-\frac{1}{2}\left(\frac{2\sqrt{2\left(\log \frac{1}{\delta}\right)V} + 2\log \frac{1}{\delta}}{\sqrt{V} + \sqrt{2\log\frac{1}{\delta}}}\right)^2\right) \leq \exp\left(-\frac{1}{2}\left(\sqrt{2\log\frac{1}{\delta}}\right)^2\right) = \exp\left(-\log\frac{1}{\delta}\right) = \delta
\end{align*}
Therefore,
\begin{align*}
    &P\left(\left|\sup_{\pi_1,\pi_2\in\Pi}\frac{n}{2U}\left|\tilde{\Delta}(\pi_1,\pi_2)-\Delta(\pi_1,\pi_2)\right| - \mathbb{E}\left[\sup_{\pi_1,\pi_2\in\Pi}\frac{n}{2U}\left|\tilde{\Delta}(\pi_1,\pi_2)-\Delta(\pi_1,\pi_2)\right|\right] \right| \geq 2\sqrt{2\left(\log \frac{1}{\delta}\right)V} + 2\log \frac{1}{\delta}\right) \\
    &\leq 2\exp\left(-\frac{1}{2}t\log(1+\frac{t}{V})\right) \leq 2\delta.
\end{align*}
This means that with probability at least $1-2\delta$:
\begin{align*}
    \sup_{\pi_1,\pi_2\in\Pi}\frac{n}{2U}\left|\tilde{\Delta}(\pi_1,\pi_2)-\Delta(\pi_1,\pi_2)\right| \leq \mathbb{E}\left[\sup_{\pi_1,\pi_2\in\Pi}\frac{n}{2U}\left|\tilde{\Delta}(\pi_1,\pi_2)-\Delta(\pi_1,\pi_2)\right|\right] + 2\sqrt{2\left(\log \frac{1}{\delta}\right)V} + 2\log \frac{1}{\delta}.
\end{align*}
Now multiplying both sides by $2U$ and dividing both sides by $n$:
\begin{align}
\label{eq:tilde_delta}
    \sup_{\pi_1,\pi_2\in\Pi}\left|\tilde{\Delta}(\pi_1,\pi_2)-\Delta(\pi_1,\pi_2)\right| \leq \mathbb{E}\left[\sup_{\pi_1,\pi_2\in\Pi}\left|\tilde{\Delta}(\pi_1,\pi_2)-\Delta(\pi_1,\pi_2)\right|\right] + \frac{4}{n}\sqrt{2U^2\left(\log \frac{1}{\delta}\right)V} + \frac{2U}{n}\log \frac{1}{\delta}.
\end{align}
Here, from Lemma \ref{lem:talagrand}, we have:
\begin{align*}
    &\mathbb{E}\left[\sup_{\pi_1, \pi_2}\sum_{i=1}^n\left(\mathcal{M}(\pi_1,\pi_2,\{s_{i,t}\}, \{\Gamma_{i,s_{1:T}}\}) - \mathbb{E}[\mathcal{M}(\pi_1,\pi_2,\{s_{i,t}\}, \{\Gamma_{i,s_{1:T}}\})]\right)^2\right] \\
    &\leq n\sup_{\pi_1,\pi_2\in \Pi}\mathbb{V}\left[\left(\mathcal{M}(\pi_1,\pi_2,\{s_{i,t}\}, \{\Gamma_{i,s_{1:T}}\})\right)\right] \\
    &+ 8U\mathbb{E}\left[\sup_{\pi^{\alpha},\pi^{\beta}}\left|\sum_{i=1}^n Z_i\left(\mathcal{M}(\pi_1,\pi_2,\{s_{i,t}\}, \{\Gamma_{i,s_{1:T}}\}) - \mathbb{E}\left[\mathcal{M}(\pi_1,\pi_2,\{s_{i,t}\}, \{\Gamma_{i,s_{1:T}}\})\right]\right)\right|\right] \\
    &\leq n\sup_{\pi_1,\pi_2\in \Pi}\mathbb{E}\left[\left(\mathcal{M}(\pi_1,\pi_2,\{s_{i,t}\}, \{\Gamma_{i,s_{1:T}}\})\right)^2\right] \\
    &+ 8U\mathbb{E}\left[\sup_{\pi^{\alpha},\pi^{\beta}}\left|\sum_{i=1}^n Z_i\left(\mathcal{M}(\pi_1,\pi_2,\{s_{i,t}\}, \{\Gamma_{i,s_{1:T}}\})\right)\right|\right] + 8U\mathbb{E}\left[\sup_{\pi^{\alpha},\pi^{\beta}}\left|\sum_{i=1}^n Z_i\left(\mathbb{E}\left[\mathcal{M}(\pi_1,\pi_2,\{s_{i,t}\}, \{\Gamma_{i,s_{1:T}}\})\right]\right)\right|\right] \\
    &\leq n\sup_{\pi_1,\pi_2\in \Pi}\mathbb{E}\left[\left(\mathcal{M}(\pi_1,\pi_2,\{s_{i,t}\}, \{\Gamma_{i,s_{1:T}}\})\right)^2\right] \\
    &+ 8U\mathbb{E}\left[\sup_{\pi^{\alpha},\pi^{\beta}}\left|\sum_{i=1}^n Z_i\left(\mathcal{M}(\pi_1,\pi_2,\{s_{i,t}\}, \{\Gamma_{i,s_{1:T}}\})\right)\right|\right] + 8U\mathbb{E}\left[\sup_{\pi^{\alpha},\pi^{\beta}}\left|\sum_{i=1}^n Z_i\left(\mathcal{M}(\pi_1,\pi_2,\{s_{i,t}\}, \{\Gamma_{i,s_{1:T}}\})\right)\right|\right] \\
    &= n\sup_{\pi_1,\pi_2\in \Pi}\mathbb{E}\left[\left(\mathcal{M}(\pi_1,\pi_2,\{s_{i,t}\}, \{\Gamma_{i,s_{1:T}}\})\right)^2\right] + 16U\mathbb{E}\left[\sup_{\pi^{\alpha},\pi^{\beta}}\left|\sum_{i=1}^n Z_i\left(\mathcal{M}(\pi_1,\pi_2,\{s_{i,t}\}, \{\Gamma_{i,s_{1:T}}\})\right)\right|\right],
\end{align*}
where the last inequality follows from Jensen by noting that:
\begin{align*}
    &\mathbb{E}\left[\sup_{\pi^{\alpha},\pi^{\beta}}\left|\sum_{i=1}^n Z_i\left(\mathbb{E}\left[\mathcal{M}(\pi_1,\pi_2,\{s_{i,t}\}, \{\Gamma_{i,s_{1:T}}\})\right]\right)\right|\right] \leq \mathbb{E}\left[\sup_{\pi^{\alpha},\pi^{\beta}}\mathbb{E}\left|\sum_{i=1}^n Z_i\left(\mathcal{M}(\pi_1,\pi_2,\{s_{i,t}\}, \{\Gamma_{i,s_{1:T}}\})\right)\right|\right] \\
    &\leq \mathbb{E}\left[\sup_{\pi^{\alpha},\pi^{\beta}}\left|\sum_{i=1}^n Z_i\left(\mathcal{M}(\pi_1,\pi_2,\{s_{i,t}\}, \{\Gamma_{i,s_{1:T}}\})\right)\right|\right].
\end{align*}
Consequently, we have:
\begin{align*}
    &\mathbb{E}\left[\sup_{\pi_1, \pi_2}\sum_{i=1}^n\left(\frac{\mathcal{M}(\pi_1,\pi_2,\{s_{i,t}\}, \{\Gamma_{i,s_{1:T}}\}) - \mathbb{E}[\mathcal{M}(\pi_1,\pi_2,\{s_{i,t}\}, \{\Gamma_{i,s_{1:T}}\})]}{2U}\right)^2\right] \\
    &\leq \frac{n}{4U^2}\sup_{\pi_1,\pi_2\in \Pi}\mathbb{E}\left[\left(\mathcal{M}(\pi_1,\pi_2,\{s_{i,t}\}, \{\Gamma_{i,s_{1:T}}\})\right)^2\right] + \frac{8}{U}\mathbb{E}\left[\sup_{\pi^{\alpha},\pi^{\beta}}\left|\sum_{i=1}^n Z_i\left(\mathcal{M}(\pi_1,\pi_2,\{s_{i,t}\}, \{\Gamma_{i,s_{1:T}}\})\right)\right|\right]
\end{align*}
Therefore, we can plug the following $V$ value into Equation (\ref{eq:tilde_delta}):
\begin{align*}
    V = \frac{n}{4U^2}\sup_{\pi_1,\pi_2\in \Pi}\mathbb{E}\left[\left(\mathcal{M}(\pi_1,\pi_2,\{s_{i,t}\}, \{\Gamma_{i,s_{1:T}}\})\right)^2\right] + \frac{8}{U}\mathbb{E}\left[\sup_{\pi^{\alpha},\pi^{\beta}}\left|\sum_{i=1}^n Z_i\left(\mathcal{M}(\pi_1,\pi_2,\{s_{i,t}\}, \{\Gamma_{i,s_{1:T}}\})\right)\right|\right],
\end{align*}
it follows that with probability at least $1-2\delta$:
\begin{align*}
    &\sup_{\pi_1,\pi_2\in\Pi}\left|\tilde{\Delta}(\pi_1,\pi_2)-\Delta(\pi_1,\pi_2)\right| \leq \mathbb{E}\left[\sup_{\pi_1,\pi_2\in\Pi}\left|\tilde{\Delta}(\pi_1,\pi_2)-\Delta(\pi_1,\pi_2)\right|\right] + \frac{4}{n}\sqrt{2U^2\left(\log \frac{1}{\delta}\right)V} + \frac{2U}{n}\log \frac{1}{\delta} \\
    &\leq \mathbb{E}\left[\sup_{\pi_1,\pi_2\in\Pi}\left|\tilde{\Delta}(\pi_1,\pi_2)-\Delta(\pi_1,\pi_2)\right|\right] + \frac{2U}{n}\log \frac{1}{\delta} + \frac{4}{n}\sqrt{\left(\log \frac{1}{\delta}\right)\frac{n}{2}\sup_{\pi_1,\pi_2\in \Pi}\mathbb{E}\left[\left(\mathcal{M}(\pi_1,\pi_2,\{s_{i,t}\}, \{\Gamma_{i,s_{1:T}}\})\right)^2\right]} \\
    &+ \frac{4}{n}\sqrt{\left(\log \frac{1}{\delta}\right)16U\mathbb{E}\left[\sup_{\pi_1,\pi_2\in\Pi}\left|\sum_{i=1}^n Z_i\left(\mathcal{M}(\pi_1,\pi_2,\{s_{i,t}\}, \{\Gamma_{i,s_{1:T}}\})\right)\right|\right]} \\
    &= \mathbb{E}\left[\sup_{\pi_1,\pi_2\in\Pi}\left|\tilde{\Delta}(\pi_1,\pi_2)-\Delta(\pi_1,\pi_2)\right|\right] + \frac{2U}{n}\log \frac{1}{\delta} + 2\sqrt{2\log\frac{1}{\delta}}\sqrt{\frac{\sup_{\pi_1,\pi_2\in \Pi}\mathbb{E}\left[\left(\mathcal{M}(\pi_1,\pi_2,\{s_{i,t}\}, \{\Gamma_{i,s_{1:T}}\})\right)^2\right]}{n}} \\
    &+ 16\sqrt{\left(\log \frac{1}{\delta}\right)\frac{U}{n}\mathbb{E}\left[\sup_{\pi^{\alpha},\pi^{\beta}}\frac{1}{n}\left|\sum_{i=1}^n Z_i\left(\mathcal{M}(\pi_1,\pi_2,\{s_{i,t}\}, \{\Gamma_{i,s_{1:T}}\})\right)\right|\right]} \\
    &= \mathbb{E}\left[\sup_{\pi_1,\pi_2\in\Pi}\left|\tilde{\Delta}(\pi_1,\pi_2)-\Delta(\pi_1,\pi_2)\right|\right] + 2\sqrt{2\log\frac{1}{\delta}}\sqrt{\frac{\sup_{\pi_1,\pi_2\in \Pi}\mathbb{E}\left[\left(\mathcal{M}(\pi_1,\pi_2,\{s_{i,t}\}, \{\Gamma_{i,s_{1:T}}\})\right)^2\right]}{n}} \\
    &+ \sqrt{\frac{O(\frac{1}{\sqrt{n}})}{n}} + O(\frac{1}{n}) \\
    &= \mathbb{E}\left[\sup_{\pi_1,\pi_2\in\Pi}\left|\tilde{\Delta}(\pi_1,\pi_2)-\Delta(\pi_1,\pi_2)\right|\right] + 2\sqrt{2\log\frac{1}{\delta}}\sqrt{\frac{\sup_{\pi_1,\pi_2\in \Pi}\mathbb{E}\left[\left(\mathcal{M}(\pi_1,\pi_2,\{s_{i,t}\}, \{\Gamma_{i,s_{1:T}}\})\right)^2\right]}{n}} + O(\frac{1}{n^{0.75}}).
\end{align*}
Combining this observation with Equation (\ref{eq:tilde_delta_o}), we have that with probability at least $1-2\delta$:
\begin{align*}
    &\sup_{\pi_1,\pi_2\in\Pi}\left|\tilde{\Delta}(\pi_1,\pi_2)-\Delta(\pi_1,\pi_2)\right| \\
    &\leq \mathbb{E}\left[\sup_{\pi_1,\pi_2\in\Pi}\left|\tilde{\Delta}(\pi_1,\pi_2)-\Delta(\pi_1,\pi_2)\right|\right] + 2\sqrt{2\log\frac{1}{\delta}}\sqrt{\frac{\sup_{\pi_1,\pi_2\in \Pi}\mathbb{E}\left[\left(\mathcal{M}(\pi_1,\pi_2,\{s_{i,t}\}, \{\Gamma_{i,s_{1:T}}\})\right)^2\right]}{n}} + O(\frac{1}{n^{0.75}}) \\
    &\leq O\left(\kappa(\Pi)\sqrt{\frac{\Upsilon^{\ast}_{\mathrm{DR}}}{n}}\right) + o(\frac{1}{\sqrt{n}}) + 2\sqrt{2\log\frac{1}{\delta}}\sqrt{\frac{\sup_{\pi_1,\pi_2\in \Pi}\mathbb{E}\left[\left(\mathcal{M}(\pi_1,\pi_2,\{s_{i,t}\}, \{\Gamma_{i,s_{1:T}}\})\right)^2\right]}{n}} + O(\frac{1}{n^{0.75}}) \\
    &= O\left(\left(\kappa(\Pi)+\sqrt{\log\frac{1}{\delta}}\right)\sqrt{\frac{\Upsilon^{\ast}_{\mathrm{DR}}}{n}}\right) + o(\frac{1}{\sqrt{n}})
\end{align*}

\end{proof}

\subsection{Proof of Lemma \ref{lem:tilde_hat_delta_dr}}
\label{sec:appendix_proof_tilde_hat_delta_dr}
\begin{proof}
First, we prove that $\sup_{\pi\in\Pi}\mathbb{E}[(\hat{Q}_{1,t}^{-k}-Q_{1,t})^2]=o(n^{-2\alpha_1})$ under Assumption \ref{aspp:dr_3}.
Let define $\rho^{\pi}_{t:t^{\prime}}=\prod_{l=t}^{t^{\prime}}\pi(a_{l}|s_{l})$ and $\tau_{t:t^{\prime}}=(s_t,a_t^1,a_t^2,\cdots,s_{t^{\prime}},a_{t^{\prime}}^1,a_{t^{\prime}}^2)$.
From Cauchy-Schwartz inequality, we have:
\begin{equation}
\label{eq:q_bound}
\begin{aligned}
    &\sup_{\pi\in\Pi}\mathbb{E}[(\hat{Q}_{1,t}^{-k}-Q_{1,t})^2] = \mathbb{E}[\sup_{\pi\in\Pi}(\hat{Q}_{1,t}^{-k}-Q_{1,t})^2] \\ 
    &= \mathbb{E}\left[\sup_{\pi\in\Pi}\left(\sum_{t^{\prime}=t}^T\gamma^{t^{\prime}-t}\sum_{\tau_{t+1:t^{\prime}}}\rho^{\pi}_{t+1:t^{\prime}}\left(\hat{R}_{t^{\prime}}^{-k}\prod_{l=t}^{t^{\prime}-1}\hat{P}_T^{-k}(s_{l+1}|s_{l},a_{l})-R_{t^{\prime}}\prod_{l=t}^{t^{\prime}-1}P_T(s_{l+1}|s_{l},a_{l})\right)\right)^2\right] \\
    &\leq \left(\sum_{t^{\prime}=t}^T\gamma^{2(t^{\prime}-t)}\right)\mathbb{E}\left[\sup_{\pi\in\Pi}\left(\sum_{t^{\prime}=t}^T\left(\sum_{\tau_{t+1:t^{\prime}}}\rho^{\pi}_{t+1:t^{\prime}}\left(\hat{R}_{t^{\prime}}^{-k}\prod_{l=t}^{t^{\prime}-1}\hat{P}_T^{-k}(s_{l+1}|s_{l},a_{l})-R_{t^{\prime}}\prod_{l=t}^{t^{\prime}-1}P_T(s_{l+1}|s_{l},a_{l})\right)\right)^2\right)\right] \\
    &\leq \left(\sum_{t^{\prime}=t}^T\gamma^{2(t^{\prime}-t)}\right)\sum_{t^{\prime}=t}^T\left(\sum_{\tau_{t+1:t^{\prime}}}1\right)\sum_{\tau_{t+1:t^{\prime}}}\mathbb{E}\left[\left(\hat{R}_{t^{\prime}}^{-k}\prod_{l=t}^{t^{\prime}-1}\hat{P}_T^{-k}(s_{l+1}|s_{l},a_{l})-R_{t^{\prime}}\prod_{l=t}^{t^{\prime}-1}P_T(s_{l+1}|s_{l},a_{l})\right)^2\right] \\
    &= \left(\sum_{t^{\prime}=t}^T\gamma^{2(t^{\prime}-t)}\right)\sum_{t^{\prime}=t}^T\sum_{\tau_{t+1:t^{\prime}}} o(n^{-2\alpha_1}) = o(n^{-2\alpha_1}),
\end{aligned}
\end{equation}
where $\sum_{\tau_{t+1:t}}\rho^{\pi}_{t+1:t}R_{t}\prod_{l=t}^{t-1}P_T(s_{l+1}|s_{l},a_{l})=R_{t}$ and $\sum_{\tau_{t+1:t}}\rho^{\pi}_{t+1:t}\hat{R}_{t}^{-k}\prod_{l=t}^{t-1}\hat{P}_T^{-k}(s_{l+1}|s_{l},a_{l})=\hat{R}_{t}^{-k}$.

Taking any policy profile $\pi \in \Pi$.
We start by rewriting the DR value estimator as follows:
\begin{align*}
    \hat{v}^{\mathrm{DR}}_1(\pi_1,\pi_2) =& \frac{1}{n}\sum_{i=1}^n\sum_{t=1}^T \gamma^{t-1}\left(\hat{\rho}_{i,t}^{-k(i)}\left(r_{i,t}-\hat{Q}_{1,i,t}^{-k(i)}\right)+\hat{\rho}_{i,t-1}^{-k(i)}\hat{V}_{1,i,t}^{-k(i)}\right) \\
    =& \frac{1}{n}\sum_{i=1}^n\left(\sum_{t=1}^T \gamma^{t-1}\hat{\rho}_{i,t}^{-k(i)}\left(r_{i,t}-\hat{Q}_{1,i,t}^{-k(i)}+\gamma\hat{V}_{1,i,t+1}^{-k(i)}\right)\right) + \frac{1}{n}\sum_{i=1}^n \hat{V}_{1,i,1}^{-k(i)}.
\end{align*}
Similarly, we have the oracle double robust estimator as follows:
\begin{align*}
    v_1^{\mathrm{DR}}(\pi_1,\pi_2) =& \frac{1}{n}\sum_{i=1}^n\left(\sum_{t=1}^T \gamma^{t-1}\rho_{i,t}\left(r_{i,t}-Q_{1,i,t}+\gamma V_{1,i,t+1}\right)\right) + \frac{1}{n}\sum_{i=1}^n V_{1,i,1}.
\end{align*}
Therefore, we can decompose the difference function $\hat{v}^{\mathrm{DR}}_1(\pi_1,\pi_2)-v^{\mathrm{DR}}_1(\pi_1,\pi_2)$ as follows:
\begin{align*}
    \hat{v}^{\mathrm{DR}}_1(\pi_1,\pi_2)-v^{\mathrm{DR}}_1(\pi_1,\pi_2) =& \frac{1}{n}\sum_{i=1}^n\left(\sum_{t=1}^T \gamma^{t-1}\hat{\rho}_{i,t}^{-k(i)}\left(r_{i,t}-\hat{Q}_{1,i,t}^{-k(i)}+\gamma\hat{V}_{1,i,t+1}^{-k(i)}\right)\right) + \frac{1}{n}\sum_{i=1}^n \hat{V}_{1,i,1}^{-k(i)} \\
    &- \frac{1}{n}\sum_{i=1}^n\left(\sum_{t=1}^T \gamma^{t-1}\rho_{i,t}\left(r_{i,t}-Q_{1,i,t}+\gamma V_{1,i,t+1}\right)\right) + \frac{1}{n}\sum_{i=1}^n V_{1,i,1} \\
    =& \sum_{t=1}^T \gamma^{t-1}\left(\frac{1}{n}\sum_{i=1}^n\left(\rho_{i,t}\left(-\hat{Q}_{1,i,t}^{-k(i)} + Q_{1,i,t} \right) + \rho_{i,t-1}\left(\hat{V}_{1,i,t}^{-k(i)} - V_{1,i,t}\right)\right)\right) \\
    &+ \sum_{t=1}^T \gamma^{t-1}\left(\frac{1}{n}\sum_{i=1}^n\left(\hat{\rho}_{i,t}^{-k(i)} - \rho_{i,t}\right)\left(r_{i,t}-Q_{1,i,t} + \gamma V_{1,i,t+1}\right)\right) \\
    &+ \sum_{t=1}^T \gamma^{t-1}\left(\frac{1}{n}\sum_{i=1}^n\left(\hat{\rho}_{i,t}^{-k(i)} - \rho_{i,t}\right)\left(-\hat{Q}_{1,i,t}^{-k(i)} + Q_{1,i,t} \right)\right) \\
    &+ \sum_{t=1}^T \gamma^{t-1}\left(\frac{1}{n}\sum_{i=1}^n\left(\hat{\rho}_{i,t-1}^{-k(i)} - \rho_{i,t-1}\right)\left(\hat{V}_{1,i,t}^{-k(i)} - V_{1,i,t}\right)\right).
\end{align*}
For each of reference, denote:
\begin{enumerate}
    \item $S_{1}^{t}(\pi)\triangleq \frac{1}{n}\sum_{i=1}^n\left(\rho_{i,t}\left(-\hat{Q}_{1,i,t}^{-k(i)} + Q_{1,i,t} \right) + \rho_{i,t-1}\left(\hat{V}_{1,i,t}^{-k(i)} - V_{1,i,t}\right)\right)$.
    \item $S_{2}^{t}(\pi)\triangleq \frac{1}{n}\sum_{i=1}^n\left(\hat{\rho}_{i,t}^{-k(i)} - \rho_{i,t}\right)\left(r_{i,t}-Q_{1,i,t} + \gamma V_{1,i,t+1}\right)$.
    \item $S_{3}^{t}(\pi)\triangleq \frac{1}{n}\sum_{i=1}^n\left(\hat{\rho}_{i,t}^{-k(i)} - \rho_{i,t}\right)\left(-\hat{Q}_{1,i,t}^{-k(i)} + Q_{1,i,t} \right)$.
    \item $S_{4}^{t}(\pi)\triangleq \frac{1}{n}\sum_{i=1}^n\left(\hat{\rho}_{i,t-1}^{-k(i)} - \rho_{i,t-1}\right)\left(\hat{V}_{1,i,t}^{-k(i)} - V_{1,i,t}\right)$.
\end{enumerate}
Hereafter, we bound $\sup\limits_{\pi^{\alpha},\pi^{\beta}\in\Pi}|S_{1}^{t}(\pi^{\alpha})-S_{1}^{t}(\pi^{\beta})|$, $\sup\limits_{\pi^{\alpha},\pi^{\beta}\in\Pi}|S_{2}^{t}(\pi^{\alpha})-S_{2}^{t}(\pi^{\beta})|$, $\sup\limits_{\pi\in\Pi}|S_{3}^{t}(\pi)|$, and $\sup\limits_{\pi\in\Pi}|S_{4}^{t}(\pi)|$ in turn.
Define further:
\begin{enumerate}
    \item $S_{1}^{t,k}(\pi)\triangleq \frac{1}{n}\sum_{\{i|k(i)=k\}}\left(\rho_{i,t}\left(-\hat{Q}_{1,i,t}^{-k(i)} + Q_{1,i,t} \right) + \rho_{i,t-1}\left(\hat{V}_{1,i,t}^{-k(i)} - V_{1,i,t}\right)\right)$.
    \item $S_{2}^{t,k}(\pi)\triangleq \frac{1}{n}\sum_{\{i|k(i)=k\}}\left(\hat{\rho}_{i,t}^{-k(i)} - \rho_{i,t}\right)\left(r_{i,t}-Q_{1,i,t} + \gamma V_{1,i,t+1}\right)$.
\end{enumerate}
Clearly, $S_{1}^{t}(\pi)=\sum_{k=1}^K S_{1}^{t,k}(\pi)$, $S_{2}^{t}(\pi)=\sum_{k=1}^K S_{2}^{t,k}(\pi)$.

Now since $\hat{Q}_{1,t}^{-k(i)}$ is computed using the rest $K-1$ folds, when we condition on the data in the rest $K-1$ folds, $\hat{Q}_{1,t}^{-k(i)}$ is fixed estimator.
Consequently, conditioned on $\hat{Q}_{1,t}^{-k(i)}$, $S_{1}^{t,k}(\pi)$ is a sum of {\bf iid} bounded random variables with zero mean, because:
\begin{align*}
    &\mathbb{E}\left[\rho_{i,t}\left(-\hat{Q}_{1,i,t}^{-k(i)} + Q_{1,i,t} \right) + \rho_{i,t-1}\left(\hat{V}_{1,i,t}^{-k(i)} - V_{1,i,t}\right)\right] \\
    &=\mathbb{E}\left[\rho_{i,t-1}\left(\mathbb{E}\left[\eta_{i,t}\left(-\hat{Q}_{1,i,t}^{-k(i)} + Q_{1,i,t} \right) + \left(\hat{V}_{1,i,t}^{-k(i)} - V_{1,i,t}\right) | s_1,a_1^1,a_1^2,\cdots,s_{t-1},a_{t-1}^1,a_{t-1}^2,s_t\right]\right)\right] = 0.
\end{align*}
Besides, as in Equation (\ref{eq:q_bound}), we can decompose $\hat{Q}_{1,t}^{-k}$ into $\pi$ and other terms that are independent of $\pi$.
Therefore, defining $S_{1,i}^{t}(\pi)=\rho_{i,t}\left(-\hat{Q}_{1,i,t}^{-k(i)} + Q_{1,i,t} \right) + \rho_{i,t-1}\left(\hat{V}_{1,i,t}^{-k(i)} - V_{1,i,t}\right)$, we can obtain the bound on $\sup_{\pi^{\alpha},\pi^{\beta}\in\Pi}|S_{1}^{t,k}(\pi^{\alpha})-S_{1}^{t,k}(\pi^{\beta})|$ as in Lemma \ref{lem:tilde_delta_dr}: $\forall \delta>0$, with probability at least $1-2\delta$,
\begin{align*}
    &K \sup_{\pi^{\alpha},\pi^{\beta}\in\Pi}|S_{1}^{t,k}(\pi^{\alpha})-S_{1}^{t,k}(\pi^{\beta})| \\
    \leq& O\left(\left(\kappa(\Pi)+\sqrt{\log\frac{1}{\delta}}\right)\sqrt{\frac{\sup\limits_{\pi^{\alpha},\pi^{\beta}\in\Pi}\mathbb{E}\left[\left( S_{1,i}^{t}(\pi^{\alpha})-S_{1,i}^{t}(\pi^{\beta})\right)^2 | \hat{Q}_{1,t}^{-k(i)} \right]}{\frac{n}{K}}}\right) + o(\frac{1}{\sqrt{n}}) \\
    \leq& 4\cdot O\left(\left(\kappa(\Pi)+\sqrt{\log\frac{1}{\delta}}\right)\sqrt{\frac{\sup\limits_{\pi\in\Pi}\mathbb{E}\left[\left\|\Gamma_i\right\|_2^2 | \hat{Q}_{1,t}^{-k(i)} \right]}{\frac{n}{K}}}\right) + o(\frac{1}{\sqrt{n}})
\end{align*}
where $\Gamma_{i}=\frac{-\hat{Q}_{1,i,t}^{-k(i)} + Q_{1,i,t}}{\prod_{t^{\prime}=1}^t\pi^b_{t^{\prime}}(a_{i,t^{\prime}}|s_{i,t^{\prime}})}A_{i,1:t} + \frac{A_{i,1:t-1}\otimes(\hat{Q}_{1,t}^{-k(i)}(s_{i,t}) - Q_{1,t}(s_{i,t}))}{\prod_{t^{\prime}=1}^{t-1}\pi^b_{t^{\prime}}(a_{i,t^{\prime}}|s_{i,t^{\prime}})}$, and the second inequality follows from Cauchy-Schwartz.
Thus, from Assumption \ref{asp:overlap_and_reward}, for any $a^1\in \mathcal{A}_1,a^2\in \mathcal{A}^2$:
\begin{align*}
    &K \sup_{\pi^{\alpha},\pi^{\beta}\in\Pi}|S_{1}^{t,k}(\pi^{\alpha})-S_{1}^{t,k}(\pi^{\beta})| \\
    \leq& 8C^t\sqrt{d} \cdot O\left(\left(\kappa(\Pi)+\sqrt{\log\frac{1}{\delta}}\right)\sqrt{\frac{K\mathbb{E}\left[\sup\limits_{\pi\in\Pi}|\hat{Q}_{1,t}^{-k(i)}(s_{i,t},a^1,a^2) - Q_{1,t}(s_{i,t},a^1,a^2)|^2 | \hat{Q}_{1,t}^{-k(i)} \right]}{n}}\right) + o(\frac{1}{\sqrt{n}}).
\end{align*}
From Equation (\ref{eq:q_bound}), it follows that $\mathbb{E}\left[\sup_{\pi\in\Pi}\left(\hat{Q}_{1,t}^{-k(i)}(s_{i,t}, a) - Q_{1,t}(s_{i,t}, a)\right)^2\right] = o(n^{-2\alpha_1})$.
Consequently, Markov's inequality immediately implies that $\sup_{\pi\in\Pi}\mathbb{E}\left[\left(\hat{Q}_{1,t}^{-k(i)}(s_{i,t}, a) - Q_{1,t}(s_{i,t}, a)\right)^2 | \hat{Q}_{1,t}^{-k(i)}\right] = o_p(n^{-2\alpha_1})$.
Therefore, from $\alpha_1>0$, we immediately have: $\sup_{\pi^{\alpha},\pi^{\beta}\in\Pi}|S_{1}^{t,k}(\pi^{\alpha})-S_{1}^{t,k}(\pi^{\beta})|=o_p(n^{-0.5-\alpha_1})+o_p(\frac{1}{\sqrt{n}})=o_p(\frac{1}{\sqrt{n}})$.
Consequently,
\begin{align*}
    \sup_{\pi^{\alpha},\pi^{\beta}\in\Pi}|S_{1}^{t}(\pi^{\alpha})-S_{1}^{t}(\pi^{\beta})| &= \sup_{\pi^{\alpha},\pi^{\beta}\in\Pi}|\sum_{k=1}^K \left(S_{1}^{t,k}(\pi^{\alpha})-S_{1}^{t,k}(\pi^{\beta})\right)| \\
    &\leq \sum_{k=1}^K \sup_{\pi^{\alpha},\pi^{\beta}\in\Pi}|S_{1}^{t,k}(\pi^{\alpha})-S_{1}^{t,k}(\pi^{\beta})| = o_p(\frac{1}{\sqrt{n}}).
\end{align*}
By exactly the same argument, we have $\sup_{\pi^{\alpha},\pi^{\beta}\in\Pi}|S_{2}^{t}(\pi^{\alpha})-S_{2}^{t}(\pi^{\beta})| = o_p(\frac{1}{\sqrt{n}})$.

Next, we bound the contribution from $S_{3}^{t}(\pi)$ as follow:
\begin{align*}
    &\sup_{\pi\in \Pi}|S_{3}^{t}(\pi)| = \sup_{\pi^{\alpha},\pi^{\beta}\in\Pi}\frac{1}{n}\left|\sum_{i=1}^n\left(\hat{\rho}_{i,t}^{-k(i)} - \rho_{i,t}\right)\left(-\hat{Q}_{1,i,t}^{-k(i)} + Q_{1,i,t} \right)\right| \\
    &\leq \frac{1}{n}\sum_{i=1}^n\sup_{\pi^{\alpha},\pi^{\beta}\in\Pi}\left|\hat{\rho}_{i,t}^{-k(i)} - \rho_{i,t}\right|\cdot\sup_{\pi^{\alpha},\pi^{\beta}\in\Pi}\left|\hat{Q}_{1,i,t}^{-k(i)} - Q_{1,i,t}\right| \\
    &\leq \sqrt{\frac{1}{n}\sum_{i=1}^n\sup_{\pi^{\alpha},\pi^{\beta}\in\Pi}\left(\hat{\rho}_{i,t}^{-k(i)} - \rho_{i,t}\right)^2}\sqrt{\frac{1}{n}\sum_{i=1}^n\sup_{\pi^{\alpha},\pi^{\beta}\in\Pi}\left(\hat{Q}_{1,i,t}^{-k(i)} - Q_{1,i,t}\right)^2},
\end{align*}
where the last inequality follows from Cauchy-Schwartz.
Taking expectation of both sides yields:
\begin{align*}
    &\mathbb{E}\left[\sup_{\pi\in \Pi}|S_{3}^{t}(\pi)|\right] \leq \mathbb{E}\left[\sqrt{\frac{1}{n}\sum_{i=1}^n\sup_{\pi^{\alpha},\pi^{\beta}\in\Pi}\left(\hat{\rho}_{i,t}^{-k(i)} - \rho_{i,t}\right)^2}\sqrt{\frac{1}{n}\sum_{i=1}^n\sup_{\pi^{\alpha},\pi^{\beta}\in\Pi}\left(\hat{Q}_{1,i,t}^{-k(i)} - Q_{1,i,t}\right)^2}\right] \\
    & \leq \sqrt{\mathbb{E}\left[\frac{1}{n}\sum_{i=1}^n\sup_{\pi^{\alpha},\pi^{\beta}\in\Pi}\left(\hat{\rho}_{i,t}^{-k(i)} - \rho_{i,t}\right)^2\right]} \sqrt{\mathbb{E}\left[\frac{1}{n}\sum_{i=1}^n\sup_{\pi^{\alpha},\pi^{\beta}\in\Pi}\left(-\hat{Q}_{1,i,t}^{-k(i)} + Q_{1,i,t}\right)^2\right]} \\
    &= \sqrt{\frac{1}{n}\sum_{i=1}^n\mathbb{E}\left[\sup_{\pi^{\alpha},\pi^{\beta}\in\Pi}\left(\hat{\rho}_{i,t}^{-k(i)} - \rho_{i,t}\right)^2\right]} \sqrt{\frac{1}{n}\sum_{i=1}^n\mathbb{E}\left[\sup_{\pi^{\alpha},\pi^{\beta}\in\Pi}\left(-\hat{Q}_{1,i,t}^{-k(i)} + Q_{1,i,t}\right)^2\right]} \\
    & \leq \sqrt{\frac{1}{n}\sum_{i=1}^n \frac{s(\frac{K-1}{K}n)}{\left(\frac{K-1}{K}n\right)^{2\alpha_2}}}\sqrt{\frac{1}{n}\sum_{i=1}^n \frac{s(\frac{K-1}{K}n)}{\left(\frac{K-1}{K}n\right)^{2\alpha_1}}} \leq \sqrt{\frac{s(\frac{K-1}{K}n)}{\left(\frac{K-1}{K}n\right)^{2\alpha_2}}}\sqrt{\frac{s(\frac{K-1}{K}n)}{\left(\frac{K-1}{K}n\right)^{2\alpha_1}}} \leq \frac{s(\frac{K-1}{K}n)}{\sqrt{\left(\frac{K-1}{K}n\right)^{2(\alpha_1+\alpha_2)}}} = o(\frac{1}{\sqrt{n}}),
\end{align*}
where the second inequality again follows from Cauchy-Schwartz and the last equality follows from $s(n)=o(1)$.
Consequently, by Markov's inequality, this equation immediately implies $\sup_{\pi\in \Pi}|S_{3}^{t}(\pi)|=o_p(\frac{1}{\sqrt{n}})$.
By exactly the same argument, we have $\sup_{\pi\in\Pi}|S_{4}^{t}(\pi)| = o_p(\frac{1}{\sqrt{n}})$.
Putting the above bound for $\sup_{\pi\in\Pi}|S_{1}^{t}(\pi)|$, $\sup_{\pi\in\Pi}|S_{2}^{t}(\pi)|$, $\sup_{\pi\in\Pi}|S_{3}^{t}(\pi)|$ and $\sup_{\pi\in\Pi}|S_{4}^{t}(\pi)|$ together, we therefore have the claim established:
\begin{align*}
    &\sup_{\pi^{\alpha},\pi^{\beta}\in\Pi} \left|\hat{\Delta}(\pi^{\alpha},\pi^{\beta}) - \tilde{\Delta}(\pi^{\alpha},\pi^{\beta})\right| \\
    &= \sup_{\pi^{\alpha},\pi^{\beta}\in\Pi} \left|\sum_{t=1}^T \gamma^{t-1}\left(S_{1}^{t}(\pi^{\alpha}) - S_{1}^{t}(\pi^{\beta}) + S_{2}^{t}(\pi^{\alpha}) - S_{2}^{t}(\pi^{\beta}) + S_{3}^{t}(\pi^{\alpha}) - S_{3}^{t}(\pi^{\beta}) + S_{4}^{t}(\pi^{\alpha}) - S_{4}^{t}(\pi^{\beta})\right)\right| \\
    &\leq \sum_{t=1}^T \gamma^{t-1}\sup_{\pi^{\alpha},\pi^{\beta}\in\Pi} \left|S_{1}^{t}(\pi^{\alpha}) - S_{1}^{t}(\pi^{\beta}) + S_{2}^{t}(\pi^{\alpha}) - S_{2}^{t}(\pi^{\beta}) + S_{3}^{t}(\pi^{\alpha}) - S_{3}^{t}(\pi^{\beta}) + S_{4}^{t}(\pi^{\alpha}) - S_{4}^{t}(\pi^{\beta})\right| \\
    &\leq  \sum_{t=1}^T \gamma^{t-1} \left(\sup_{\pi^{\alpha},\pi^{\beta}\in\Pi}|S_{1}^{t}(\pi^{\alpha})-S_{1}^{t}(\pi^{\beta})| + \sup_{\pi^{\alpha},\pi^{\beta}\in\Pi}|S_{2}^{t}(\pi^{\alpha})-S_{2}^{t}(\pi^{\beta})| + 2\sup_{\pi\in\Pi}|S_{3}^{t}(\pi)| + 2\sup_{\pi\in\Pi}|S_{4}^{t}(\pi)|\right) \\
    &= o_p(\frac{1}{\sqrt{n}}).
\end{align*}

\end{proof}

\subsection{Proof of Lemma \ref{lem:hat_delta_dr}}
\label{sec:appendix_proof_hat_delta_dr}
\begin{proof}
We have:
\begin{align*}
    &\sup_{\pi^{\alpha},\pi^{\beta}\in \Pi}\left|\hat{\Delta}(\pi^{\alpha}, \pi^{\beta}) - \Delta(\pi^{\alpha}, \pi^{\beta})\right| \\
    &\leq \sup_{\pi^{\alpha},\pi^{\beta}\in \Pi}\left|\hat{\Delta}(\pi^{\alpha}, \pi^{\beta}) - \tilde{\Delta}(\pi^{\alpha}, \pi^{\beta}) - \Delta(\pi^{\alpha}, \pi^{\beta}) + \tilde{\Delta}(\pi^{\alpha}, \pi^{\beta})\right| \\
    &\leq \sup_{\pi^{\alpha},\pi^{\beta}\in \Pi}\left|\hat{\Delta}(\pi^{\alpha}, \pi^{\beta}) - \tilde{\Delta}(\pi^{\alpha}, \pi^{\beta})\right| + \sup_{\pi^{\alpha},\pi^{\beta}\in \Pi}\left|\tilde{\Delta}(\pi^{\alpha}, \pi^{\beta}) - \Delta(\pi^{\alpha}, \pi^{\beta})\right|.
\end{align*}
Therefore, based on Lemmas \ref{lem:tilde_delta_dr} and \ref{lem:tilde_hat_delta_dr}, for $\delta>0$, there exists $C>0$, $N_{\delta}>0$, such that with probability at least $1-2\delta$ and for all $n\geq N_{\delta}$:
\begin{align}
    \sup_{\pi^{\alpha},\pi^{\beta}\in \Pi}\left|\hat{\Delta}(\pi^{\alpha},\pi^{\beta})-\Delta(\pi^{\alpha},\pi^{\beta})\right|\leq C\left(\left(\kappa(\Pi)+\sqrt{\log\frac{1}{\delta}}\right)\sqrt{\frac{\Upsilon^{\ast}_{\mathrm{DR}}}{n}}\right).
\end{align}
\end{proof}

\subsection{Proof of Lemma \ref{lem:tilde_delta_drl}}
\label{sec:regret_bound_drl}
\begin{proof}
Since the proof of Lemma \ref{lem:tilde_delta_drl} is almost same as Lemma \ref{lem:tilde_delta_dr}, we omit the proof.
\end{proof}

\subsection{Proof of Lemma \ref{lem:rademacher}}
\begin{proof}
First, we introduce the following definitions:
\begin{definition}
Given the state space $\mathcal{S}$, a policy profile class $\Pi$, a set of $n$ state trajectories $\{\{s_{1,t}\}, \cdots, \{s_{n,t}\}\}$, define:
\begin{enumerate}
    \item Hamming distance between any two policy profiles $\pi^{\alpha}$ and $\pi^{\beta}$ in $\Pi$: $H_{n}(\pi^{\alpha},\pi^{\beta})=\frac{1}{n}\sum_{i=1}^n\bm{1}(\{\bigvee_{t=1}^T\pi_{1,t}^{\alpha}(s_{i,t})\neq\pi_{1,t}^{\beta}(s_{i,t})\}\vee\{\bigvee_{t=1}^T\pi_{2,t}^{\alpha}(s_{i,t})\neq\pi_{2,t}^{\beta}(s_{i,t})\})$.
    \item $\epsilon$-Hamming covering number of the set $\{\{s_{1,t}\}, \cdots, \{s_{n,t}\}\}\}$: $N_{H}(\epsilon, \Pi, \{\{s_{1,t}\}, \cdots, \{s_{n,t}\}\}\})$ is the smallest number $K$ of policy profiles $\{\pi_1,\cdots,\pi_K\}$ in $\Pi$, such that $\forall \pi \in \Pi, \exists \pi_i,H_n(\pi,\pi_i)\leq \epsilon$.
    \item $\epsilon$-Hamming covering number of $\Pi$: $N_{H}(\epsilon, \Pi)=\sup\{N_{H}(\epsilon, \Pi, \{\{s_{1,t}\}, \cdots, \{s_{m,t}\}\}\}) ~|~ m \geq 1, \{s_{1,t}\}, \cdots, \{s_{m,t}\}\}$.
    \item Entropy integral: $\kappa(\Pi)=\int_{0}^{\infty}\sqrt{\log N_{H}(\epsilon^2, \Pi)}d\epsilon$.
\end{enumerate}
\end{definition}

\begin{definition}
Given a set of $n$ state trajectories $\{\{s_{1,t}\}, \cdots, \{s_{n,t}\}\}\}$, and a set of $n$ weights $\bm{\Gamma}=\{\{\Gamma_{1,t}\}_{t=1}^T,\cdots,\{\Gamma_{n,t}\}_{t=1}^T\}$, we define the following distances $I_{\Gamma}(\pi_1,\pi_2)$ between two policy profiles $\pi_1$ and $\pi_2$ in $\Pi$ and the corresponding covering number $N_{I_{\Gamma}}(\epsilon, \Pi, \{\{s_{1,t}\}, \cdots, \{s_{n,t}\}\}\})$ as follows:
\begin{enumerate}
    \item $I_{\Gamma}(\pi_1,\pi_2)=\sqrt{\frac{\sum_{i=1}^n|\sum_{s_{1:T}} \langle \pi_1(s_{1:T})-\pi_2(s_{1:T}), \Gamma_{i,s_{1:T}} \rangle|^2}{\sup_{\pi^{\alpha},\pi^{\beta}\in \Pi}\sum_{i=1}^n|\sum_{s_{1:T}} \langle \pi^{\alpha}(s_{1:T})-\pi^{\beta}(s_{1:T}), \Gamma_{i,s_{1:T}} \rangle|^2}}$, where we set $0 \triangleq \frac{0}{0}$.
    \item $N_{I_{\Gamma}}(\epsilon, \Pi, \{\{s_{1,t}\}, \cdots, \{s_{n,t}\}\}\})$: the minimum number of policy profiles needed to $\epsilon$-cover $\Pi$ under $I_{\Gamma}$.
\end{enumerate}
\end{definition}

Based on these definitions, we introduce the following lemma.
\begin{lemma}
\label{lem:dist}
For any $n$, any $\bm{\Gamma}=\{\{\Gamma_{1,t}\}_{t}^T,\cdots, \{\Gamma_{n,t}\}_{t}^T\}$ and any $\{\{s_{1,t}\}, \cdots, \{s_{n,t}\}\}\}$:
\begin{enumerate}
    \item Triangle inequality holds for sum of inner product distance: $I_{\Gamma}(\pi_1,\pi_2) \leq I_{\Gamma}(\pi_1,\pi_3) + I_{\Gamma}(\pi_3,\pi_2)$.
    \item $N_{I_{\Gamma}}(\epsilon, \Pi, \{\{s_{1,t}\}, \cdots, \{s_{n,t}\}\}\})\leq N_{H}(\epsilon^2,\Pi)$.
\end{enumerate}
\end{lemma}

Here, we break the proof into four main components.
\paragraph{Step 1: Policy profile approximations.}

Set $\epsilon_j=\frac{1}{2^j}$ and let $S_0, S_1, S_2, \cdots, S_J$ be a sequence of policy profile classes such that $S_j$ $\epsilon_j$-cover $\Pi$ under the sum of inner product distance:
\begin{align*}
    \forall \pi\in \Pi, \exists \pi^{\prime}\in S_j, I_{\Gamma}(\pi, \pi^{\prime})\leq \epsilon_j,
\end{align*}
where $J=\lceil \log_2(n)(1-\omega)\rceil$.
Note that by definition of the covering number under the sum of inner product distance, we can choose the $j$-th policy profile class $S_j$ such that $|S_j|=N_{I_{\Gamma}}(2^{-j}, \Pi, \{\{s_{1,t}\}, \cdots, \{s_{n,t}\}\}\})$.
Additionally, we define refining approximation operators $A_j: \Pi\to \Pi ~(j=0,\cdots, J)$ as follows:
\begin{align*}
    A_j(\pi) = 
    \begin{cases}
    \argmin_{\pi^{\prime}\in S_J}I_{\Gamma}(\pi, \pi^{\prime}) & (j = J) \\
    \argmin_{\pi^{\prime}\in S_j}I_{\Gamma}(A_{j+1}(\pi), \pi^{\prime}) & (j \neq J)
    \end{cases}
    .
\end{align*}

By these definitions, we can obtain the following properties:
\begin{enumerate}
    \item $\max_{\pi\in \Pi}I_{\Gamma}(\pi, A_J(\pi))\leq 2^{-J}$: \\
    Pick any $\pi\in \Pi$.
    By the definition of $S_J$, $\exists \pi^{\prime}\in S_J, I_{\Gamma}(\pi, \pi^{\prime})\leq \epsilon_J$.
    By the definition of $A_J$, we have $I_{\Gamma}(\pi, A_J(\pi)) \leq I_{\Gamma}(\pi, \pi^{\prime}) \leq \epsilon_J = 2^{-J}$.
    Taking maximum over all $\pi\in \Pi$ verifies this property.
    \item $|\{A_j(\pi) | \pi\in \Pi\}| \leq N_{I_{\Gamma}}(2^{-j}, \Pi, \{\{s_{1,t}\}, \cdots, \{s_{n,t}\}\}\})$, for every $j=0,\cdots, J$: \\
    Since $A_j(\pi)=\argmin_{\pi^{\prime}\in S_j}I_{\Gamma}(A_{j+1}(\pi),\pi^{\prime})$, $A_j(\pi)\in S_j$ for every $\pi\in \Pi$.
    Consequently, we have: $$|\{A_j(\pi)|\pi\in \Pi\}| \leq |S_j| = N_{I_{\Gamma}}(2^{-j}, \Pi, \{\{s_{1,t}\}, \cdots, \{s_{n,t}\}\}\})$$.
    \item $\max_{\pi\in \Pi}I_{\Gamma}(A_j(\pi), A_{j+1}(\pi))\leq 2^{-(j-1)}$, for very $j=0,\cdots, J-1$: \\
    From Lemma \ref{lem:dist}, since $I_{\Gamma}$ satisfies the triangle inequality, we have:
    \begin{align*}
        \max_{\pi\in \Pi}I_{\Gamma}(A_j(\pi), A_{j+1}(\pi)) &\leq \max_{\pi\in \Pi}\left(I_{\Gamma}(A_j(\pi), \pi) + I_{\Gamma}(A_{j+1}(\pi), \pi)\right) \\
        &\leq \max_{\pi\in \Pi}I_{\Gamma}(A_j(\pi), \pi) + \max_{\pi\in \Pi}I_{\Gamma}(A_{j+1}(\pi), \pi) \\
        &\leq 2^{-j} + 2^{-(j+1)} \leq 2^{-(j-1)}.
    \end{align*}
    \item For any $J\geq j^{\prime} \geq j \geq 0$, $|\{(A_j(\pi), A_{j^{\prime}}(\pi)) | \pi\in \Pi\}| \leq N_{I_{\Gamma}}(2^{-j^{\prime}}, \Pi, \{\{s_{1,t}\}, \cdots, \{s_{n,t}\}\}\})$: \\
    If $A_{j^{\prime}}(\pi) = A_{j^{\prime}}(\tilde{\pi})$, then by the definition of $A_j$, we have:
    \begin{align*}
        A_{j^{\prime}-1}(\pi) = \argmin_{\pi^{\prime}\in S_{j^{\prime}}} I_{\Gamma}(A_{j^{\prime}}(\pi), \pi^{\prime}) = \argmin_{\pi^{\prime}\in S_{j^{\prime}}} I_{\Gamma}(A_{j^{\prime}}(\tilde{\pi}), \pi^{\prime}) = A_{j^{\prime}-1}(\tilde{\pi}).
    \end{align*}
    Consequently, by backward induction, it then follows that $A_j(\pi) = A_j(\tilde{\pi})$.
    Therefore, 
    \begin{align*}
        |\{(A_j(\pi), A_{j^{\prime}}(\pi)) | \pi\in \Pi\}| = |\{A_{j^{\prime}}(\pi) | \pi \in \Pi\}| \leq N_{I_{\Gamma}}(2^{-j^{\prime}}, \Pi, \{\{s_{1,t}\}, \cdots, \{s_{n,t}\}\}\})
    \end{align*}
\end{enumerate}

\paragraph{Step 2: Chaining with concentration inequalities in the negligible regime.}

For each policy profile $\pi \in \Pi$, we can write it in term of the approximation policy profiles as: $\pi=A_0(\pi) + \sum_{j=1}^{\underline{J}}(A_j(\pi)-A_{j-1}(\pi)) + (A_J(\pi) - A_{\underline{J}}(\pi)) + (\pi - A_J(\pi))$, where $\underline{J} = \lfloor\frac{1}{2}(1-\omega)\log_2(n)\rfloor$.
Therefore, we have:
\begin{align}
\label{eq:pol_diff}
\begin{aligned}
\pi^{\alpha} - \pi^{\beta} =& \left(A_0(\pi^{\alpha}) + \sum_{j=1}^{\underline{J}}(A_j(\pi^{\alpha})-A_{j-1}(\pi^{\alpha})) + (A_J(\pi^{\alpha}) - A_{\underline{J}}(\pi^{\alpha})) + (\pi^{\alpha} - A_J(\pi^{\alpha}))\right) \\
& - \left(A_0(\pi^{\beta}) + \sum_{j=1}^{\underline{J}}(A_j(\pi^{\beta})-A_{j-1}(\pi^{\beta})) + (A_J(\pi^{\beta}) - A_{\underline{J}}(\pi^{\beta})) + (\pi^{\beta} - A_J(\pi^{\beta}))\right) \\
=& \left((\pi^{\alpha} - A_J(\pi^{\alpha})) - (\pi^{\beta} - A_J(\pi^{\beta}))\right) + \left((A_J(\pi^{\alpha})-A_{\underline{J}}(\pi^{\alpha}))+(A_J(\pi^{\beta})-A_{\underline{J}}(\pi^{\beta}))\right) \\
&+ \left(\sum_{j=1}^{\underline{J}}(A_j(\pi^{\alpha})-A_{j-1}(\pi^{\alpha})) - \sum_{j=1}^{\underline{J}}(A_j(\pi^{\beta})-A_{j-1}(\pi^{\beta}))\right),
\end{aligned}
\end{align}
where the second equality follows from that $\{A_0(\pi)\}$ is a singleton set.
Hereafter, for simplicity, we define:  $$\mathcal{M}(\pi^{\alpha},\pi^{\beta},\{s_{i,t}\}, \{\Gamma_{i,s_{1:T}}\})=\sum_{s_{1:T}} \langle \pi^{\alpha}(s_{1:T})-\pi^{\beta}(s_{1:T}), \Gamma_{i,s_{1:T}} \rangle$$.
In this step, we establish two claims, for any $\pi\in \Pi$:
\begin{enumerate}
    \item $\lim_{n\to \infty} \sqrt{n}\mathbb{E}\left[\sup_{\pi\in \Pi}\left|\frac{1}{n}\sum_{i=1}^n Z_i\sum_{s_{1:T}} \langle \pi(s_{1:T})-A_J(\pi)(s_{1:T}), \Gamma_{i,s_{1:T}} \rangle\right|\right] = 0$: \\
    By Cauchy-Schwartz inequality and $I_{\Gamma}(\pi,A_J(\pi))\leq 2^{-J},\forall \pi\in\Pi$, we have:
    \begin{align*}
        &\sup_{\pi\in \Pi}\left|\frac{1}{n}\sum_{i=1}^n Z_i\mathcal{M}(\pi,A_J(\pi),\{s_{i,t}\}, \{\Gamma_{i,s_{1:T}}\}) \right| \\
        &\leq \sup_{\pi \in \Pi}\sqrt{\frac{1}{n}\left(\sum_{i=1}^n\left|\mathcal{M}(\pi,A_J(\pi),\{s_{i,t}\}, \{\Gamma_{i,s_{1:T}}\})\right|^2\right)} \\
        &= \sup_{\pi \in \Pi}\sqrt{\frac{\sum_{i=1}^n\left|\mathcal{M}(\pi,A_J(\pi),\{s_{i,t}\}, \{\Gamma_{i,s_{1:T}}\})\right|^2}{\sup_{\pi^{\alpha},\pi^{\beta}}\sum_{i=1}^n\left|\mathcal{M}(\pi^{\alpha},\pi^{\beta},\{s_{i,t}\}, \{\Gamma_{i,s_{1:T}}\})\right|^2}}\sqrt{\frac{\sup_{\pi^{\alpha},\pi^{\beta}}\sum_{i=1}^n\left|\mathcal{M}(\pi^{\alpha},\pi^{\beta},\{s_{i,t}\}, \{\Gamma_{i,s_{1:T}}\})\right|^2}{n}} \\
        &= \sup_{\pi \in \Pi}I_{\Gamma}(\pi,A_J(\pi))\sqrt{\frac{\sup_{\pi^{\alpha},\pi^{\beta}}\sum_{i=1}^n\left|\mathcal{M}(\pi^{\alpha},\pi^{\beta},\{s_{i,t}\}, \{\Gamma_{i,s_{1:T}}\})\right|^2}{n}} \\
        &\leq \sup_{\pi \in \Pi}I_{\Gamma}(\pi,A_J(\pi))\sqrt{\frac{\sup_{\pi^{\alpha},\pi^{\beta}}\sum_{i=1}^n\left(\sum_{s_{1:T}}\left|\left\langle\pi^{\alpha}(s_{1:T})-\pi^{\beta}(s_{1:T}), \Gamma_{i,s_{1:T}}\right\rangle\right|\right)^2}{n}} \\
        &\leq \sup_{\pi \in \Pi}I_{\Gamma}(\pi,A_J(\pi))\sqrt{\frac{\sup_{\pi^{\alpha},\pi^{\beta}}\sum_{i=1}^n\left(\sum_{s_{1:T}} \left\|\pi^{\alpha}(s_{1:T})-\pi^{\beta}(s_{1:T})\right\|_2\left\| \Gamma_{i,s_{1:T}}\right\|_2\right)^2}{n}} \\
        &\leq \sqrt{2}\sup_{\pi \in \Pi}I_{\Gamma}(\pi,A_J(\pi))\sqrt{\frac{\sum_{i=1}^n\left(\sum_{s_{1:T}} \left\|\Gamma_{i,s_{1:T}}\right\|_2\right)^2}{n}} \\
        &\leq \sqrt{2}\left(|\mathcal{S}|^{2T}\max_{i,s_{1:T}}\left\|\Gamma_{i,s_{1:T}}\right\|_{\infty}\right)\sup_{\pi \in \Pi}I_{\Gamma}(\pi,A_J(\pi)) \leq \sqrt{2}\left(|\mathcal{S}|^{2T}\max_{i,s_{1:T}}\left\|\Gamma_{i,s_{1:T}}\right\|_{\infty}\right)2^{-J} \\
        &= \sqrt{2}\left(|\mathcal{S}|^{2T}\max_{i,s_{1:T}}\left\|\Gamma_{i,s_{1:T}}\right\|_{\infty}\right)2^{-\lceil \log_2(n)(1-\omega)\rceil} \leq \frac{\sqrt{2}}{n^{1-\omega}}\left(|\mathcal{S}|^{2T}\max_{i,s_{1:T}}\left\|\Gamma_{i,s_{1:T}}\right\|_{\infty}\right).
    \end{align*}
    Consequently, we have:
    \begin{align*}
        \sqrt{n}\sup_{\pi\in \Pi}\left|\frac{1}{n}\sum_{i=1}^n Z_i\mathcal{M}(\pi,A_J(\pi),\{s_{i,t}\}, \{\Gamma_{i,s_{1:T}}\}) \right| \leq \frac{\sqrt{2}}{n^{0.5-\omega}}\left(|\mathcal{S}|^{2T}\max_{i,s_{1:T}}\left\|\Gamma_{i,s_{1:T}}\right\|_{\infty}\right).
    \end{align*}
    Since $\Gamma_{i,s_{1:T}}$ is bounded, consequently, $\sqrt{n}\mathbb{E}\left[\sup_{\pi\in \Pi}\left|\frac{1}{n}\sum_{i=1}^n Z_i\mathcal{M}(\pi,A_J(\pi),\{s_{i,t}\}, \{\Gamma_{i,s_{1:T}}\})\right|\right] = O(\frac{1}{n^{0.5-\omega}})$, which then immediately implies $\lim_{n\to \infty}\sqrt{n}\mathbb{E}\left[\sup_{\pi\in \Pi}\left|\frac{1}{n}\sum_{i=1}^n Z_i\mathcal{M}(\pi,A_J(\pi),\{s_{i,t}\}, \{\Gamma_{i,s_{1:T}}\})\right|\right] = 0$.
    \item $\lim_{n\to \infty} \sqrt{n}\mathbb{E}\left[\sup_{\pi\in \Pi}\left|\frac{1}{n}\sum_{i=1}^n Z_i\mathcal{M}(A_J(\pi),A_{\underline{J}}(\pi),\{s_{i,t}\}, \{\Gamma_{i,s_{1:T}}\}) \right|\right] = 0$: \\
    Conditioned on $\{\{s_{i,t}\}, \{\Gamma_{i,s_{1:T}}\}\}_{i=1}^n$, the random variables $Z_i\mathcal{M}(A_{\underline{J}}(\pi),A_{J}(\pi),\{s_{i,t}\}, \{\Gamma_{i,s_{1:T}}\})$ are independent and zero-mean (since $Z_i$'s are Rademacher random variables).
    Further, each $Z_i\mathcal{M}(A_{\underline{J}}(\pi),A_{J}(\pi),\{s_{i,t}\}, \{\Gamma_{i,s_{1:T}}\})$ is bounded between $a_i=-\left|\mathcal{M}(A_{\underline{J}}(\pi),A_{J}(\pi),\{s_{i,t}\}, \{\Gamma_{i,s_{1:T}}\})\right|$ and $b_i=\left|\mathcal{M}(A_{\underline{J}}(\pi),A_{J}(\pi),\{s_{i,t}\}, \{\Gamma_{i,s_{1:T}}\})\right|$.
    
    By the definition, we have: $I_{\Gamma}(A_{\underline{J}}(\pi),A_J(\pi))=\sqrt{\frac{\sum_{i=1}^n|\mathcal{M}(A_{\underline{J}}(\pi),A_{J}(\pi),\{s_{i,t}\}, \{\Gamma_{i,s_{1:T}}\})|^2}{\sup_{\pi^{\alpha},\pi^{\beta}\in \Pi}\sum_{i=1}^n|\mathcal{M}(\pi^{\alpha},\pi^{\beta},\{s_{i,t}\}, \{\Gamma_{i,s_{1:T}}\})|^2}}$.
    Therefore, we have:
    \begin{align*}
        &I_{\Gamma}(A_{\underline{J}}(\pi),A_J(\pi))^2\sup_{\pi^{\alpha},\pi^{\beta}\in \Pi}\sum_{i=1}^n|\mathcal{M}(\pi^{\alpha},\pi^{\beta},\{s_{i,t}\}, \{\Gamma_{i,s_{1:T}}\})|^2 = \sum_{i=1}^n|\mathcal{M}(A_{\underline{J}}(\pi),A_{J}(\pi),\{s_{i,t}\}, \{\Gamma_{i,s_{1:T}}\})|^2.
    \end{align*}
    By Hoeffding's inequality:
    \begin{align*}
        &P\left[\left|\sum_{i=1}^n Z_i\mathcal{M}(A_{\underline{J}}(\pi),A_{J}(\pi),\{s_{i,t}\}, \{\Gamma_{i,s_{1:T}}\})\right| \geq t \right] \leq 2\exp\left(-\frac{2t^2}{\sum_{i=1}^n(b_i-a_i)^2}\right) \\
        &= 2\exp\left(-\frac{t^2}{2\sum_{i=1}^n\left|\mathcal{M}(A_{\underline{J}}(\pi),A_{J}(\pi),\{s_{i,t}\}, \{\Gamma_{i,s_{1:T}}\})\right|^2}\right).
    \end{align*}
    Let $t=a2^{3-\underline{J}}\sqrt{n\left(|\mathcal{S}|^{2T}\max_{i,s_{1:T}}\left\|\Gamma_{i,s_{1:T}}\right\|_{\infty}\right)^2}$, we have:
    \begin{align*}
        & P\left[\left|\frac{1}{\sqrt{n}}\sum_{i=1}^n Z_i\mathcal{M}(A_{\underline{J}}(\pi),A_{J}(\pi),\{s_{i,t}\}, \{\Gamma_{i,s_{1:T}}\})\right| \geq a2^{3-\underline{J}}\sqrt{\left(|\mathcal{S}|^{2T}\max_{i,s_{1:T}}\left\|\Gamma_{i,s_{1:T}}\right\|_{\infty}\right)^2} \right] \\
        &= P\left[\left|\sum_{i=1}^n Z_i\mathcal{M}(A_{\underline{J}}(\pi),A_{J}(\pi),\{s_{i,t}\}, \{\Gamma_{i,s_{1:T}}\})\right| \geq t \right] \\
        &\leq 2\exp\left(-\frac{t^2}{2\sum_{i=1}^n\left|\mathcal{M}(A_{\underline{J}}(\pi),A_{J}(\pi),\{s_{i,t}\}, \{\Gamma_{i,s_{1:T}}\})\right|^2}\right) \\
        &= 2\exp\left(-\frac{t^2}{2\frac{\sum_{i=1}^n\left|\mathcal{M}(A_{\underline{J}}(\pi),A_{J}(\pi),\{s_{i,t}\}, \{\Gamma_{i,s_{1:T}}\})\right|^2}{\sup_{\pi^{\alpha}, \pi^{\beta}}\sum_{i=1}^n|\mathcal{M}(\pi^{\alpha},\pi^{\beta},\{s_{i,t}\}, \{\Gamma_{i,s_{1:T}}\})|^2}\sup_{\pi^{\alpha}, \pi^{\beta}}\sum_{i=1}^n|\mathcal{M}(\pi^{\alpha},\pi^{\beta},\{s_{i,t}\}, \{\Gamma_{i,s_{1:T}}\})|^2}\right) \\
        &\leq 2\exp\left(-\frac{t^2}{2\frac{\sum_{i=1}^n\left|\mathcal{M}(A_{\underline{J}}(\pi),A_{J}(\pi),\{s_{i,t}\}, \{\Gamma_{i,s_{1:T}}\})\right|^2}{\sup_{\pi^{\alpha}, \pi^{\beta}}\sum_{i=1}^n|\mathcal{M}(\pi^{\alpha},\pi^{\beta},\{s_{i,t}\}, \{\Gamma_{i,s_{1:T}}\})|^2}2\left(|\mathcal{S}|^{2T}\max_{i,s_{1:T}}\left\|\Gamma_{i,s_{1:T}}\right\|_{\infty}\right)^2 n}\right) \\
        &= 2\exp\left(-\frac{a^2 4^{3-\underline{J}}}{4\frac{\sum_{i=1}^n\left|\mathcal{M}(A_{\underline{J}}(\pi),A_{J}(\pi),\{s_{i,t}\}, \{\Gamma_{i,s_{1:T}}\})\right|^2}{\sup_{\pi^{\alpha}, \pi^{\beta}}\sum_{i=1}^n|\mathcal{M}(\pi^{\alpha},\pi^{\beta},\{s_{i,t}\}, \{\Gamma_{i,s_{1:T}}\})|^2}}\right) = 2\exp\left(-\frac{a^2 4^{3-\underline{J}}}{4I_{\Gamma}(A_{\underline{J}}(\pi),A_J(\pi))^2}\right) \\
        &\leq 2\exp\left(-\frac{a^2 4^{3-\underline{J}}}{4\left(\sum_{j=\underline{J}}^{J-1}I_{\Gamma}(A_{j}(\pi),A_{j+1}(\pi))\right)^2}\right) \leq 2\exp\left(-\frac{a^2 4^{3-\underline{J}}}{4\left(\sum_{j=\underline{J}}^{J-1}2^{-(j-1)}\right)^2}\right) \\
        &= 2\exp\left(-\frac{a^2 4^{3-\underline{J}}}{4\left(\frac{2^{-(\underline{J}-1)}(1-2^{-J+\underline{J}})}{1-2^{-1}}\right)^2}\right) = 2\exp\left(-\frac{a^2 4^{3-\underline{J}}}{4\left(2^{-\underline{J}+2}(1-2^{-J+\underline{J}})\right)^2}\right) \\
        &\leq 2\exp\left(-\frac{a^2 4^{3-\underline{J}}}{4\left(2^{-\underline{J}+2}\right)^2}\right) = 2\exp\left(-\frac{a^2 4^{3-\underline{J}}}{4^{3-\underline{J}}}\right) = 2\exp(-a^2).
    \end{align*}
    Since this equation holds for any $\pi\in \Pi$, by a union bound, we have:
    \begin{align*}
        & P\left[\sup_{\pi\in\Pi}\left|\frac{1}{\sqrt{n}}\sum_{i=1}^n Z_i\mathcal{M}(A_{\underline{J}}(\pi),A_{J}(\pi),\{s_{i,t}\}, \{\Gamma_{i,s_{1:T}}\})\right| \geq a2^{3-\underline{J}}\sqrt{\left(|\mathcal{S}|^{2T}\max_{i,s_{1:T}}\left\|\Gamma_{i,s_{1:T}}\right\|_{\infty}\right)^2} \right] \\
        &\leq 2\left|\{(A_{\underline{J}}(\pi), A_{J}(\pi)) | \pi\in \Pi\}\right|\exp\left(-a^2\right) \leq  2N_{I_{\Gamma}}(2^{-J}, \Pi, \{\{s_{1,t}\}, \cdots, \{s_{n,t}\}\}\})\exp(-a^2) \\
        &\leq 2N_{H}(2^{-2J},\Pi)\exp(-a^2) \leq 2C\exp(D2^{2J\omega})\exp(-a^2) \leq 2C\exp(D2^{2\omega(1-\omega)\log_2(n)} -a^2),
    \end{align*}
    where the second inequality follows from Property 4 in Step 1, the third inequality follows from Lemma \ref{lem:dist}, the fourth inequality follows from Assumption \ref{asp:covering_number} and the last inequality follows from $J=\lceil (1-\omega)\log_2(n)\rceil \leq (1-\omega)\log_2(n)+1$ (and the term $2^{2\omega}$ is absorbed into the constant $D$).
    Next, set $a=\frac{2^{\underline{J}}}{\sqrt{\log n\left(|\mathcal{S}|^{2T}\max_{i,s_{1:T}}\left\|\Gamma_{i,s_{1:T}}\right\|_{\infty}\right)^2 }}$, we have:
    \begin{align*}
        & P\left[\sup_{\pi\in\Pi}\left|\frac{1}{\sqrt{n}}\sum_{i=1}^n Z_i\mathcal{M}(A_{\underline{J}}(\pi),A_{J}(\pi),\{s_{i,t}\}, \{\Gamma_{i,s_{1:T}}\})\right| \geq \frac{8}{\sqrt{\log n}} \right] \leq 2C\exp(D2^{2\omega(1-\omega)\log_2(n)} -a^2) \\
        &= 2C\exp\left(D2^{2\omega(1-\omega)\log_2(n)} - \frac{2^{(1-\omega)\log_2(n)}}{\log n\left(|\mathcal{S}|^{2T}\max_{i,s_{1:T}}\left\|\Gamma_{i,s_{1:T}}\right\|_{\infty}\right)^2 }\right) \\
        &= 2C\exp\left(D2^{2\omega(1-\omega)\log_2(n)} - \frac{2^{(1-2\omega+2\omega)(1-\omega)\log_2(n)}}{ U^2\log n}\right) \\
        &= 2C\exp\left(2^{2\omega(1-\omega)\log_2(n)}\left(D - \frac{2^{(1-2\omega)(1-\omega)\log_2(n)}}{U^2\log n}\right)\right) = 2C\exp\left(-n^{2\omega(1-\omega)}\left(\frac{n^{(1-2\omega)(1-\omega)}}{ U^2\log n}-D\right)\right),
    \end{align*}
    where $U=\left(|\mathcal{S}|^{2T}\max_{i,s_{1:T}}\left\|\Gamma_{i,s_{1:T}}\right\|_{\infty}\right)$.
    Since $\omega<\frac{1}{2}$ by Assumption \ref{asp:covering_number}, $\lim_{n\to \infty}\frac{n^{(1-2\omega)(1-\omega)}}{ U^2\log n}=\infty$.
    This mean for all large $n$, with probability at least $1-2C\exp(-n^{2\omega(1-\omega)})$, $\sup_{\pi\in\Pi}\left|\frac{1}{\sqrt{n}}\sum_{i=1}^n Z_i\mathcal{M}(A_{\underline{J}}(\pi),A_{J}(\pi),\{s_{i,t}\}, \{\Gamma_{i,s_{1:T}}\})\right| \leq \frac{8}{\sqrt{\log n}}$, therefore immediately implying: $\lim_{n\to \infty} \sqrt{n}\mathbb{E}\left[\sup_{\pi\in \Pi}\left|\frac{1}{n}\sum_{i=1}^n Z_i\mathcal{M}(A_{\underline{J}}(\pi),A_{J}(\pi),\{s_{i,t}\}, \{\Gamma_{i,s_{1:T}}\})\right|\right] = 0$.
\end{enumerate}

\paragraph{Step 3: Chaining with concentration inequalities in the effective regime.}

By expanding the Rademacher complexity using the approximation policy profiles, we can show:
\begin{align}
\label{eq:rademacher}
\begin{aligned}
        &\mathcal{R}_n(\Pi^D) = \mathbb{E}\left[\sup_{\pi^{\alpha},\pi^{\beta}\in \Pi}\frac{1}{n} \left|\sum_{i=1}^n Z_i\mathcal{M}(\pi^{\alpha},\pi^{\beta},\{s_{i,t}\}, \{\Gamma_{i,s_{1:T}}\})\right|\right], \\
        \leq& 2\mathbb{E}\left[\sup_{\pi\in \Pi}\frac{1}{n} \left|\sum_{i=1}^n Z_i\mathcal{M}(\pi,A_J(\pi),\{s_{i,t}\}, \{\Gamma_{i,s_{1:T}}\})\right|\right] + 2\mathbb{E}\left[\sup_{\pi\in \Pi}\frac{1}{n} \left|\sum_{i=1}^n Z_i\mathcal{M}(A_J(\pi),A_{\underline{J}}(\pi),\{s_{i,t}\}, \{\Gamma_{i,s_{1:T}}\})\right|\right] \\
        &+ 2\mathbb{E}\left[\sup_{\pi\in \Pi}\frac{1}{n} \left|\sum_{i=1}^n Z_i\sum_{s_{1:T}} \left\langle \sum_{j=1}^{\underline{J}}\left(A_j(\pi)(s_{1:T})-A_{j-1}(\pi)(s_{1:T})\right),\Gamma_{i,s_{1:T}}\right\rangle\right|\right] \\
        =& 2\mathbb{E}\left[\sup_{\pi\in \Pi}\frac{1}{n} \left|\sum_{i=1}^n Z_i\sum_{s_{1:T}} \left\langle \sum_{j=1}^{\underline{J}}\left(A_j(\pi)(s_{1:T})-A_{j-1}(\pi)(s_{1:T})\right),\Gamma_{i,s_{1:T}}\right\rangle\right|\right] + o(\frac{1}{\sqrt{n}}).
\end{aligned}
\end{align}
Consequently, it now remains to bound $\mathbb{E}\left[\sup_{\pi\in \Pi}\frac{1}{n} \left|\sum_{i=1}^n Z_i\sum_{s_{1:T}} \left\langle \sum_{j=1}^{\underline{J}}\left(A_j(\pi)(s_{1:T})-A_{j-1}(\pi)(s_{1:T})\right),\Gamma_{i,s_{1:T}}\right\rangle\right|\right]$.
For each $j\in \{1,\cdots, \underline{J}\}$, setting $t_j=a_j2^{2-j}\sqrt{\sup_{\pi^{\alpha}, \pi^{\beta}}\sum_{i=1}^n|\mathcal{M}(\pi^{\alpha},\pi^{\beta},\{s_{i,t}\}, \{\Gamma_{i,s_{1:T}}\})|^2}$ and applying Hoeffding's inequality:
\begin{align*}
        & P\left[\left|\frac{1}{\sqrt{n}}\sum_{i=1}^n Z_i\mathcal{M}(A_{j}(\pi),A_{j-1}(\pi),\{s_{i,t}\}, \{\Gamma_{i,s_{1:T}}\})\right| \geq a_j2^{2-j}\sqrt{\frac{\sup_{\pi^{\alpha}, \pi^{\beta}}\sum_{i=1}^n|\mathcal{M}(\pi^{\alpha},\pi^{\beta},\{s_{i,t}\}, \{\Gamma_{i,s_{1:T}}\})|^2}{n}}\right] \\
        &= P\left[\left|\sum_{i=1}^n Z_i\mathcal{M}(A_{j}(\pi),A_{j-1}(\pi),\{s_{i,t}\}, \{\Gamma_{i,s_{1:T}}\})\right| \geq t_j \right] \\
        &\leq 2\exp\left(-\frac{t_j^2}{2\sum_{i=1}^n\left|\mathcal{M}(A_{j}(\pi),A_{j-1}(\pi),\{s_{i,t}\}, \{\Gamma_{i,s_{1:T}}\})\right|^2}\right) \\
        &= 2\exp\left(-\frac{a_j^2 4^{2-j}}{2\frac{\sum_{i=1}^n\left|\mathcal{M}(A_{j}(\pi),A_{j-1}(\pi),\{s_{i,t}\}, \{\Gamma_{i,s_{1:T}}\})\right|^2}{\sup_{\pi^{\alpha}, \pi^{\beta}}\sum_{i=1}^n|\mathcal{M}(\pi^{\alpha},\pi^{\beta},\{s_{i,t}\}, \{\Gamma_{i,s_{1:T}}\})|^2}}\right) = 2\exp\left(-\frac{a_j^2 4^{2-j}}{2I_{\Gamma}(A_j(\pi),A_{j-1}(\pi))^2}\right) \\
        &\leq 2\exp\left(-\frac{a_j^2 4^{2-j}}{2\cdot 4^{-(j-2)}}\right) = 2\exp\left(-\frac{a_j^2}{2}\right),
\end{align*}
where the last inequality follows from Property 3 in Step 1.
For the rest of this step, we denote for notational convenience $M(\Pi)\triangleq \sup_{\pi^{\alpha}, \pi^{\beta}}\sum_{i=1}^n|\mathcal{M}(\pi^{\alpha},\pi^{\beta},\{s_{i,t}\}, \{\Gamma_{i,s_{1:T}}\})|^2$, as this term will be repeatedly used.
Setting $a_j^2=2\log\left(\frac{2j^2}{\delta}N_H(4^{-j},\Pi)\right)$, we then apply a union bound to obtain:
\begin{align*}
    & P\left[\sup_{\pi\in\Pi}\left|\frac{1}{\sqrt{n}}\sum_{i=1}^n Z_i\mathcal{M}(A_{j}(\pi),A_{j-1}(\pi),\{s_{i,t}\}, \{\Gamma_{i,s_{1:T}}\})\right| \geq a_j2^{2-j}\sqrt{\frac{M(\Pi)}{n}}\right] \\
    &\leq 2\left|\{(A_{j}(\pi), A_{j-1}(\pi)) | \pi\in \Pi\}\right|\exp\left(-\frac{a_j^2}{2}\right) \leq  2N_{I_{\Gamma}}(2^{-j}, \Pi, \{\{s_{1,t}\}, \cdots, \{s_{n,t}\}\}\})\exp\left(-\frac{a_j^2}{2}\right) \\
    &\leq 2N_{H}(4^{-j},\Pi)\exp\left(-\frac{a_j^2}{2}\right) = 2N_{H}(4^{-j},\Pi)\exp\left(-\log\left(\frac{2j^2}{\delta}N_H(4^{-j},\Pi)\right)\right) = \frac{\delta}{j^2}.
\end{align*}
Consequently, by a further union bound:
\begin{align*}
    & P\left[\sup_{\pi\in\Pi}\left|\frac{1}{\sqrt{n}}\sum_{i=1}^n Z_i\sum_{s_{1:T}} \left\langle \sum_{j=1}^{\underline{J}}\left(A_{j}(\pi)(s_{1:T})-A_{j-1}(\pi)(s_{1:T})\right),\Gamma_{i,s_{1:T}}\right\rangle\right| \geq \sum_{j=1}^{\underline{J}}a_j2^{2-j}\sqrt{\frac{M(\Pi)}{n}}\right] \\
    &\leq P\left[\sum_{j=1}^{\underline{J}}\sup_{\pi\in\Pi}\left|\frac{1}{\sqrt{n}}\sum_{i=1}^n Z_i\mathcal{M}(A_{j}(\pi),A_{j-1}(\pi),\{s_{i,t}\}, \{\Gamma_{i,s_{1:T}}\})\right| \geq \sum_{j=1}^{\underline{J}}a_j2^{2-j}\sqrt{\frac{M(\Pi)}{n}}\right] \\
    &\leq \sum_{j=1}^{\underline{J}}P\left[\sup_{\pi\in\Pi}\left|\frac{1}{\sqrt{n}}\sum_{i=1}^n Z_i\mathcal{M}(A_{j}(\pi),A_{j-1}(\pi),\{s_{i,t}\}, \{\Gamma_{i,s_{1:T}}\})\right| \geq a_j2^{2-j}\sqrt{\frac{M(\Pi)}{n}}\right] \leq \sum_{j=1}^{\underline{J}}\frac{\delta}{j^2} < \sum_{j=1}^{\infty}\frac{\delta}{j^2} < 1.7\delta.
\end{align*}
Take $\delta_k=\frac{1}{2^k}$ and apply the above bound to each $\delta_k$ yields that with probability at least $1-\frac{1.7}{2^k}$,
\begin{align*}
    &\sup_{\pi\in\Pi}\left|\frac{1}{\sqrt{n}}\sum_{i=1}^n Z_i\sum_{s_{1:T}} \left\langle \sum_{j=1}^{\underline{J}}\left(A_{j}(\pi)(s_{1:T})-A_{j-1}(\pi)(s_{1:T})\right),\Gamma_{i,s_{1:T}}\right\rangle\right| \leq \sum_{j=1}^{\underline{J}}a_j2^{2-j}\sqrt{\frac{M(\Pi)}{n}} \\
    &= 4\sqrt{2}\sum_{j=1}^{\underline{J}}\sqrt{\log\left(2^{k+1}j^2N_H(4^{-j},\Pi)\right)}2^{-j}\sqrt{\frac{M(\Pi)}{n}} = 4\sqrt{2}\sqrt{\frac{M(\Pi)}{n}}\sum_{j=1}^{\underline{J}}2^{-j}\sqrt{\log\left(2^{k+1}j^2N_H(4^{-j},\Pi)\right)} \\
    &\leq 4\sqrt{2}\sqrt{\frac{M(\Pi)}{n}}\sum_{j=1}^{\underline{J}}2^{-j}\left(\sqrt{k+1}+\sqrt{2\log j}+\sqrt{\log N_H(4^{-j},\Pi)}\right) \\
    &\leq 4\sqrt{2}\sqrt{\frac{M(\Pi)}{n}}\left(\sqrt{k+1}\sum_{j=1}^{\infty}2^{-j}\left(1+\sqrt{2\log j}\right)+\sum_{j=1}^{\underline{J}}2^{-j}\sqrt{\log N_H(4^{-j},\Pi)}\right) \\
    &\leq 4\sqrt{2}\sqrt{\frac{M(\Pi)}{n}}\left(\sqrt{k+1}\sum_{j=1}^{\infty}\frac{2j}{2^{j}}+\frac{1}{2}\sum_{j=1}^{J}2^{-j}\sqrt{\log N_H(4^{-j},\Pi)}\right) \\
    &= 4\sqrt{2}\sqrt{\frac{M(\Pi)}{n}}\left(\sqrt{k+1}\frac{2\cdot2^{-1}}{(1-2^{-1})^2}+\frac{1}{2}\sum_{j=1}^{J}2^{-j}\left(\sqrt{\log N_H(4^{-j},\Pi)}+\sqrt{\log N_H(1,\Pi)}\right)\right) \\
    &= 4\sqrt{2}\sqrt{\frac{M(\Pi)}{n}}\left(4\sqrt{k+1}+\sum_{j=0}^{J}2^{-j-1}\sqrt{\log N_H(4^{-j},\Pi)}\right) < 4\sqrt{2}\sqrt{\frac{M(\Pi)}{n}}\left(4\sqrt{k+1}+\int_{0}^{1}\sqrt{\log N_H(\epsilon^2,\Pi)}d\epsilon\right) \\
    &= 4\sqrt{2}\sqrt{\frac{\sup_{\pi^{\alpha}, \pi^{\beta}}\sum_{i=1}^n|\mathcal{M}(\pi^{\alpha},\pi^{\beta},\{s_{i,t}\}, \{\Gamma_{i,s_{1:T}}\})|^2}{n}}\left(4\sqrt{k+1}+\kappa(\Pi)\right),
\end{align*}
where the last inequality follows from setting $\epsilon=2^{-j}$ and upper bounding the sum using the integral.
Consequently, for each $k=0,1,\cdots,$ we have:
\begin{align*}
    &P\left[\sup_{\pi\in\Pi}\left|\frac{1}{\sqrt{n}}\sum_{i=1}^n Z_i\sum_{s_{1:T}} \left\langle \sum_{j=1}^{\underline{J}}\left(A_{j}(\pi)(s_{1:T})-A_{j-1}(\pi)(s_{1:T})\right),\Gamma_{i,s_{1:T}}\right\rangle\right| \right. \\
    &\left. \geq 4\sqrt{2}\sqrt{\frac{\sup_{\pi^{\alpha}, \pi^{\beta}}\sum_{i=1}^n|\mathcal{M}(\pi^{\alpha},\pi^{\beta},\{s_{i,t}\}, \{\Gamma_{i,s_{1:T}}\})|^2}{n}}\left(4\sqrt{k+1}+\kappa(\Pi)\right)\right] \leq \frac{1.7}{2^k}
\end{align*}
We next turn the probability bound given in this equation into a bound on its (conditional) expectation.
Specifically, define the (non-negative) random variable $R=\sup_{\pi\in\Pi}\left|\frac{1}{\sqrt{n}}\sum_{i=1}^n Z_i\sum_{s_{1:T}} \left\langle \sum_{j=1}^{\underline{J}}\left(A_{j}(\pi)(s_{1:T})-A_{j-1}(\pi)(s_{1:T})\right),\Gamma_{i,s_{1:T}}\right\rangle\right|$ and let $F_R(\cdot)$ be its cumulative distribution function (conditioned on $\{\{s_{i,t}\}, \{\Gamma_{i,s_{1:T}}\}\}_{i=1}^n$).
Per its definition, we have:
\begin{align*}
    1 - F_R\left(4\sqrt{2}\sqrt{\frac{\sup_{\pi^{\alpha}, \pi^{\beta}}\sum_{i=1}^n|\mathcal{M}(\pi^{\alpha},\pi^{\beta},\{s_{i,t}\}, \{\Gamma_{i,s_{1:T}}\})|^2}{n}}\left(4\sqrt{k+1}+\kappa(\Pi)\right)\right)\leq \frac{1.7}{2^k}.
\end{align*}
Consequently, we have:
\begin{align*}
    &\mathbb{E}\left[R ~|~ \{\{s_{i,t}\}, \{\Gamma_{i,s_{1:T}}\}\}_{i=1}^n\right] = \int_{0}^{\infty} (1-F_R(r))dr \\
    &\leq \sum_{k=0}^{\infty}\frac{1.7}{2^k}4\sqrt{2}\sqrt{\frac{\sup_{\pi^{\alpha}, \pi^{\beta}}\sum_{i=1}^n|\mathcal{M}(\pi^{\alpha},\pi^{\beta},\{s_{i,t}\}, \{\Gamma_{i,s_{1:T}}\})|^2}{n}}\left(4\sqrt{k+1}+\kappa(\Pi)\right) \\
    &= 6.8\sqrt{2}\sqrt{\frac{\sup_{\pi^{\alpha}, \pi^{\beta}}\sum_{i=1}^n|\mathcal{M}(\pi^{\alpha},\pi^{\beta},\{s_{i,t}\}, \{\Gamma_{i,s_{1:T}}\})|^2}{n}}\left(\sum_{k=0}^{\infty}\frac{1}{2^k}4\sqrt{k+1}+\sum_{k=0}^{\infty}\frac{1}{2^k}\kappa(\Pi)\right) \\
    &\leq 6.8\sqrt{2}\sqrt{\frac{\sup_{\pi^{\alpha}, \pi^{\beta}}\sum_{i=1}^n|\mathcal{M}(\pi^{\alpha},\pi^{\beta},\{s_{i,t}\}, \{\Gamma_{i,s_{1:T}}\})|^2}{n}}\left(\sum_{k=0}^{\infty}\frac{4(k+1)}{2^k}+2\kappa(\Pi)\right) \\
    &= 6.8\sqrt{2}\sqrt{\frac{\sup_{\pi^{\alpha}, \pi^{\beta}}\sum_{i=1}^n|\mathcal{M}(\pi^{\alpha},\pi^{\beta},\{s_{i,t}\}, \{\Gamma_{i,s_{1:T}}\})|^2}{n}}\left(16+2\kappa(\Pi)\right).
\end{align*}
Taking expectation with respect to $\{\{s_{i,t}\}, \{\Gamma_{i,s_{1:T}}\}\}_{i=1}^n$, we obtain:
\begin{align}
\begin{aligned}
\label{eq:under_J}
    &\mathbb{E}\left[\sup_{\pi\in\Pi}\left|\frac{1}{\sqrt{n}}\sum_{i=1}^n Z_i\sum_{s_{1:T}}\left\langle \sum_{j=1}^{\underline{J}}\left(A_{j}(\pi)(s_{1:T})-A_{j-1}(\pi)(s_{1:T})\right),\Gamma_{i,s_{1:T}}\right\rangle\right|\right] \\
    &\leq 6.8\sqrt{2}\left(16+2\kappa(\Pi)\right)\mathbb{E}\left[\sqrt{\frac{\sup_{\pi^{\alpha}, \pi^{\beta}}\sum_{i=1}^n|\mathcal{M}(\pi^{\alpha},\pi^{\beta},\{s_{i,t}\}, \{\Gamma_{i,s_{1:T}}\})|^2}{n}}\right] \\
    &\leq 13.6\sqrt{2}\left(8+\kappa(\Pi)\right)\sqrt{\mathbb{E}\left[\frac{\sup_{\pi^{\alpha}, \pi^{\beta}}\sum_{i=1}^n|\mathcal{M}(\pi^{\alpha},\pi^{\beta},\{s_{i,t}\}, \{\Gamma_{i,s_{1:T}}\})|^2}{n}\right]}
\end{aligned}
\end{align}

\paragraph{Step 4: Refining the lower range bound using Talagrand's inequality.}
To obtain a bound on $\mathbb{E}\!\left[\!\frac{\sup_{\pi^{\alpha}, \pi^{\beta}}\sum_{i=1}^n|\mathcal{M}(\pi^{\alpha},\pi^{\beta},\{s_{i,t}\}, \{\Gamma_{i,s_{1:T}}\})|^2}{n}\right]$ , we use the following version of Talagrand's concentration inequality in \cite{gine2006concentration, zhou2018offline}:
\begin{lemma}
\label{lem:talagrand}
Let $X_1, \cdots, X_n$ be independent $\mathcal{X}$-valued random variables and $\mathcal{F}$ be a class of functions where $\sup_{x\in \mathcal{X}} |f(x)| \leq U$ for some $U>0$, and let $Z_i$ be {\bf iid} Rademacher random variables: $P(Z_i=1)=P(Z_i=-1)=\frac{1}{2}$.
We have:
\begin{align*}
    \mathbb{E}\left[\sup_{f\in \mathcal{F}}\sum_{i=1}^n f^2(X_i)\right] \leq n\sup_{f\in \mathcal{F}}\mathbb{E}[f^2(X_i)] + 8U\mathbb{E}\left[\sup_{f\in \mathcal{F}}\left|\sum_{i=1}^n Z_i f(X_i)\right|\right]
\end{align*}
\end{lemma}
We apply Lemma \ref{lem:talagrand} to the current context: we identify $X_i$ in Lemma \ref{lem:talagrand} with $(\{s_{i,t}\}, \{\Gamma_{i,s_{1:T}}\})$ here and $f(\{s_{i,t}\}, \{\Gamma_{i,s_{1:T}}\})=\mathcal{M}(\pi^{\alpha},\pi^{\beta},\{s_{i,t}\}, \{\Gamma_{i,s_{1:T}}\})$.
Since $\Gamma_{i,s_{1:T}}$ is bounded, for some constant $U$, $\forall \pi^{\alpha}, \pi^{\beta}\in \Pi, ~|f(\{s_{i,t}\}, \{\Gamma_{i,s_{1:T}}\})|\leq \sum_{s_{1:T}} \|\pi^{\alpha}(s_{1:T})-\pi^{\beta}(s_{1:T})\|_2 \|\Gamma_{i,s_{1:T}}\|_2 \leq \sqrt{2}\sum_{s_{1:T}} \|\Gamma_{i,s_{1:T}}\|_2 \leq U$.
Consequently, we have:
\begin{align*}
    &\mathbb{E}\left[\sup_{\pi^{\alpha}, \pi^{\beta}}\sum_{i=1}^n\left(\mathcal{M}(\pi^{\alpha},\pi^{\beta},\{s_{i,t}\}, \{\Gamma_{i,s_{1:T}}\})\right)^2\right] \\
    &\leq n \sup_{\pi^{\alpha}, \pi^{\beta}}\mathbb{E}\left[\left(\mathcal{M}(\pi^{\alpha},\pi^{\beta},\{s_{i,t}\}, \{\Gamma_{i,s_{1:T}}\})\right)^2\right] + 8U\mathbb{E}\left[\sup_{\pi^{\alpha},\pi^{\beta}}\sum_{i=1}^n \left|Z_i\left(\mathcal{M}(\pi^{\alpha},\pi^{\beta},\{s_{i,t}\}, \{\Gamma_{i,s_{1:T}}\})\right)\right|\right].
\end{align*}
Dividing both sides by $n$ then yields:
\begin{align}
\begin{aligned}
\label{eq:talagrand}
    &\mathbb{E}\left[\frac{\sup_{\pi^{\alpha}, \pi^{\beta}}\sum_{i=1}^n\left(\mathcal{M}(\pi^{\alpha},\pi^{\beta},\{s_{i,t}\}, \{\Gamma_{i,s_{1:T}}\})\right)^2}{n}\right] \\
    &\leq \sup_{\pi^{\alpha}, \pi^{\beta}}\mathbb{E}\left[\left(\mathcal{M}(\pi^{\alpha},\pi^{\beta},\{s_{i,t}\}, \{\Gamma_{i,s_{1:T}}\})\right)^2\right] + 8U\mathbb{E}\left[\sup_{\pi^{\alpha},\pi^{\beta}}\frac{1}{n}\sum_{i=1}^n \left|Z_i\left(\mathcal{M}(\pi^{\alpha},\pi^{\beta},\{s_{i,t}\}, \{\Gamma_{i,s_{1:T}}\})\right)\right|\right] \\
    &= \sup_{\pi^{\alpha}, \pi^{\beta}}\mathbb{E}\left[\left(\mathcal{M}(\pi^{\alpha},\pi^{\beta},\{s_{i,t}\}, \{\Gamma_{i,s_{1:T}}\})\right)^2\right] + 8U\mathcal{R}_n(\Pi^D).
\end{aligned}
\end{align}
Therefore, by combining Equation (\ref{eq:under_J}) with Equation (\ref{eq:talagrand}), we have:
\begin{align*}
    &\mathbb{E}\left[\sup_{\pi\in\Pi}\left|\frac{1}{\sqrt{n}}\sum_{i=1}^n Z_i\sum_{s_{1:T}} \left\langle \sum_{j=1}^{\underline{J}}\left(A_{j}(\pi)(s_{1:T})-A_{j-1}(\pi)(s_{1:T})\right),\Gamma_{i,s_{1:T}}\right\rangle\right|\right] \\
    &\leq 13.6\sqrt{2}\left(8+\kappa(\Pi)\right)\sqrt{\sup_{\pi^{\alpha}, \pi^{\beta}}\mathbb{E}\left[\left(\mathcal{M}(\pi^{\alpha},\pi^{\beta},\{s_{i,t}\}, \{\Gamma_{i,s_{1:T}}\})\right)^2\right] + 8U\mathcal{R}_n(\Pi^D)}.
\end{align*}
Finally, combining Equation (\ref{eq:rademacher}), we have:
\begin{align*}
    &\sqrt{n}\mathcal{R}_n(\Pi^D) = \mathbb{E}\left[\sup_{\pi^{\alpha},\pi^{\beta}\in \Pi}\frac{1}{\sqrt{n}} \left|\sum_{i=1}^n Z_i\mathcal{M}(\pi^{\alpha},\pi^{\beta},\{s_{i,t}\}, \{\Gamma_{i,s_{1:T}}\})\right|\right] \\
    &\leq 2\mathbb{E}\left[\sup_{\pi\in \Pi}\frac{1}{\sqrt{n}} \left|\sum_{i=1}^n Z_i\sum_{s_{1:T}} \left\langle \sum_{j=1}^{\underline{J}}\left(A_{j}(\pi)(s_{1:T})-A_{j-1}(\pi)(s_{1:T})\right),\Gamma_{i,s_{1:T}}\right\rangle\right|\right] + o(1) \\
    &\leq 27.2\sqrt{2}\left(8+\kappa(\Pi)\right)\sqrt{\sup_{\pi^{\alpha}, \pi^{\beta}}\mathbb{E}\left[\left(\mathcal{M}(\pi^{\alpha},\pi^{\beta},\{s_{i,t}\}, \{\Gamma_{i,s_{1:T}}\})\right)^2\right] + 8U\mathcal{R}_n(\Pi^D)} + o(1).
\end{align*}
Dividing both sided of the above inequality by $\sqrt{n}$ yields:
\begin{align}
\begin{aligned}
\label{eq:R_n}
    \mathcal{R}_n(\Pi^D) &\leq 27.2\sqrt{2}\left(8+\kappa(\Pi)\right)\sqrt{\frac{\sup_{\pi^{\alpha}, \pi^{\beta}}\mathbb{E}\left[\left(\mathcal{M}(\pi^{\alpha},\pi^{\beta},\{s_{i,t}\}, \{\Gamma_{i,s_{1:T}}\})\right)^2\right] + 8U\mathcal{R}_n(\Pi^D)}{n}} + o(\frac{1}{\sqrt{n}}) \\
    &\leq 27.2\sqrt{2}\left(8+\kappa(\Pi)\right)\left(\sqrt{\frac{\sup_{\pi^{\alpha}, \pi^{\beta}}\mathbb{E}\left[\left(\mathcal{M}(\pi^{\alpha},\pi^{\beta},\{s_{i,t}\}, \{\Gamma_{i,s_{1:T}}\})\right)^2\right]}{n}} + \sqrt{\frac{8U\mathcal{R}_n(\Pi^D)}{n}}\right) + o(\frac{1}{\sqrt{n}}).
\end{aligned}
\end{align}
The above equation immediately implies $\mathcal{R}_n(\Pi)=O(\sqrt{\frac{1}{n}})+O(\sqrt{\frac{\mathcal{R}_n(\Pi)}{n}})$, which one can solve to obtain $\mathcal{R}_n(\Pi)=O(\sqrt{\frac{1}{n}})$.
Plugging it into Equation (\ref{eq:R_n}) then results:
\begin{align*}
    \mathcal{R}_n(\Pi^D) &\leq 27.2\sqrt{2}\left(8+\kappa(\Pi)\right)\left(\sqrt{\frac{\sup_{\pi^{\alpha}, \pi^{\beta}}\mathbb{E}\left[\left(\mathcal{M}(\pi^{\alpha},\pi^{\beta},\{s_{i,t}\}, \{\Gamma_{i,s_{1:T}}\})\right)^2\right]}{n}} + \sqrt{\frac{O(\sqrt{\frac{1}{n}})}{n}}\right) + o(\frac{1}{\sqrt{n}}) \\
    & \leq 27.2\sqrt{2}\left(8+\kappa(\Pi)\right)\sqrt{\frac{\sup_{\pi^{\alpha}, \pi^{\beta}}\mathbb{E}\left[\left(\mathcal{M}(\pi^{\alpha},\pi^{\beta},\{s_{i,t}\}, \{\Gamma_{i,s_{1:T}}\})\right)^2\right]}{n}} + o(\frac{1}{\sqrt{n}}) \\
    & \leq O\left(\kappa(\Pi)\sqrt{\frac{\sup_{\pi^{\alpha}, \pi^{\beta}}\mathbb{E}\left[\left(\mathcal{M}(\pi^{\alpha},\pi^{\beta},\{s_{i,t}\}, \{\Gamma_{i,s_{1:T}}\})\right)^2\right]}{n}}\right) + o(\frac{1}{\sqrt{n}})
\end{align*}
\end{proof}

\subsection{Proof of Lemma \ref{lem:dist}}
\begin{proof}
$\forall \pi_1,\pi_2,\pi_3\in \Pi$:
\begin{align*}
    &I_{\Gamma}(\pi_1,\pi_2)^2 = \frac{\sum_{i=1}^n|\sum_{s_{1:T}}\left\langle\pi_1(s_{1:T})-\pi_3(s_{1:T})+\pi_3(s_{1:T})-\pi_2(s_{1:T}),\Gamma_{i,s_{1:T}}\right\rangle|^2}{\sup_{\pi^{\alpha},\pi^{\beta}}\sum_{i=1}^n|\mathcal{M}(\pi_{\alpha},\pi_{\beta},\{s_{i,t}\}, \{\Gamma_{i,s_{1:T}}\})|^2} \\
    =& \frac{\sum_{i=1}^n|\mathcal{M}(\pi_1,\pi_3,\{s_{i,t}\}, \{\Gamma_{i,s_{1:T}}\}) + \mathcal{M}(\pi_3,\pi_2,\{s_{i,t}\}, \{\Gamma_{i,s_{1:T}}\})|^2}{\sup_{\pi^{\alpha},\pi^{\beta}}\sum_{i=1}^n|\mathcal{M}(\pi_{\alpha},\pi_{\beta},\{s_{i,t}\}, \{\Gamma_{i,s_{1:T}}\})|^2} \\
    \leq& \frac{\sum_{i=1}^n\left(|\mathcal{M}(\pi_1,\pi_3,\{s_{i,t}\}, \{\Gamma_{i,s_{1:T}}\})|^2 + |\mathcal{M}(\pi_3,\pi_2,\{s_{i,t}\}, \{\Gamma_{i,s_{1:T}}\})|^2\right)}{\sup_{\pi^{\alpha},\pi^{\beta}}\sum_{i=1}^n|\mathcal{M}(\pi_{\alpha},\pi_{\beta},\{s_{i,t}\}, \{\Gamma_{i,s_{1:T}}\})|^2} \\
    &+ \frac{2\sum_{i=1}^n\left|\mathcal{M}(\pi_1,\pi_3,\{s_{i,t}\}, \{\Gamma_{i,s_{1:T}}\})\right|\left|\mathcal{M}(\pi_3,\pi_2,\{s_{i,t}\}, \{\Gamma_{i,s_{1:T}}\})\right|}{\sup_{\pi^{\alpha},\pi^{\beta}}\sum_{i=1}^n|\mathcal{M}(\pi_{\alpha},\pi_{\beta},\{s_{i,t}\}, \{\Gamma_{i,s_{1:T}}\})|^2} \\
    \leq& \frac{\sum_{i=1}^n\left(|\mathcal{M}(\pi_1,\pi_3,\{s_{i,t}\}, \{\Gamma_{i,s_{1:T}}\})|^2 + |\mathcal{M}(\pi_3,\pi_2,\{s_{i,t}\}, \{\Gamma_{i,s_{1:T}}\})|^2\right)}{\sup_{\pi^{\alpha},\pi^{\beta}}\sum_{i=1}^n|\mathcal{M}(\pi_{\alpha},\pi_{\beta},\{s_{i,t}\}, \{\Gamma_{i,s_{1:T}}\})|^2} \\
    &+ \frac{2\sqrt{\left(\sum_{i=1}^n\left|\mathcal{M}(\pi_1,\pi_3,\{s_{i,t}\}, \{\Gamma_{i,s_{1:T}}\})\right|^2\right)\left(\sum_{i=1}^n\left|\mathcal{M}(\pi_3,\pi_2,\{s_{i,t}\}, \{\Gamma_{i,s_{1:T}}\})\right|^2\right)}}{\sup_{\pi^{\alpha},\pi^{\beta}}\sum_{i=1}^n|\mathcal{M}(\pi_{\alpha},\pi_{\beta},\{s_{i,t}\}, \{\Gamma_{i,s_{1:T}}\})|^2} \\
    =& \left(\sqrt{\frac{\sum_{i=1}^n|\mathcal{M}(\pi_1,\pi_3,\{s_{i,t}\}, \{\Gamma_{i,s_{1:T}}\})|^2}{\sup_{\pi^{\alpha},\pi^{\beta}}\sum_{i=1}^n|\mathcal{M}(\pi_{\alpha},\pi_{\beta},\{s_{i,t}\}, \{\Gamma_{i,s_{1:T}}\})|^2}} + \sqrt{\frac{\sum_{i=1}^n|\mathcal{M}(\pi_3,\pi_2,\{s_{i,t}\}, \{\Gamma_{i,s_{1:T}}\})|^2}{\sup_{\pi^{\alpha},\pi^{\beta}}\sum_{i=1}^n|\mathcal{M}(\pi_{\alpha},\pi_{\beta},\{s_{i,t}\}, \{\Gamma_{i,s_{1:T}}\})|^2}}\right)^2 \\
    =& \left(I_{\Gamma}(\pi_1,\pi_3) + I_{\Gamma}(\pi_3,\pi_2)\right)^2.
\end{align*}
Thus, we get $I_{\Gamma}(\pi_1,\pi_2) \leq I_{\Gamma}(\pi_1,\pi_3) + I_{\Gamma}(\pi_3,\pi_2)$.

Next, to prove the second statement, let $K=N_H(\epsilon^2,\Pi)$.
Without loss of generality, we can assume $K<\infty$, otherwise, the above inequality automatically holds.
Fix any $n$ state trajectories $\{\{s_{1,t}\}, \cdots, \{s_{n,t}\}\}\}$.
Denote by $\{\pi_1,\cdots,\pi_K\}$ the set of $K$ policy profiles that $\epsilon^2$-cover $\Pi$.
This means that for any $\pi\in \Pi$, there exists $\pi_j$, such that:
\begin{align*}
    \forall M > 0, \forall \{\{\tilde{s}_{1,t}\},\cdots, \{\tilde{s}_{M,t}\}\}, H_M(\pi,\pi_j) = \frac{1}{M}\sum_{i=1}^M \bm{1}(\{\bigvee_{t=1}^T\pi_{1,t}(s_{i,t})\neq\pi_{j,1,t}(s_{i,t})\}\vee\{\bigvee_{t=1}^T\pi_{2,t}(s_{i,t})\neq\pi_{j,2,t}(s_{i,t})\} \leq \epsilon^2.
\end{align*}
Pick $M=m\sum_{i=1}^n\lceil \frac{m|\mathcal{M}(\pi,\pi_j,\{s_{i,t}\}, \{\Gamma_{i,s_{1:T}}\})|^2}{\sup_{\pi^{\alpha},\pi^{\beta}}\sum_{i=1}^n|\mathcal{M}(\pi^{\alpha},\pi^{\beta},\{s_{i,t}\}, \{\Gamma_{i,s_{1:T}}\})|^2}\rceil + \sum_{i=1}^n\lceil \frac{m|\mathcal{M}(\pi,\pi_j,\{s_{i,t}\}, \{\Gamma_{i,s_{1:T}}\})|^2}{\sup_{\pi^{\alpha},\pi^{\beta}}\sum_{i=1}^n|\mathcal{M}(\pi^{\alpha},\pi^{\beta},\{s_{i,t}\}, \{\Gamma_{i,s_{1:T}}\})|^2}\rceil$ (where $m$ is some positive integer) and 
\begin{align*}
    \{\{\tilde{s}_{1,t}\},\cdots,\{\tilde{s}_{M,t}\}\}=\{\{s_{1,t}\},\cdots,\{s_{1,t}\},\{s_{2,t}\},\cdots,\{s_{2,t}\},\cdots,\{s_{n,t}\},\cdots,\{s_{n,t}\},\{s_{\ast,t}\},\cdots,\{s_{\ast,t}\}\},
\end{align*}
where $\{s_{i,t}\}~(1\leq i\leq n)$ appears $\lceil \frac{m|\mathcal{M}(\pi,\pi_j,\{s_{i,t}\}, \{\Gamma_{i,s_{1:T}}\})|^2}{\sup_{\pi^{\alpha},\pi^{\beta}}\sum_{i=1}^n|\mathcal{M}(\pi^{\alpha},\pi^{\beta},\{s_{i,t}\}, \{\Gamma_{i,s_{1:T}}\})|^2}\rceil$ times and  $\{s_{\ast,t}\}$ appears \\$m\sum_{i=1}^n\lceil \frac{m|\mathcal{M}(\pi,\pi_j,\{s_{i,t}\}, \{\Gamma_{i,s_{1:T}}\})|^2}{\sup_{\pi^{\alpha},\pi^{\beta}}\sum_{i=1}^n|\mathcal{M}(\pi^{\alpha},\pi^{\beta},\{s_{i,t}\}, \{\Gamma_{i,s_{1:T}}\})|^2}\rceil$ times.

Here, we pick $\{s_{\ast,t}\}$ such that $\bm{1}(\{\bigvee_{t=1}^T\pi_{1,t}(s_{\ast,t})\neq\pi_{j,1,t}(s_{\ast,t})\}\vee\{\bigvee_{t=1}^T\pi_{2,t}(s_{\ast,t})\neq\pi_{j,2,t}(s_{\ast,t})\}=1$.
Per the definition of $M$, we have:
\begin{align*}
    M &= (m+1)\sum_{i=1}^n\lceil \frac{m|\mathcal{M}(\pi,\pi_j,\{s_{i,t}\}, \{\Gamma_{i,s_{1:T}}\})|^2}{\sup_{\pi^{\alpha},\pi^{\beta}}\sum_{i=1}^n|\mathcal{M}(\pi^{\alpha},\pi^{\beta},\{s_{i,t}\}, \{\Gamma_{i,s_{1:T}}\})|^2}\rceil \\
    &\leq (m+1)\sum_{i=1}^n\left(\frac{m|\mathcal{M}(\pi,\pi_j,\{s_{i,t}\}, \{\Gamma_{i,s_{1:T}}\})|^2}{\sup_{\pi^{\alpha},\pi^{\beta}}\sum_{i=1}^n|\mathcal{M}(\pi^{\alpha},\pi^{\beta},\{s_{i,t}\}, \{\Gamma_{i,s_{1:T}}\})|^2}+1\right) \\
    &= (m+1)\frac{m\sum_{i=1}^n|\mathcal{M}(\pi,\pi_j,\{s_{i,t}\}, \{\Gamma_{i,s_{1:T}}\})|^2}{\sup_{\pi^{\alpha},\pi^{\beta}}\sum_{i=1}^n|\mathcal{M}(\pi^{\alpha},\pi^{\beta},\{s_{i,t}\}, \{\Gamma_{i,s_{1:T}}\})|^2} + n \leq (m+1)(m + n).
\end{align*}
Further, from the number of appearances of $\{s_{i,t}\}~(1\leq i\leq n)$ and $\{s_{\ast,t}\}$, we have:
\begin{align*}
    &H_M(\pi, \pi_j) = \frac{1}{M}\sum_{i=1}^M\bm{1}(\{\bigvee_{t=1}^T\pi_{1,t}(s_{i,t})\neq\pi_{j,1,t}(s_{i,t})\}\vee\{\bigvee_{t=1}^T\pi_{2,t}(s_{i,t})\neq\pi_{j,2,t}(s_{i,t})\} \\
    &= \frac{1}{M}\sum_{i=1}^n\lceil \frac{m|\mathcal{M}(\pi,\pi_j,\{s_{i,t}\}, \{\Gamma_{i,s_{1:T}}\})|^2}{\sup_{\pi^{\alpha},\pi^{\beta}}\sum_{i=1}^n|\mathcal{M}(\pi^{\alpha},\pi^{\beta},\{s_{i,t}\}, \{\Gamma_{i,s_{1:T}}\})|^2}\rceil\bm{1}(\{\bigvee_{t=1}^T\pi_{1,t}(s_{i,t})\neq\pi_{j,1,t}(s_{i,t})\}\vee\{\bigvee_{t=1}^T\pi_{2,t}(s_{i,t})\neq\pi_{j,2,t}(s_{i,t})\} \\
    &+ \frac{m}{M}\sum_{i=1}^n\lceil \frac{m|\mathcal{M}(\pi,\pi_j,\{s_{i,t}\}, \{\Gamma_{i,s_{1:T}}\})|^2}{\sup_{\pi^{\alpha},\pi^{\beta}}\sum_{i=1}^n|\mathcal{M}(\pi^{\alpha},\pi^{\beta},\{s_{i,t}\}, \{\Gamma_{i,s_{1:T}}\})|^2}\rceil \\
    &\geq \frac{m}{(m+1)(m+n)}\sum_{i=1}^n \frac{m|\mathcal{M}(\pi,\pi_j,\{s_{i,t}\}, \{\Gamma_{i,s_{1:T}}\})|^2}{\sup_{\pi^{\alpha},\pi^{\beta}}\sum_{i=1}^n|\mathcal{M}(\pi^{\alpha},\pi^{\beta},\{s_{i,t}\}, \{\Gamma_{i,s_{1:T}}\})|^2} \\
    &= \frac{m^2}{(m+1)(m+n)} \frac{\sum_{i=1}^n|\mathcal{M}(\pi,\pi_j,\{s_{i,t}\}, \{\Gamma_{i,s_{1:T}}\})|^2}{\sup_{\pi^{\alpha},\pi^{\beta}}\sum_{i=1}^n|\mathcal{M}(\pi^{\alpha},\pi^{\beta},\{s_{i,t}\}, \{\Gamma_{i,s_{1:T}}\})|^2} = \frac{m^2}{(m+1)(m+n)}I_{\Gamma}(\pi,\pi_j)^2
\end{align*}
Letting $m\to \infty$ yields: $\lim_{m\to\infty} H_M(\pi,\pi_j)\geq I_{\Gamma}(\pi,\pi_j)^2$.
Therefore, we have:
\begin{align*}
    I_{\Gamma}(\pi,\pi_j) \leq \epsilon.
\end{align*}
Consequently, the above argument establishes that for any $\pi\in \Pi$, there exists $\pi_j\in\{\pi_1,\cdots,\pi_K\}$, such that $I_{\Gamma}(\pi,\pi_j)\leq \epsilon$, and therefore $N_{I_{\Gamma}}(\epsilon, \Pi, \{\{s_{1,t}\}, \cdots, \{s_{n,t}\}\}\})\leq K = N_{H}(\epsilon^2, \Pi)$
\end{proof}

\section{Additional results of Experiment}
\label{sec:appendix_results}
Tables \ref{tab:exp_ope_rps1_2}, \ref{tab:exp_ope_rps2_2} show the results in the experiments in Section 6.2.
We provide additional results from the experiment in Section
\begin{table}[h!]
    \centering
    \caption{Off-policy exploitability evaluation in RBRPS1: RMSE (and standard errors).}
    \begin{tabular}{c|c|c|c|c|c} \hline
         $N$ & $\hat{v}^{\mathrm{exp}}_{\mathrm{IS}}$ & $\hat{v}^{\mathrm{exp}}_{\mathrm{MIS}}$ & $\hat{v}^{\mathrm{exp}}_{\mathrm{DM}}$ & $\hat{v}^{\mathrm{exp}}_{\mathrm{DR}}$ & $\hat{v}^{\mathrm{exp}}_{\mathrm{DRL}}$ \\ \hline \hline
         $250$ & \begin{tabular}{c}
              $0.085$ \\
              $(8.48\times 10^{-3})$
         \end{tabular} & \begin{tabular}{c}
              $0.232$ \\
              $(3.69\times 10^{-3})$
         \end{tabular} & \begin{tabular}{c}
              $4.8\times 10^{-3}$ \\
              $(4.4\times 10^{-4})$
         \end{tabular} & \begin{tabular}{c}
              $\mathbf{3.6\times 10^{-3}}$ \\
              $(3.4\times 10^{-4})$
         \end{tabular} & \begin{tabular}{c}
              $4.5\times 10^{-3}$ \\
              $(4.2\times 10^{-4})$
         \end{tabular}  \\ \hline
         $500$ & \begin{tabular}{c}
              $0.065$ \\
              $(6.4\times 10^{-3})$
         \end{tabular} & \begin{tabular}{c}
              $0.230$ \\
              $(2.8\times 10^{-3})$
         \end{tabular} & \begin{tabular}{c}
              $6.9\times 10^{-5}$ \\
              $(6.4\times 10^{-6})$
         \end{tabular} & \begin{tabular}{c}
              $\mathbf{3.6\times 10^{-5}}$ \\
              $(3.5\times 10^{-6})$
         \end{tabular} & \begin{tabular}{c}
              $6.1\times 10^{-5}$ \\
              $(5.8\times 10^{-6})$
         \end{tabular} \\ \hline
         $1000$ & \begin{tabular}{c}
              $0.044$ \\
              $(4.3\times 10^{-3})$
         \end{tabular} & \begin{tabular}{c}
              $0.226$ \\
              $(1.8\times 10^{-3})$
         \end{tabular} & \begin{tabular}{c}
              $2.9\times 10^{-9}$ \\
              $(2.8\times 10^{-10})$
         \end{tabular} & \begin{tabular}{c}
              $\mathbf{1.1\times 10^{-9}}$ \\
              $(1.1\times 10^{-10})$
         \end{tabular} & \begin{tabular}{c}
              $2.5\times 10^{-9}$ \\
              $(2.4\times 10^{-10})$
         \end{tabular} \\ \hline
    \end{tabular}
    \label{tab:exp_ope_rps1_2}
\end{table}

\begin{table}[h!]
    \centering
    \caption{Off-policy exploitability evaluation in RBRPS2: RMSE (and standard errors).}
    \begin{tabular}{c|c|c|c|c|c} \hline
         $N$ & $\hat{v}^{\mathrm{exp}}_{\mathrm{IS}}$ & $\hat{v}^{\mathrm{exp}}_{\mathrm{MIS}}$ & $\hat{v}^{\mathrm{exp}}_{\mathrm{DM}}$ & $\hat{v}^{\mathrm{exp}}_{\mathrm{DR}}$ & $\hat{v}^{\mathrm{exp}}_{\mathrm{DRL}}$ \\ \hline\hline
         $250$ & \begin{tabular}{c}
              $36.6$ \\
              $(2.54)$
         \end{tabular} & \begin{tabular}{c}
              $11.3$ \\
              $(0.83)$
         \end{tabular} & \begin{tabular}{c}
              $7.07$ \\
              $(0.16)$
         \end{tabular} & \begin{tabular}{c}
              $8.98$ \\
              $(0.88)$
         \end{tabular} & \begin{tabular}{c}
              $\mathbf{6.52}$ \\
              $(0.38)$
         \end{tabular} \\ \hline
         $500$ & \begin{tabular}{c}
              $21.7$ \\
              $(1.55)$
         \end{tabular} & \begin{tabular}{c}
              $11.2$ \\
              $(0.68)$
         \end{tabular} & \begin{tabular}{c}
              $6.04$ \\
              $(0.25)$
         \end{tabular} & \begin{tabular}{c}
              $6.10$ \\
              $(0.61)$
         \end{tabular} & \begin{tabular}{c}
              $\mathbf{5.56}$ \\
              $(0.39)$
         \end{tabular} \\ \hline
         $1000$ & \begin{tabular}{c}
              $15.5$ \\
              $(1.22)$
         \end{tabular} & \begin{tabular}{c}
              $11.1$ \\
              $(0.50)$
         \end{tabular} & \begin{tabular}{c}
              $4.87$ \\
              $(0.32)$
         \end{tabular} & \begin{tabular}{c}
              $\mathbf{4.33}$ \\
              $(0.41)$
         \end{tabular} & \begin{tabular}{c}
              $4.39$ \\
              $(0.34)$
         \end{tabular} \\ \hline
    \end{tabular}
    \label{tab:exp_ope_rps2_2}
\end{table}

% \newpage
% \bibliographystyle{plain} %参考文献出力スタイル
% \bibliography{ref}

% \end{document}

%%%%%%%%%%%%%%%%%%%%%%%%%%%%%%%%%%%%%%%%%%%%%%%%%%%%%%%%%%%%%%%%%%%%%%%%

\end{document}